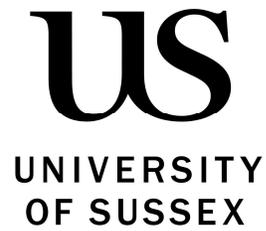

# Can humans help BERT gain "confidence"?

### Piyush Agrawal | 250944

# Declaration

I hereby declare that this project has not been and will not be submitted in whole or in part to another University for the award of any other degree.

Piyush Agrawal, 30th Aug 2022

# Abstract

The advancements in artificial intelligence over the last decade have opened a multitude of avenues for interdisciplinary research. Since the idea of artificial intelligence was inspired by the working of neurons in the brain, it seems quite practical to combine the two fields and take the help of cognitive data to train AI models. Not only it will help to get a deeper understanding of the technology, but of the brain as well. In this thesis, I conduct novel experiments to integrate cognitive features from the Zurich Cognitive Corpus (ZuCo) (Hollenstein et al., 2018) with a transformer-based encoder model called BERT. I show how EEG and eye-tracking features from ZuCo can help to increase the performance of the NLP model. I confirm the performance increase with the help of a robustness-checking pipeline and derive a word-EEG lexicon to use in benchmarking an external dataset that does not have any cognitive features associated with it. Further, I analyze the internal working mechanism of BERT and explore a potential method for model explainability by correlating it with a popular model agnostic explainability framework called LIME (Ribeiro et al., 2016). Finally, I discuss the possible directions to take this research forward.

# Acknowledgments

I would like to thank Dr. Julie Weeds for supervising me on this thesis and for all of her support, suggestions, advice, and patience throughout the course. The meetings and conversations were vital in inspiring me and encouraging me to keep learning. I would also like to thank Ahmed Younes, Lorenzo Bertolini, and Justina for all their help and support.

Piyush Agrawal, 30th Aug 2022

# Contents

**Appendices**

# Chapter 1: Introduction

Natural language processing (NLP) has progressed a lot in recent years. The development of transformers has made a huge impact in the field of artificial intelligence and it is being used for a plethora of tasks (Vaswani et al., 2017). While transformers are the current state-of-the-art for many NLP tasks they don't understand language (Bishop, 2021; Dubova, 2022). In contrast, the human brain is an effortless language machine. An average human can easily perform language tasks without needing to train on vast amounts of data. Advancements in Electroencephalography (EEG) and eye-tracking systems have enabled us to record brain data while performing reading tasks. This enables us to see how the brain responds while performing any natural reading task and decode it to find patterns of comprehension (Ling et al., 2019). This encourages the use of computational neuroscience integrated with artificial intelligence to reinforce artificial models with some comprehension capability (McClelland et al., 2020). Since we are the domain experts, we should be able to provide the assistance of our domain knowledge to the NLP models so we can take a step forward in building true artificial language understanding models. Physiological processes recorded from the brain are studied using specific frequency bands or "brain oscillations/waves" named in Greek alphabets delta, theta, alpha, beta, and gamma. An abundance of research explains these oscillations depending on the tasks performed. For language, the brain resolves attention, memory, and comprehension (syntactic and semantic). Williams et al. (2019) suggest that theta activity increases when focusing attention on the current task that involves short-term memory. This corroborates the observation by Bastiaansen et al. (2002) suggesting an increase in theta activity during real-time language comprehension. A real-time comprehension function can be a response to a semantic anomaly that increases the delta/theta activity (Prystauka & Lewis, 2019). The role of the alpha band is long-term memory involving autonomous processes suggests Williams et al. (2019). Kilmesh (2012) reports the involvement of alpha-band activity during temporal attention. Language translation task involves temporal attention and Grabner et al. (2007) observed that theta and alpha oscillations are sensitive to the success and the difficulty of translation tasks. Beta-band has been associated with more complex linguistic functions such as semantic retrieval of lexicons, parsing sequences, and generating correct sentences (Weiss and Muller, 2012). Lastly, Prystauka & Lewis (2019) suggest that gamma-band activity is sensitive to semantic manipulations and factual inconsistencies. They found that gamma activity increased when a piece of incorrect factual information was presented. The N400 response in the brain was discovered in the 1980s as an indicator of reading comprehension and linguistic manipulations. N400 is a late event-related potential that manifests around 400ms after the event as a negative peak (Holcomb, 1993). Since then, the N400 response has been shown to manifest not only in the language domain but as a shared event between other domains with comparable cognitive resolution requirements (Dudschig, 2022). Nevertheless, the N400 response is a good candidate to identify comprehension onsets. Grabner et al. (2007) report theta-band response around 200-600ms after linguistic stimulus such as word presentation, including alpha, and beta-band inhibition 200-400ms after the stimulus. The increase in the gamma-band activity due to factually incorrect stimulus was also around 400-600ms after the stimulus (Prystauka & Lewis, 2019)

There is an increasing interest in combining neuroscience and machine learning in the natural language processing domain. Many researchers are trying to incorporate the fields to capitalize on the benefits. For example, Heilbron et al. (2022) used the language model GPT-2 to show evidence of predictive processing in the brain. In other words, the brain's response to a word is driven by probabilistic predictions. Likewise, cognitive and/or physiological features like EEG and eye-tracking data have been used to improve NLP models. In a study by Sood et al. (2020), they proposed a hybrid text saliency model (TSM) that can predict the human gaze in free text reading. They show a high correlation of the proposed model with human gaze ground truth. Further, they integrate their model into two attention-type NLP models for paraphrase generation and sentence compression tasks. They show that their model outperforms the previous state-of-the-art (in 2020) with a BLEU-4 score increase of 10%. In another research, Barrett et al. (2016) trained a Hidden-Markov-Model with maximum entropy with raw texts, a dictionary, and gaze feature data to improve parts-of-speech tagging. Other studies by Hollenstein et al. (2019) and Hollenstein et al. (2021) which this thesis builds on, are discussed later.

The aim of this research is to leverage cognitive features from the Zurich Cognitive Corpus (ZuCo) which provides EEG and eye-tracking recordings from human subjects during natural reading tasks (Hollenstein et al., 2018) to modify the architecture of BERT which is a transformer-based encoder model. The expectation is that the models augmented with the cognitive features should perform better than the vanilla BERT model in a text classification task.

## 1.1 A brief background of word embeddings

Recent advancements in NLP started with word embeddings which are n-dimensional vector representations of words in a document. Previous methods used statistical techniques for natural language processing tasks. Some of the important ones are Term Frequency Inverse-document Frequency (TFIDF) for extracting word importance by counting the frequency of occurrence of words and filtering out very rare words like proper nouns and highly repeated words like articles, prepositions, etc. as they generally might not contribute to overall document understanding (Uther et al., 2011), Latent Semantic Analysis (LSA) to extract contextually similar documents by counting co-occurrence of words in a document and assuming that the words which frequently appear together share contextual information (Foltz, 1996), and Latent Dirichlet Allocation (LDA) which is used for topic modeling. It assumes that each document can have multiple topics and each topic can have frequently occurring words that define that topic to calculate the probabilities of finding them. For example, if a document's content is about sports and finance, the words "football", "golf", etc. will appear in sports while "stock", "trade", etc. will appear in finance. This can be segregated using LDA (David et al., 2003). All of these techniques can be used individually or in combination to generate document representations, calculate document similarity, and other backend NLP tasks. One disadvantage that statistical models face is they are non-contextual at the core and they discover relations based on assumptions without taking semantics into account (Hogenboom et al., 2010).

Mikolov et al. (2013) introduced Word2Vec. They proposed two novel methods to capture the similarity between texts in a document. They used neural networks to train a continuous-bag-of-words (CBoW) model which used the distributed representation of the words to try and find the probability of a word occurring in a given set of surrounding words. The other proposed approach was the Skip-Gram model which is the opposite of the CBoW model where the input is the target word and the model tries to predict the words around it.

A year later, Pennigton et al. (2014) proposed the Global Vectors for Word Representation (GloVe) model. They focused primarily on the co-occurrence frequency of the words which was not given as much importance by Word2Vec. The word embeddings that were generated as a result of co-occurrences of words in the whole corpus, reflected the probability of two words occurring together within some window.

Other embedding models have been proposed since then (Bojanowski et al., 2017). These models were pre-trained on a large corpus since one word can appear in the vicinity of lots of other words. The result represents the words in the form of n-dimensional vectors and the words that are similar in meaning occupy neighboring vector spaces. For example, "woman" and "queen" should be closer to each other in the vector space as compared to "man" and "queen". These vectors can be accessed from the models as lookup tables. These embedding models are still not contextual since they are in the form of a lookup table which is static. This means that the word "playing" in "Playing with my friend" and "Playing with my brother" will have the same vector. To turn these into contextual embeddings, neural networks can be used as discussed in the next chapter.

## 1.2 Bidirectional Encoder Representation from Transformers (BERT)

Sequence models such as Recurrent Neural Networks (RNN) enjoyed all the "attention" until Vaswani et al. (2017) proposed the transformer model. In summary, RNNs are a type of neural network that take can resolve temporal data or sequential data like text. It has a cell-type structure where each cell takes one word and the previous hidden state as input and generates and the output from the last cell is used as shown in figure 1 (Sherstinsky, 2020). There are many variations in RNNs in terms of using their outputs, for example, the output of each cell is used as opposed to just the last cell for named entity recognition but the one described above is for text classification.

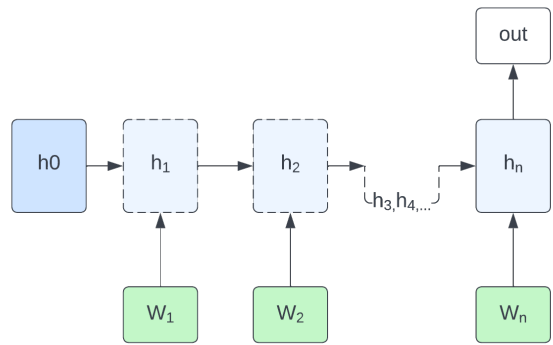

**Figure 1 Recurrent neural network for text classification:** h0 through hn are the hidden state where h0 is the initial hidden state. W1 through Wn are the input tokens at each time step, the number of tokens determines the number of time steps. The last hidden state gives the output in text classification.

The contextual embeddings are the hidden states from the model which is learned during training and word embeddings can be initialized using Word2Vec, GloVe, or any other embedding models. The RNN faced an issue of vanishing gradient that made it difficult for the model to learn longer sequences. Due to this, the Long-short term memory (LSTM) was proposed which incorporated a memory module to output the current cell state to keep track of longer sequences (Sherstinsky, 2020). These cell states could be understood as an information highway to transfer data between cells.

While LSTMs alleviated the issue of vanishing gradient, they still couldn't process very long sequences like paragraphs and took a long time to train due to being sequential in nature, in other words, the processing couldn't be parallelized. In 2017, Vaswani et al. proposed the transformer architecture that aimed to resolve all these issues. It became one of the most influential papers in natural language processing as it could do everything that the sequence models could and more. Figure 2 shows the transformer architecture.

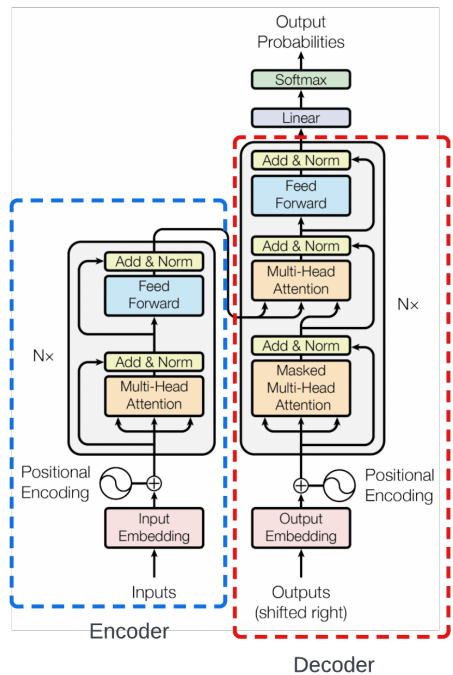

**Figure 2 Architecture of the Transformer (Vaswani et al., 2017; Modified):** The left block (in blue) of the transformer model is the encoder and the right block (in red) is the decoder.

The model is made of two components, an encoder, and a decoder. The encoder aims to resolve the internal semantic relation in the input sequence and encodes it in the form of hidden states. The decoder takes input from the encoder and tries to learn how it relates to the output. The authors trained the transformer model for neural machine translation which is a conventional task for encoder-decoder models and reported state-of-the-art performance. This study focuses on BERT which as its name suggests, uses only the encoder from the transformer. Devlin et al. (2019) leveraged the encoder from the transformer to create a pre-trained framework that can be fine-tuned for multiple NLP tasks. They proposed a model with stacked encoders to generate contextual embeddings where each word appearing in different contexts will have different embeddings, unlike the Word2Vec or GloVe models. It is bi-directional because instead of reading individual tokens per time step (from left to right or right to left) as RNNs do, it can read the entire sequence at once. Figure 3 shows the encoder with self-attention functionality.

The encoder consists of two embedding layers, one to generate word embeddings for each word in the input sentence and the other to generate position embeddings for the position of each word in the input sentence. The words in the input sentences are represented by integer tokens called input ids or token ids and for each token id, the word embedding layer generates an n-dimensional embedding vector. The positions of the words in the input sentence are represented as a count vector [1,2,3……, n] and the position embedding layer generates an n-dimensional vector for each position integer. Then the word embedding is summed with the position embedding.

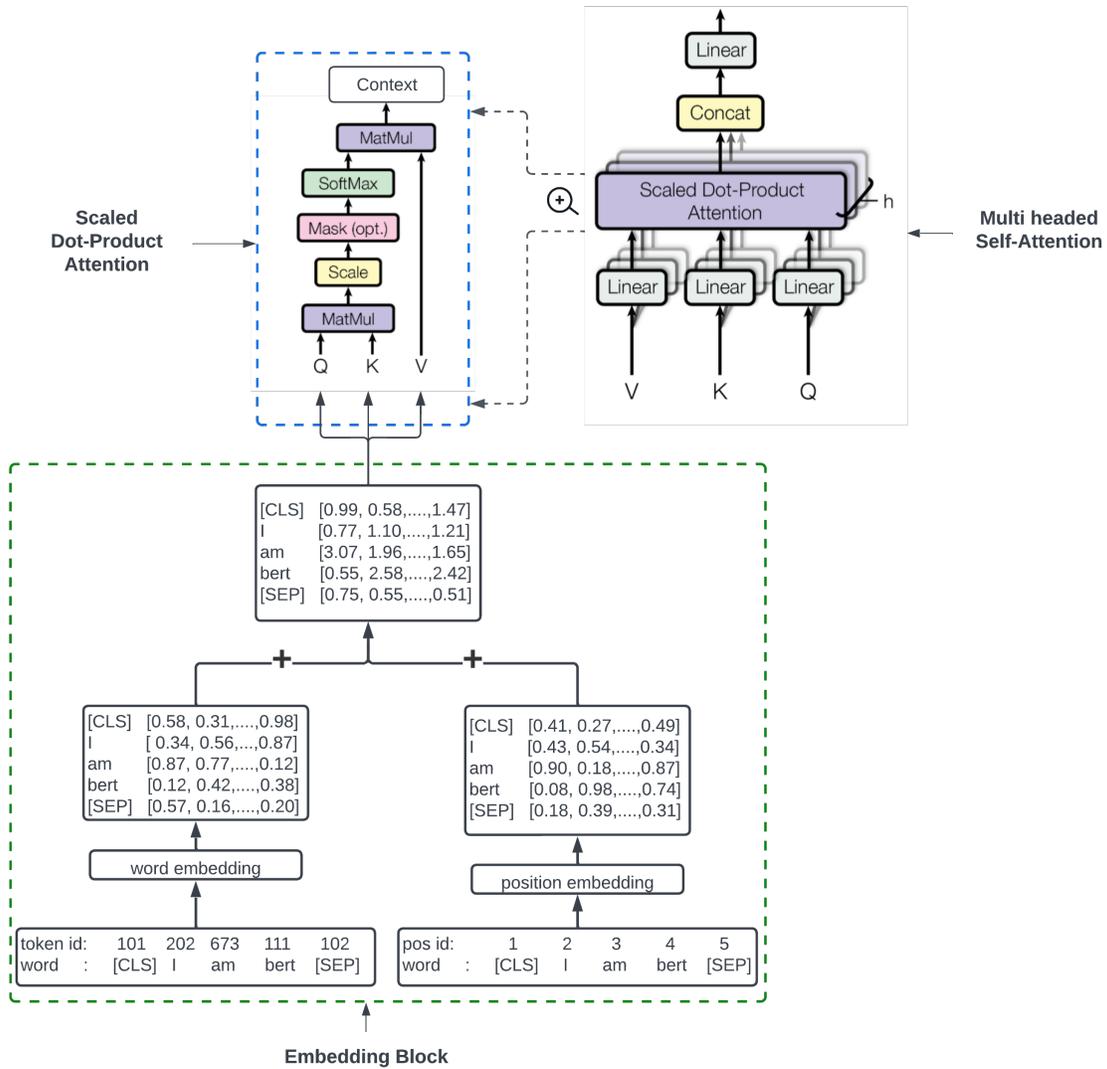

**Figure 3 Embedding and Self-attention mechanism in the default BERT (Vaswani et al., 2017; Modified):** The block in blue shows the exploded view of the Scaled Dot-Product Attention. The bottom block (in green) shows the two embedding layers (word and position) in the default BERT model. The final value after addition for each word is an n-dimensional embedding vector. The contexts are concatenated and passed to the linear layer as shown in the Multi-Headed Self-Attention as shown in the right block.

The result of the sum is passed to the self-attention module as query, key, and value so that the self-attention heads can calculate attention scores for each word. The query, key, and value receive the same embeddings as each word has to resolve its relationship with every other word in the sentence. The attention score is calculated by multiplying the query vector with the transpose of the key vector using matrix multiplication so the output vector represents how much attention each word is assigned to every other word in the sentence. Then these scores are added with an attention mask, scaled, and converted into probabilities with the softmax layer. When these probabilities are again multiplied with the same word embeddings which is the value vector, the output represents the final context (please refer to figure 3 for a visual explanation). This is how the vanilla BERT works and the cognitive features will be integrated by adding additional embedding layers or modifying the attention scores in this architecture.

The authors designed the model to be trained on large datasets so that the model could encode significant amounts of contexts and learn from many variations of the semantic patterns for a word. They trained it using two unsupervised learning tasks to help the model resolve contexts autonomously, Masked language Modeling (MLM), and Next Sentence Prediction (NSP). In MLM, the authors randomly masked 15% of the words in the input sequence by replacing the token with [MASK] and training the model to predict those tokens. Since they designed the model to be used for fine-tuning in other downstream tasks which can be different from MLM, if a token was selected randomly for masking, they replaced it with [MASK] 80% of the time, and the other 20% they replaced the token with a random token or didn't replace it all. This way the model can learn in an unsupervised manner and also expect inputs without any masking if the fine-tuning task is different. The second unsupervised task they trained the model in is Next Sentence Prediction (NSP) where the model learns to predict if two sentences occur subsequently in a document. The model receives two sentences as input where 50% of the time the second sentence is subsequent to the first in the original document and the other 50% a random sentence is selected from the document as the second sentence. The task assumes that the random sentence will be disconnected in context from the first sentence. The authors added two special tokens to the sentences [CLS] or "classification" with the token integer 101, to represent the meaning of the sentence so that it can be used to predict if the second sentence is subsequent to the first, and [SEP] or "separator" with the token integer 102, to differentiate between the two sentences. To identify which tokens belong to which sentence, they added sentence embedding that indicates the sentence type (A or B) for each token. The model is trained in both tasks together with the target of minimizing the combined loss function.

## 1.3    Related work

Hollenstein et al. (2019), used a relation classification system by Rotsztejn et al. (2018) which is a combination of convolutional neural networks and recurrent neural networks to integrate it with the cognitive features from ZuCo. They extracted word-level EEG features from the dataset and appended them to GloVe embeddings to be used as input to the relation classification model which increased the performance. They also proposed a method to generate a dictionary of word-level EEG features that can be used with other datasets that do not provide any cognitive features. In another research, Hollenstein et al. (2021) showed the advantage of using EEG features with BERT embeddings. However, their setup was complex which is bound to increase training time and add a lot of additional parameters on top of the large language model. They use BERT embeddings as input to a bi-LSTM model while keeping the BERT parameters trainable as the baseline. Further, they experiment with two approaches to decode the EEG features by using bi-LSTM and an inception Convolution Neural Network that uses multiple filters of different lengths to extract features from the EEG data (Szegedy et al., 2015). They concatenate the output of the EEG decoder to the output of the BERT bi-LSTM to be passed to the classifier. This setting does not directly modify the BERT model but uses it as input to their proposed models. In contrast, experimentation with the architecture of the BERT model to incorporate the cognitive features has not been done before. This research studies the impact of architectural modification at different levels of the BERT model using cognitive features to check if performance and "comprehension" can be increased while keeping the complexity low to match the training time and computation resource requirement of fine-tuning the vanilla BERT model. The motivation for experimenting with the architecture of the BERT model is to explore new potential opportunities to develop human cognition-inspired NLP models that can co-relate better with the human comprehension of language while also providing insights into the internal working mechanisms of large language models like BERT.

## 1.4   ZuCo

The Zurich cognitive corpus (ZuCo) dataset contains EEG and eye-tracking recordings from 12 subjects performing natural reading tasks. The task for the scope of this research is the relation classification task. All the subjects were assessed for their English language comprehension before the experiments through the LexTALE test (Lexical Test for Advanced Learners of English). The sentences for the relation classification task were chosen from the Wikipedia relation extraction dataset has 1110 paragraphs mentioning 4681 relations in 53 relation types. ZuCo roughly has 40 sentences each in 8 relation types (award, education, job title, political affiliation, wife, visited, nationality, founder) that were presented to the subjects.

The sentences were presented on screen with a grey background and black text with double-spaced words and triple-spaced lines. A maximum of 7 lines were present simultaneously in this task. This setting provided enough room to keep track of the eye movements. The subjects had access to a control pad to switch the sentences and answer the questions allowing them to read the sentences at their own pace and replicate a natural reading environment. This also gave the subjects freedom to fixate on words for as long as they deemed and in the pattern that they wanted. Interestingly, the reading speed for this task for the subjects was lower than passively reading similar sentences as they had to look for a specific relation type to answer the questions. They had to report whether a relation type was present for each sentence and there were 72 control sentences in the mix that did not have any relation type to check if there is actual comprehension of the sentences. Due to the presence of the control sentences, the sentences were presented in chunks of the same relation type so that the subjects did not have to read the questions again. Before starting the experiment, they had a practice round to familiarize themselves with the control and the stimuli.

The instructions were presented as follows (Hollenstein et al., 2018):

"AWARD; while reading the following sentences please watch out for the relation between a person or their work and the award they/it received or were nominated for."

"Please read the following sentences. After you read each sentence, answer the question below. Press 6 when you are ready."

**The question after the sentence**: "Does this sentence contain the *award* relation? [1] = Yes, [2] = No"

# Chapter 2: Cognitive features from ZuCo

## 2.1    Eye-tracking

The EyeLink 1000 tracker system was used to identify fixation, saccades, eye position, and blinks. Saccades are frequent eye movements that orient the eye to different points in the visual field and determine the current input to the visual system i.e. the current word in the sentence (Foulsham, 2015). The tracker was calibrated with a 9-point grid in the monitor where participants were instructed to pay attention to dots presented randomly at 9 locations on the screen and the calibration process was repeated until the error rate was within the recommended average of 1 degree for all the points. The unit of measurement of the eye movement speed is deg/sec and the acceleration and velocity thresholds for the tracker to detect a saccade are 8000deg/sec2, and 30deg/sec respectively and the eye has to deflect within 0.1 degrees to be categorized as a saccadic movement by the tracker.

The tracker has a sampling rate of 500Hz and spatial resolution of <0.01 degrees which is supposed to provide high precision tracking. The duration of eye fixation is measured as the time period without saccades where the eye pauses at a location for a short time interval. The fixations were assigned over the x and y coordinates of pixels in the screen where the words represent the x-coordinate and the changing line represents the y-coordinate. Furthermore, a Gaussian mixture model was trained on the y-axis (lines) data to improve the fixation accuracy by clustering the gaze points (x-axis) to their respective gaussians (y-axis) where the number of lines defined the number of gaussians to fit.

A Gaussian mixture model is a clustering technique that can identify distributions in a 2-D space along with the shape or pattern of the distribution. The number of Gaussians are the number of clusters to allocate to the data points (Reynolds, 2009). Ideally, the model result should be distinct lines in a 2-D space with the x coordinates (words) clearly allocated across those lines so that each word that is paid attention to is in a specific line. Figure 4 shows an illustration of the trained Gaussian mixture model. The blinks were detected when pupil diameter was zero or when the x and y gaze positions were zero and discarded along with fixations shorter than 100ms as fixations shorter than 100ms most likely are irrelevant for reading tasks (Sereno & Rayner, 2003). A total of six eye tracking features i.e. number of fixations, GD (Gaze duration), FFD (First-fixation duration), TRT (Total reading time), SFD (Single fixation duration), and GPT (Go-past time) were extracted from the process.

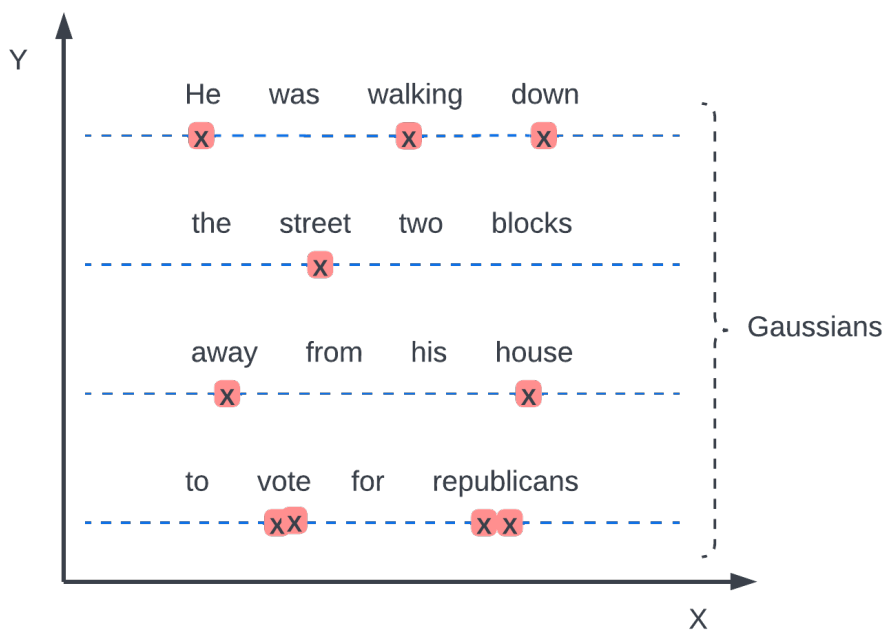

**Figure 4 Illustration of the trained Gaussian Mixture Model:** The blue lines in the figure represent Gaussians in the y-axis and the red 'x's in the x-axis represent the gaze points. There are 4 Gaussians to train and allocate the gaze points. Each gaze point has a (x,y) coordinate that denotes its position in the pixels of the screen.

## 2.2 Electroencephalography (EEG)

The EEG data was recorded with a 128-channel non-invasive electrode system from EGI. The data was recorded with a sampling rate of 500Hz and filtered to retain wave frequency between 0.1 – 100Hz using the bandpass filtering method. The EEG system was calibrated to maintain the electrode impedance of 40kOhm and checked after every 60 sentences as a high impedance can decrease the EEG amplitude and increase the noise in the data. Out of the 128 channels, 105 were used for recording the scalp, 9 were used as EOG channels for artifact removal, and the rest of the channels that were placed on the neck and the face were discarded as they did not provide any essential data for processing. The artifacts in question are the signals generated from eye movements and other muscular movements like eye blinks. These can produce signals that are higher in magnitude as compared to the EEG signals and they can travel across the scalp which can add noise to the scalp electrodes (Zeng & Song, 2014). The EOG channels were placed on the subject's forehead, the inner (where the eye joins the nose), and the outer canthi (the opposite side). The artifacts were identified using MARA (Multiple Artifact Rejection Algorithm) which is an open source plug-in for an effective supervised machine learning algorithm that analyses the EEG data so that the artifacts can be automatically rejected.

MARA is trained on six features in the spatial, spectral, and temporal domains to analyze ICA (Independent component analysis) components from the EOG signals and calculates the probability of the component being an artifact, and rejects it if the probability is greater than 0.52 like a binary classification task (Winkler et al., 2011). The identified artifacts were removed by linearly regressing them from the EEG signals. Figure 5 shows the example of the raw EEG data and the pre-processed EEG data for a sentence.

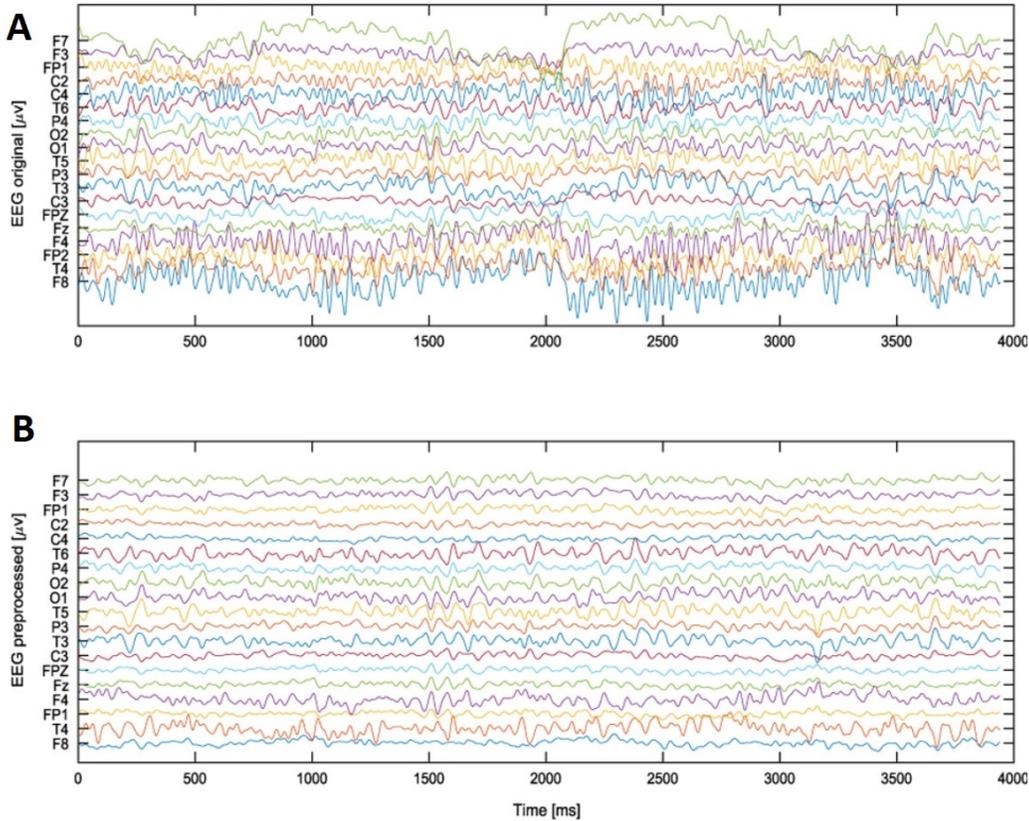

**Figure 5 Raw and Pre-processed EEG data for an example sentence (Hollenstein et al., 2018): (**A) shows the raw EEG data, (B) shows the pre-processed EEG data. The y-axis shows the brain regions where electrodes were placed. F=frontal, FP=pre-frontal, C=central, T=temporal, P=parietal, O=occipital. The x-axis shows the time.

       To extract word-level EEG features, the EEG and eye-tracking data were synchronized by identifying shared events using the EYE EEG extension that can time lock the EEG data to the onset of eye fixations. The EEG features are spread across multiple frequency bands with each band serving its own cognitive function as discussed earlier. A total of 8 bands were identified and the data was band-pass filtered to get theta1 (t1: 4-6Hz), theta2 (t2: 6.5-8Hz), alpha1 (a1: 8.5-10Hz), alpha2 (a2: 10.5-13Hz), beta1 (b1: 13.5-18Hz), beta2 (b2: 18.5-30Hz), gamma1 (g1: 30.5-40Hz), and gamma2 (g2: 40-49.5Hz) frequencies. The final EEG features are power measures for each of these frequency bands in 105 channels that are the electrodes placed on the scalp. The power measurement of the EEG signals indicates the total activity in each frequency band per channel (Xiao et al., 2018). Calculating the power of a frequency band was a multi-step process, the first was applying Hilbert transformation to all the frequency bands to estimate their amplitude and phase since it maintains the temporal information required to identify the segments of fixation in the synchronized eye-tracking data. Hilbert transformation is a method to shift the phase of a signal by 90 degrees in the complex plane and derive its imaginary part which is the transformed signal. Hence, for the next step, it was possible to compute the power for those fixated segments to get one value for each transformed frequency band in 105 channels. Also, the sentence-level EEG feature was derived by calculating the power of each frequency band over the full spectrum in 105 channels. Each of the eye-tracking features mentioned above (FFD, TRT, GD, GPT, and SFD) are the events that have associated EEG data in all eight frequency bands. Hence, the resulting EEG features are 8*5(8=#frequency bands, 5=#gaze features) 105-dimensional vectors for each fixated word in a sentence and eight 105-dimensional vectors for the full sentence.

# Chapter 3: Dataset and feature modeling

The cognitive features of the subject with the highest evaluation scores from the ZuCo study are used as Hollenstein et al, (2019) suggested that this single-subject model performed better than the model with averaged features across all the subjects.

## 3.1 Eye-tokens

Five eye-tracking features are used to generate the Eye tokens. The number of fixations is the number of times a word was fixated on, FFD (First-fixation duration) is the duration of fixating on a word for the first time, TRT (Total-reading time) is the sum of all duration of fixations of the word, GD (Gaze duration) is the sum of all fixations of the word during the first pass before the eye moves out of the word, and GPT (Go-past time) which is the sum of all fixations before the eye moves to the next word towards the right. The eye tokens are calculated as:

$$E_t = nFixation * (FFD + TRT + GD + GPT)$$

Where $E_t$ is the eye token for the current word that is fixated upon. If a word is not fixated on by the subject, it is assigned an eye token of 0. The resulting Eye token values were normalized and scaled to hold values between 0-100 for each sentence.

## 3.2 EEG tokens

Analogous to the word token that represents a word using an integer that is used as the input to the word embeddings, the EEG token is an integer generated from the word-level EEG features in the ZuCo dataset. It represents the mean brain activity for each word that was fixated upon. In the dataset, each word has a 105-dimension EEG vector in 8 frequency bands for five fixation-related potentials (FRP) i.e a total of 8*5 105-dimension vectors for that word in the sentence, and 4*8 of them were used to generate the EEG tokens. The four fixation-related events are FFD, TRT, GD, and GPT. A sample FFD set for a word looks like (FFD_t1, FFD_t2, FFD_a1, FFD_a2, FFD_b1, FFD_b2, FFD_g1, FFD_g2) and this is the same for other FRPs. The column-wise mean of all 4*8 EEG vectors generates a single 105-dimension EEG vector for a word and the scaled mean of all 105 values in this single vector is the EEG token for the word. The words that were not fixated on by the subject are assigned 0 as their EEG token.

## 3.3 Sentence-level EEG data

The ZuCo dataset provides 105-dimensional sentence-level EEG feature vectors in eight frequency bands, theta1, theta2, alpha1, alpha2, beta1, beta2, gamma1, gamma2. The column-wise mean of these EEG vectors gives a single 105-dimension vector for a sentence.

## 3.4 Cognitive attention mask

Bert uses an attention mask to manipulate the attention scores. It is required so that the model can ignore the [PAD] tokens. The attention mask that Bert uses by default assigns a value of -0 to indexes containing a token including [CLS] and [SEP] and -10000 to indexes containing [PAD]. When this mask is added to the attention score, the content tokens retain their score while the attention score of the [PAD] token diminishes. The cognitive feature mask for each word can be defined as a function of the number of fixations.

$$f(nFixation) = \begin{cases} -0, & nFixation > 1 \\ -10000, & nFiaxtion = 0 \end{cases}$$

The cognitive attention mask assigns a score of -10000 to the words that were not paid attention to by the subject along with the [PAD] token. If the number of fixations for a word is greater than 1, it is assigned -0 as the mask and for all others, it is -10000.

There are a total of 334 sentences in the ZuCo dataset in 9 relation types. A total of 302 sentences in 8 relation types were used and it was divided into training and testing sets with an 80-20% split resulting in 236 training and 66 test samples. The sentences were pre-processed by lowercasing each word and removing punctuations. The trainer library from huggingface[1] was used to train and evaluate the models so the dataset was converted into a[1] dictionary format using the dataset library from huggingface[2] containing the 'input_ids', 'token_type_ids', 'attention_mask', and 'labels' for both training and testing sets to pass to the trainer. A database is maintained to store all the cognitive features described in the features section for each of the 302 sentences. The database was created by tokenizing the input sequence using the Bert tokenizer and mapping the tokens to their respective cognitive features. A lookup function is added to the original code that accepts a batch of input ids, loops through each, and returns the cognitive features for that batch from the database.

---

# Chapter 4: Augmenting BERT

The aim is to augment the architecture of the BERT model with the cognitive corpus in meaningful ways to check if it can improve model performance and analyze the changes in the model. The cognitive features are incorporated at three different levels in BERT's architecture. BERT-base from huggingface[12] is used for all the experiments with a max sequence length of 512 tokens and 768-dimensional embedding vectors.

## 4.1    Embedding function

The default embedding layer sums the input and the position embeddings which informs the model about the semantic relation of the words in a sentence. The intuition behind augmenting the embedding layer was to check if the cognitive features can add information about the "human way" of text comprehension using different cognitive features. Much like the BERT model learns the position embedding during training (Wang et al., 2021), the hypothesis was the EEG tokens can help the model to learn the pattern of brain activity recorded during reading in a relation classification task and the Eye Tokens can help the model to learn the relevance pattern of words during classification. Adding the input, position, and cognitive features should tell the model about the word content, its position, and its "activity" i.e. the average band power for the EEG token and relevance for the Eye token. To achieve this, additional embedding layer(s) was added to the BERT embedding function that takes the cognitive features as input and generates cognitive embeddings. This process makes the cognitive features compatible for addition with the input and the position embeddings. A total of three experiments were performed with the word-level EEG tokens, the word-level Eye tokens, and both word-level EEG and Eye tokens together. In the case of word-level EEG and Eye tokens together, two embedding layers were added to take both the features as input and generate their respective embeddings. Figure 6 shows the embedding mechanism after augmentation.

Input to the word embedding layer are the token ids for a sequence. Each sentence is represented as a vector of 512 tokens where the token 0 signifies [PAD], token 101 is for [CLS], and token 102 is [SEP]. The embedding layer outputs an embedding vector of 768 dimensions for each of those 512 tokens. Hence the final output shape is (batch_size, 512, 768) for the sentences. The position embedding layer takes a range vector starting from 1 and ending at 512 [1, 2, 3, .....,512] as input and outputs a 768 dimension embedding for each number in the range. Hence, the final shape remains the same as the word embedding layer, i.e. (batch_size, 512, 768).

Since the word-level, EEG and Eye tokens hold a value for each word in a sequence including padding, they have the same length as the sequence. Hence, the EEG and Eye embedding layer that takes their respective feature tokens as input share the same output dimensions as the word and position embedding layers, and they can be added together.

Pre-trained weights for BERT-base were initialized for fine-tuning on the dataset. All the models were trained for 15 epochs with a batch size of 8. The learning rate was initially set to 5e-05 and a learning rate scheduler gradually diminishes it to 0 as the training iteration progresses. This is an optimal setting because of the high number of training parameters. Due to the small size of the dataset, i.e. 236 samples for training and 66 samples for testing, the experiments were repeated 10 times for each of the cognitive features. Average evaluation metrics for all the runs are reported.

---

[1] https://huggingface.co/bert-base-uncased

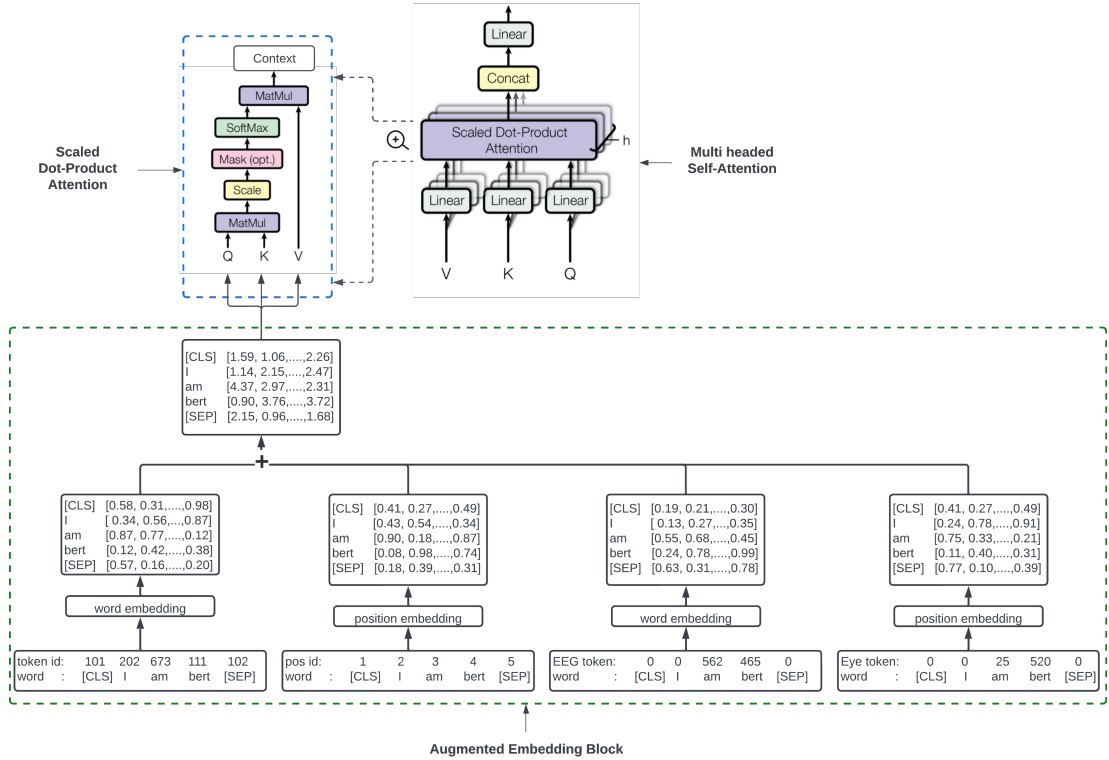

**Figure 6 Augmented embedding block with self-attention mechanism (Vaswani et al., 2017; Modified):** The left block (in blue) shows the exploded view of the Scaled Dot-Product Attention. The bottom block (in green) shows the four embedding layers (word, position, EEG token, and Eye token) in the augmented BERT model. The self-attention mechanism remains the same. The EEG and Eye tokens can be added individually or together depending on the experiment but the figures show them added together.

## 4.2    Self-attention function

The intuition behind modifying the self-attention function was to filter out or "not-pay" attention to the words that the human subject did not pay attention to. In a relation classification type task, the human subjects pay attention to keywords without the need to gather a lot of context about the sentence which has been empirically studied and shown by (Hollenstein et al., 2018). The hypothesis was to inject similar behavior using an attention mask derived from the tokens that the human paid attention to and develop a keyword representation as opposed to a contextual representation of sentences. This might seem counter-intuitive to learning about the language but it is a logical experiment when it pertains to text classification. For example, consider the sentence "Sam is a physicist, he won the Nobel prize". To relate this sentence to the label "AWARD", the necessary keywords to look for are "won", "Nobel", and "prize". These human-deemed important keywords can point to the class label. In contrast, gathering contextual information is necessary for tasks such as question answering where the word "he" needs to be related to "Sam" when answering the question "Who won the Nobel prize?"

The variable of interest in the self-attention function is the attention mask. By default, in the Bert self-attention function, the attention score is calculated by adding it to an attention mask.

The attention mask assigns a value of -0 in positions where the tokens exist including [CLS] and [SEP] and -10000 where [PAD] exists. For example, the attention mask for the sentence [[CLS], "he", "won", "the", "nobel", "prize", [SEP], [PAD], [PAD]] looks like [-0, -0, -0, -0, -0, -0, -0, -10000, -10000]. The output of this default attention mask only diminishes the attention score for the [PAD] token so that the model can ignore it. With the proposed modification, the new attention mask looks like [-0, -10000, -0, -10000, -0, -0, -0, -10000, -10000]. Since the words "he" and "the" are not relevant for the class label, they are assigned a low score for the model to ignore. Though there isn't a cognitive feature representation for the [CLS] token, it is necessary to have the model pay attention to it as the output of the final hidden state for the [CLS] token is used for classification. Figure 7 shows the augmented attention mask in the self-attention module.

The attention score is calculated for each token, hence the attention score has a shape (batch_size, 1, 1, 512). The default attention mask has the same shape as the attention score so they are compatible for addition. The proposed attention mask has a shape of (batch_size, 512) and it had to be reshaped to (batch_size, 1, 1, 512). This isn't a problem as the dimension of 1 can be added to the shape of any vector any number of times and it doesn't change the original vector. This shape means that there are 512 scores for each sentence in the batch.

The new attention mask was extracted from the cognitive lookup function for each batch and reshaped to match the dimensions of the attention score. The original attention mask was replaced with the proposed attention mask during addition. The model was trained for 15 epochs with a linearly diminishing learning rate starting from 5e-05. The experiment was repeated 10 times and average evaluation metrics are reported.

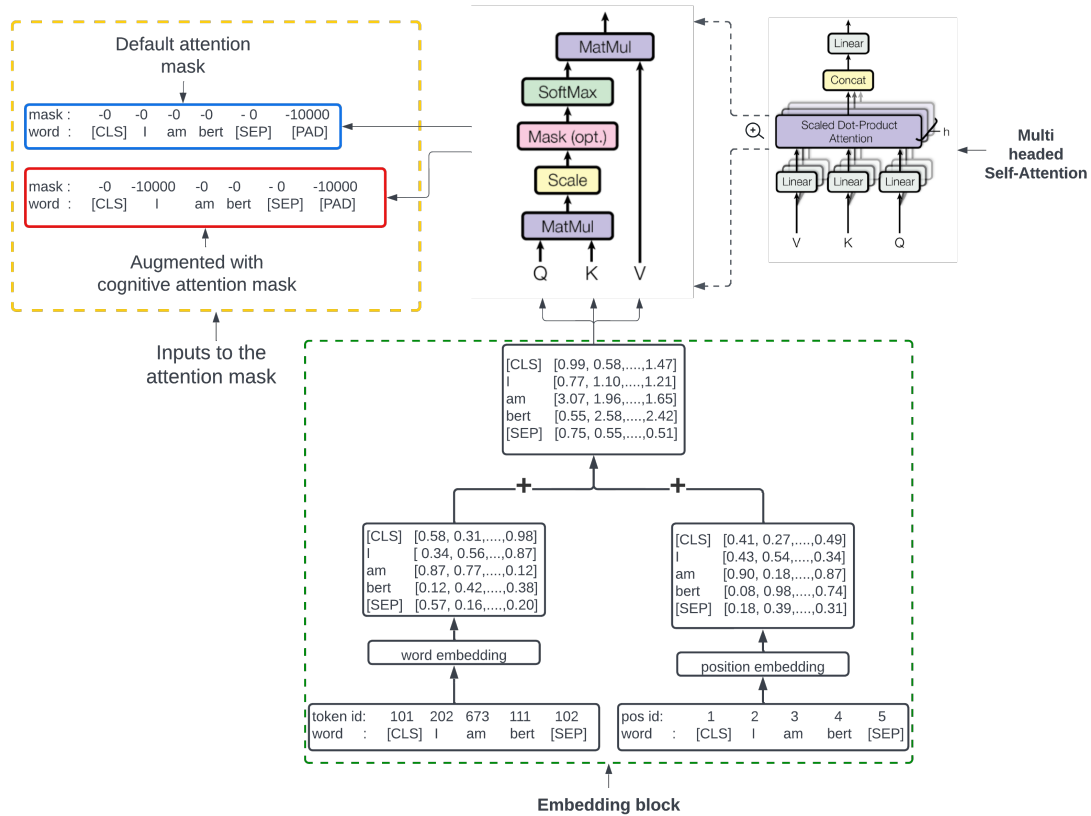

**Figure 7 Augmented attention mask in the self-attention module (Vaswani et al., 2017; Modified):**
The left block (in yellow) shows the input to the attention mask for the default model (in blue box) and augmented model (in red box). This figure assumes that the word "I" was not fixated upon by the human subject hence it was assigned a diminished value similar to [PAD]. The embedding block (in green) remains the same as the default BERT model.

## 4.3    Bert for sequence classification function

The Bert for sequence classification function is used to fine-tune the embeddings for text classification. A classifier is implemented in the function that takes the pooled output i.e. the output of the [CLS] token from the last hidden state as input to a linear layer and outputs the labels. Experimenting with the downstream task function is logical as modifying the model at lower levels i.e. the embedding and self-attention function modifies the model's interpretation of the input and the cognitive features become an inherent part of the model output. Modifying the downstream function and incorporating the cognitive features on top of the model output results in the model tweaking its own interpretation to accommodate the cognitive features during training which is in line with (Hollenstein et al., 2021). In this setting, sentence-level cognitive features can be used as the pooled output also represents the model interpretation at the sentence level. In the embedding and self-attention function experiments, the cognitive features were the word-level representation of the features where each word in a sentence was required to have its EEG and Eye token and an attention mask to make it compatible for integration with the functions. Since the pooling output represents the sentence, it can be incorporated with the sentence-level EEG features which is a 105-dimension vector for each sentence. It is important to experiment with downstream functions as it can shed light on the ability of the model to incorporate cognitive features at the architecture level compared to its ability to fine-tune its own interpretation to accommodate the additional features. Three methods were tested to integrate the sentence-level EEG features with the pooled output. These experiments were chosen to keep the training time and/or training parameters on par with the vanilla BERT model so that the model can benefit from the performance increase while maintaining the time and compute cost. Figure 8 shows an illustration of the experiments done to integrate sentence-level EEG to the pooled output.

### 4.3.1    Concatenation

This experiment was performed in two ways. The first was concatenating the 105-dimension EEG vector for each sentence to the 768-dimension pooled output from the model. The second method was to pass the EEG features into the neural network which is described below and concatenate the output from the last layer of that network to the pooled output. As the default classifier accepts 768-dimensional input, it had to be changed to accept 873 dimensions for the first method and to 1536 dimensions for the second one as input after concatenation.

### 4.3.2    Multiplication

Every value of the pooled output was multiplied with each value of the EEG vector and averaged. This operation is a bit complex but it integrates the EEG vector over the full pooled output and doesn't change the original dimension of 768 vectors.

$$W_i = \frac{\sum_{j=0}^{j=105} W_i * X_j}{768}$$

### 4.3.3    Addition

This method allows the direct addition of the EEG vectors to the pooled output by passing the vector to a simple neural network containing three layers. The first layer takes the 105-dimensions as input and outputs a 768-dimension vector. The second and the third layer takes 768-dimensions as input and outputs the same dimensional vector. GELU activation is used in the first and the second layers. This makes the EEG vector representation trainable through the neural network so it can be updated during training to fit the data better.

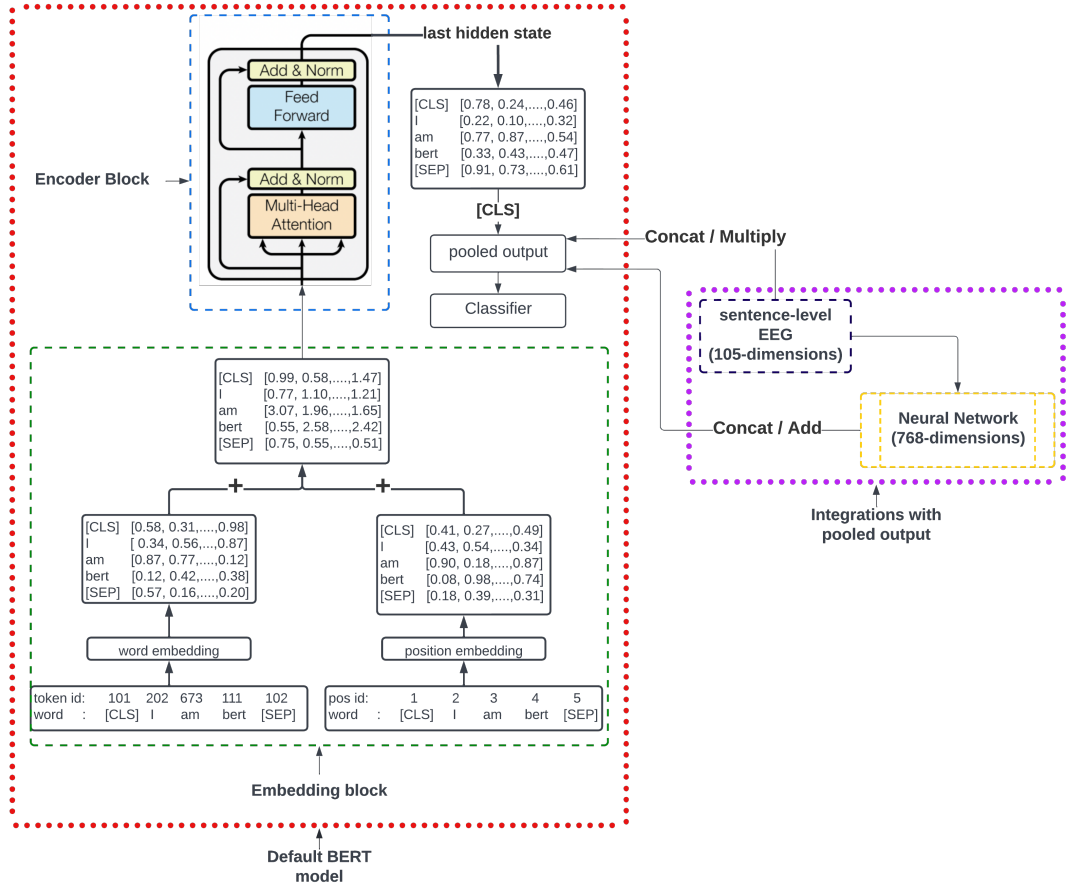

**Figure 8 Integration of sentence-level EEG features to the pooled output (Vaswani et al., 2017; Modified):** The red block shows the default BERT model without any additions to the pooled output. The purple block shows four ways of integrating sentence-level EEG features to the pooled output.

Each method was trained for 15 epochs with the learning rate scheduler starting from 5e-05 and linearly decaying to 0. The neural network experiment was repeated 10 times, the concatenation experiments were repeated 5 times, and the multiplication experiment was performed once. The average evaluation metrics are reported depending on the experiments.

# Chapter 5: Results and analysis

The hypothesis that various cognitive features can be incorporated at different levels in the BERT architecture to increase model performance holds. Precision, recall, F-1 score, and accuracy are reported on the test set which is not seen during training. All of these metrics convey important details about the results. Precision informs about the number of correctly predicted positive cases out of the total number of positive case predictions.

$$Precision = \frac{True\ Positives}{True\ Positives + False\ Positives}$$

Recall, on the other hand, informs about the number of correctly predicted positive observations out of all the observations in the class.

$$Recall = \frac{True\ Positives}{True\ Positives + False\ Negatives}$$

The F1 score combines precision and recall as a weighted average which can take both false positives and false negatives into account.

$$F1 = 2 * \frac{Recall * Precision}{Recall + Precision}$$

Lastly, accuracy is the ratio of the total correct observations to the total number of observations. Accuracy can be a tempting metric as it is very intuitive. However, when dealing with multiple classes, the imbalance between the number of records in each class can be high. Accuracy would not serve as the correct prediction metric in the case of imbalanced classes as the model can have a high overall accuracy if the model correctly predicts only the abundant classes. In our scenario, there isn't a huge difference between classes in the dataset hence accuracy is also reported.

$$Accuracy = \frac{True\ Positives + True\ Negatives}{True\ Positives + False\ Positives + True\ Negatives + False\ Negatives}$$

Since there is still class imbalance, the macro average of precision, recall, and F1 scores are reported as it adjusts for the imbalance by reporting the metrics per class without taking the proportion of each class into account.

## 5.1 Model evaluation

Table 1 shows the evaluation metrics for the experiments. The pre-trained weights for BERT base with 110million parameters were used as the baseline and augmentations. The first row shows the average evaluation the baseline model for comparison. The addition of EEG tokens in the Embedding layer of the BERT model was to check if the model can learn the brain activity pattern and increase classification performance. The model did perform better as compared to the baseline with the addition of EEG tokens in the embedding layer which can be observed from the second row of table 1. There was an average increase of 1.5% across all the metrics. Also, out of the 10 runs for the experiments, the vanilla BERT model reached a max accuracy of 67% with a probability of 2/10 whereas the model augmented with EEG tokens reached a max accuracy of 70%, and the probability of getting 67-70% accuracy was 6/10.

Augmenting the model with Eye Tokens in the embedding function shows comparable performance to the baseline model. The third row of table 1 shows the performance of this experiment. Both models reached a max accuracy of 67% out of 10 runs and share similar metric results along with similar probabilities of getting the max accuracy. Since the intuition was to check if the model can learn to capture the relevance pattern of words or if the model can learn to pay higher attention to keywords that define a class, another dimension of evaluation is presented i.e. analyzing the change in the attention layers of the model which discussed later.

The next logical step was to check if the model can learn both relevance and activity from the cognitive features by adding both EEG tokens and the Eye tokens to the embedding layer. Row 4 in table 1 shows the performance of using both features. The performance did increase by an average of 0.7% as compared to the baseline while maintaining the same training time as the vanilla BERT but it fell short in precision and recall by about 1% as compared to using only EEG tokens. Also, the highest accuracy model out of 10 runs was 67%. The attention heads of the model favorably changed to accommodate both the features which is discussed in the attention analysis part.

Using the cognitive mask feature to modify the self-attention function of the model was an interesting experiment. Unfortunately, the evaluation metrics did not turn out to be great as seen in row 5 of table 1. The training time increased by 4x for the same number of epochs as compared to the vanilla model. This might be solvable with efficient coding but due to this and not observing good performance, this experiment was repeated 5 times where the model reached a max accuracy of 64%. Since BERT is an encoded representation of language, its architecture requires relating each word to every other word in the sentence to "understand" or encode language. With the cognitive mask, the words that were not fixated upon by the human subject were also trained to be ignored by the model as they were assigned a diminished attention score of -10000 which meant that the model could not use those words to relate them with other words in the sentence. This might be one of the reasons that the model performed poorly in the classification task as it was not able to form the encoded representation as per its requirements. On the other hand, interesting sets of observations emerged from the attention analysis which are discussed later.

Concatenating is a very practical approach to integrating features from different sources as it does not alter the data and keeps all the information from the sources. There were two options for concatenating the sentence-level EEG features with the pooled output. Either the 105-dimensional feature vector can be directly concatenated with the pooled output with 768 dimensions from BERT resulting in an 873-dimension final vector or concatenating the 768-dimensional output from the neural network to the pooled output resulting in a 1536-dimensional vector as input to the classifier. (Do et al., 2017) suggests that using non-linearity helps to form a richer representation of the features which is why the second method of concatenation was tested. The performance is reported in row 6 of table 1. Both of these methods unfortunately did not perform well in this case. In fact, the performance was about 3 times worse than the vanilla model. A possible explanation for these results is that concatenation might require a more complex decoding structure than simple neural networks which is analogous to the experiments performed by Hollenstein et al. (2021).

The results for the multiplication method fell apart very quickly. Table 1, row 7 shows the performance for this experiment. It was a way to incorporate all the 105 values from the sentence-level EEG vector into the pooled output without changing the default 768 dimensions of the BERT model. However, the training time for the implemented code increased by 32x as compared to the training time of the vanilla model which is why this experiment only had a single run. The reason why this experiment might have failed is that the multiplication process completely changed the model's interpretation of the input sequence by a great margin.

The best metric results were achieved by passing the sentence-level EEG feature which is a 105-dimension vector to a simple three-layered neural network and adding it to the pooled output from the BERT model before passing it to the classifier. The performance increased by 2-3% across all the metrics while keeping the training time same as the vanilla model. This method is an addition to the downstream function for sequence classification and it does not change the model architecture. The ninth row in table 1 shows the average metrics of 10 repetitions of the experiment. In addition, the highest accuracy out of the 10 runs was 70% with this model, with a probability of 6/10 for reaching 67-70% accuracy compared to a 2/10 probability of 67% of the vanilla model. The reason for the increase in performance using this setup is that the classifier is essentially receiving two different features to take advantage of. The sentence-level EEG acts as an additional feature-set for the classifier to rely on if it fails to increase performance with the pooled output. (Ramachandram and Taylor, 2017) published an elaborate survey on how setups like this can be complementary for complex classification tasks as richer permutations can emerge with the combination to fit the data better which is in line with the experiments performed by (Hollenstein et al., 2021)

| Model | Pre-trained Bert Base | | | |
|---|---|---|---|---|
| | **P** | **R** | **F-1 (std)** | **A** |
| Vanilla \| Baseline | 0.6502 | 0.6489 | 0.6353 (0.0211) | 0.6492 |
| Embedding function +EEG Token | 0.6612 | 0.6623 | 0.6447 (0.0280) | 0.6597 |
| Embedding function +Eye Token | 0.6402 | 0.6552 | 0.6339 (0.0107) | 0.6492 |
| Embedding function +EEG&Eye Token | 0.6494 | 0.6586 | 0.6415 (0.0113) | 0.6552 |
| Cognitive attention mask | 0.6053 | 0.6062 | 0.5921 (0.0375) | 0.6059 |
| Pooling Concatenation sent-EEG | 0.1372 | 0.2055 | 0.1264 (0.0164) | 0.2477 |
| Pooling Concatenation  NN-EEG | 0.1219 | 0.1986 | 0.1232 (0.0185) | 0.2461 |
| Pooling Multiplication | 0.1418 | 0.1943 | 0.1153 (0) | 0.2312 |
| Pooling Addition with NN | **0.6710** | **0.6792** | **0.6611 (0.0231)** | **0.6731** |

**Table 1: Evaluation results of comparison between the baseline and augmented models:** The table shows the comparative results between all the models using pre-trained weights from BERT-base. P=Precision, R=Recall, A=Accuracy, F-1 score with std = standarad deviation. **Bold =** best results

## 5.2    Robustness check

To determine the importance of each cognitive feature and to assess if the increase in performance was in fact due to the integration of cognitive features, a robustness check is performed by training the BERT model with the same experiments without using any pre-trained weights. BERT for built for training on large datasets and training it from scratch with a small dataset like ZuCo is bound to give poor performance. Nevertheless, the poor performance should be consistent across all the experiments if the cognitive features are not helping the model. So, this setup provides a good basis to check if the features are responsible for the performance increase. A randomizing function was used to refresh the layers of the pre-trained BERT-base model. Each experiment was repeated 5 times and trained for 10 epochs. Table 2 shows the results from the robustness check. The first row depicts the vanilla model without any augmentations, the model did not perform well as expected and only reached an average accuracy of 27%. The EEG token model performed better than the baseline in the pre-trained experiment and the same is true for the scratch experiment with a 43.5% accuracy other metrics can be observed in row 2. The augmentation with Eye tokens in the embedding function of the model performed similarly to the vanilla model which is in line with the pre-trained scenario as can be seen in row 3 of the table. Using EEG and Eye tokens together also results in a performance increase over the baseline with 39.7% accuracy which is shown in row 4 of the table. The neural network model had the highest performance in the pre-trained scenario which is also true in the scratch test with an accuracy of 53%. This result corroborates how the classifier might be relying on the EEG features when the pooled output alone cannot perform well. Using the cognitive attention mask in the scratch model did not perform well which is again, in line with the pre-trained experiments. Overall, the results from the robustness check conformed to the expectations and corroborated the pre-trained experiments. The cognitive features do help the model.

| Model | Scratch Bert Base | | | |
|---|---|---|---|---|
| | P | R | F-1 (std) | A |
| Vanilla \| Baseline | 0.1874 | 0.2893 | 0.2094 (0.0560) | 0.2716 |
| Embedding function +EEG Token | 0.4120 | 0.4304 | 0.3985 (0.0927) | 0.4358 |
| Embedding function +Eye Token | 0.2618 | 0.2988 | 0.2520 (0.0615) | 0.2805 |
| Embedding function +EEG&Eye Token | 0.3887 | 0.3657 | 0.3522 (0.0539) | 0.3970 |
| Cognitive attention mask | 0.2602 | 0.2578 | 0.2490 (0) | 0.2835 |
| Pooling Addition with NN | **0.5272** | **0.5079** | **0.4737 (0.0426)** | **0.5313** |

**Table 2: Robustness evaluation results for the baseline and augmented models:** The table shows the comparative results between all the models with randomized weights in BERT-base. Results for concatenation and multiplication with the pooled output were not recorded due to initial poor performance. P=Precision, R=Recall, A=Accuracy, F-1 score with std = standarad deviation. **Bold =** best results

## 5.3   Benchmarking

Now that the benefits of using cognitive features are evident, there is one caveat. The cognitive features are required at test time to see any performance increase. If the features are removed during test time by importing the cognitive feature augmented models on top of the vanilla BERT code, the performance decreases again. A possible solution to this problem is to use the word-level EEG features to generate a word-EEG lexicon by taking the average of the 105-dimensional EEG vector for each word occurrence in the dataset. The final result will be a dictionary of words and a single 105-dimensional EEG feature representing that word from the whole corpus. This lexicon contains 2253 words that can be used to derive a rough sentence-level EEG feature vector by taking the average of word-level EEG data for the words that appear in the sentence. This crud implementation was tested as a benchmark method to check if the model can gain a performance increase on the dataset that doesn't have any associated cognitive features. For benchmarking, only the neural network method was tested as the focus was to check the performance rather than explainability, and the neural network model performed the best in the ZuCo dataset. The Wikipedia relation extraction dataset by (Culotta et al., 2006) was used for the benchmarking. 1126 training sentences and 275 test sentences were extracted from the dataset for the same 8 relation types (*award, jobtitle, political_affiliation, wife, visited, education, nationality, founder*) used in ZuCo. Each experiment was repeated 10 times and trained for 10 epochs with hyper-parameters same as the previous experiments. Table 3 shows the performance of the benchmark for vanilla vs the neural network-integrated model. The performance of the neural network EEG integrated model did increase as compared to the vanilla model using only a small word-EEG lexicon database and a loose representation of the actual sentence-level EEG features. Moreover, the augmented model reached the highest accuracy of 73.8% compared to 73% of the vanilla model. These results further corroborate the potential benefits of using cognitive features with BERT.

| Model | Benchmark results | | | |
|---|---|---|---|---|
| | P | R | F-1 (std) | A |
| Vanilla \| Baseline | 0.64749 | 0.7201 | 0.6736 (0.0110) | 0.7214 |
| Pooling Addition with NN | **0.6556** | **0.7187** | **0.6795 (0.0114)** | **0.7262** |

**Table 3: Benchmark result comparison Vanilla vs Neural Network:** The table shows the comparative results between all the baseline and Neural Network augmented models for the benchmark dataset. BERT-base was used and initialized with pre-trained weights. Benchmarking was done only with these two models as the Neural Network augmented model showed the highest performance metrics increase. P=Precision, R=Recall, A=Accuracy, F-1 score with std = standard deviation. **Bold =** best results

An important point to note is that lower or similar performance for this particular task of text classification is not necessarily an adverse result as model explainability and attention are crucial dimensions of evaluation that are discussed in the upcoming sections. A single sentence can have keywords relating to multiple classes which might not be present in the training or testing set. For example, consider the sentence,

"Clinton received a degree from the school of foreign service at georgetown university in washington dc and won a rhodes scholarship to the university of oxford"

This sentence can be classified as having an "education" relation since it contains keywords like "school", "university", "degree", and "scholarship". It can also be classified as having an "award" type relation as it contains keywords like "won", "rhodes", and "scholarship". A very important insight that can be derived from the attention and explainability analysis is that, if the model is classifying this sentence as either of those classes, the keywords mentioned above for each class should emerge as the prime contributor that led to those classifications through model explainability.

## 5.4 Model explainability

The classification results alone cannot account for model performance as explained in the previous section. It is important to know why the model assigned specific classes to sentences. In other words, which keywords were important in the sentence for the model to make the classification decision? Artificial intelligence models are known to be black boxes which means that the internal working mechanisms are difficult to keep track of due to the complexity and large parameters. Popular explainability techniques have emerged in recent years to try and tackle this issue in ingenious ways. The one used in this research is called Local Interpretable Model-Agnostic Explanations or LIME for short (Ribeiro et al., 2016). LIME is a model agnostic framework which means that it can explain any type of model such as statistical (SVM, KNN) or fuzzy (Neural networks). The inputs to the model are slightly changed by randomly removing words from the input sentence and generating new input sequences for the model to predict and the outputs are observed. Then the cosine distance between the perturbed sequences and the original sequence is calculated to determine the weights. Finally, a surrogate model which is usually a linear regression model is trained using the perturbed sequences and the observed labels with the cosine distances as weights. If the perturbations are far from the keyword of interest, then it adversely affects the fit of the surrogate model. Hence, each of the original words in the sequence can be explained using this technique. (Szczepański et al., 2021) have shown the benefits of combining BERT with LIME and Anchor which is another explainability technique to identify the behavioral pattern of predictions in the model while detecting fake news. The transformer-interpret library which is built on top of Facebook's Captum[1] framework is used to extract the LIME weights for the test data predictions. Consider the examples shown in figure 9. Figure 9 (A) shows the sentence with the "founder" relation type and figure 9(B) shows the sentence with the "education" relation type. All the models classified these sentences accurately. The first observation is that the models were correctly able to identify the keywords "graduated", [3]"university", and "school" for the "education" relation type and "founded" for the "founder" relation type. The second observation is the LIME scores for keywords in the models augmented with cognitive features are higher as compared to the vanilla model which means that the keywords that define the relation type are assigned higher importance in cognitive models than in the baseline. This observation is consistent across the test set for the relation types that are predicted the same by all the models as seen in the examples in figure 9 except for the cognitive attention mask as it had the least metric scores. (please see appendix A for full results).

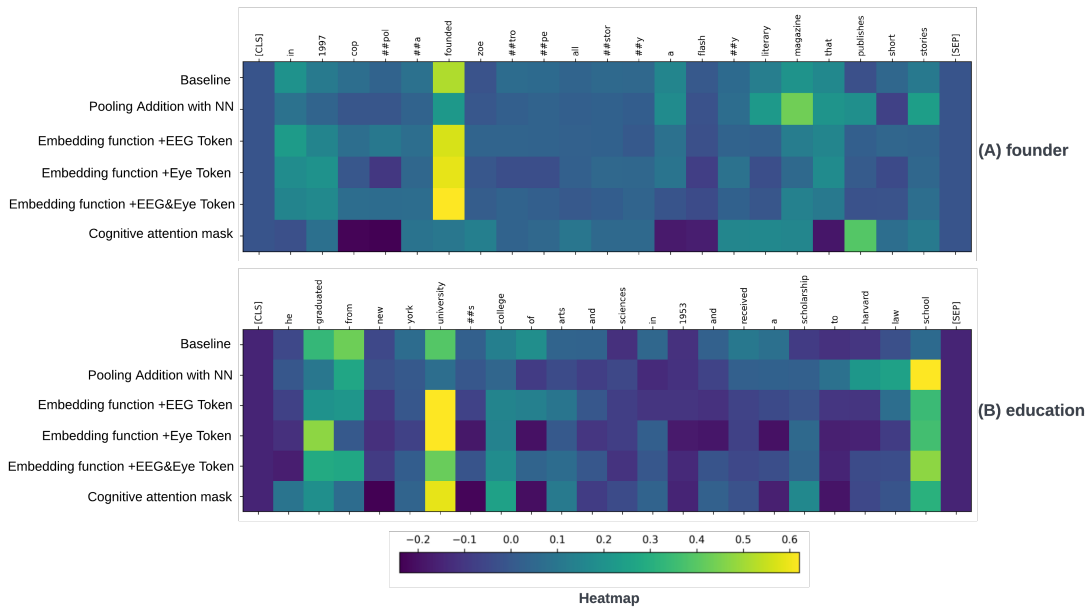

**Figure 9 Heatmap of Lime scores for all models:** The figure shows the LIME score heat map for six models for two example sentences with relation types (A) founder and (B) education. Models with concatenation and multiplication were left out due to poor performance during evaluation.

## 5.5 Attention weights analysis

While LIME can explain which words are important for the model to predict a class, it cannot explain the internal workings of the model as it is a model agnostic framework. Finding insights into the internal workings of models like BERT is necessary to explain how the model "thinks" before answering and most importantly if those insights can co-relate with LIME explanations. This would be another way of model explainability which would not only justify the model predictions but also answer why is it making those predictions, so one could have the ability to make modifications to the model to get the desired results. Self-attention is the core component of BERT that encodes language to learn context and an essential part of this research is to analyze how the attention weights change with the cognitive augmentations and if high attention weighted words can explain model predictions.

---

[3] https://captum.ai/

The self-attention function in BERT helps the model relate each word to one another in a sentence and the attention weights define the pattern of attention that the model assigns to the sentence during training. Fine-tuning as the name suggests tunes these attention weights for the downstream task it is trained on. Since each word is related to every other word in the sentence, they have incoming attention values.

There are a total of 12 hidden layers in BERT-base with each layer having 12 attention heads i.e. 144 attention heads in total. The attention weights are plotted for the test set. The incoming attention for each word in the sentence was extracted by first, a layer-wise summation of the heads resulting in 12 attention heads. Then a head-wise summation results in a single head, and finally, a column-wise summation results in one incoming attention value for each word from all the layers and all the heads as shown in figure 10. All the attention weights are derived from the best-performing model in each experiment.

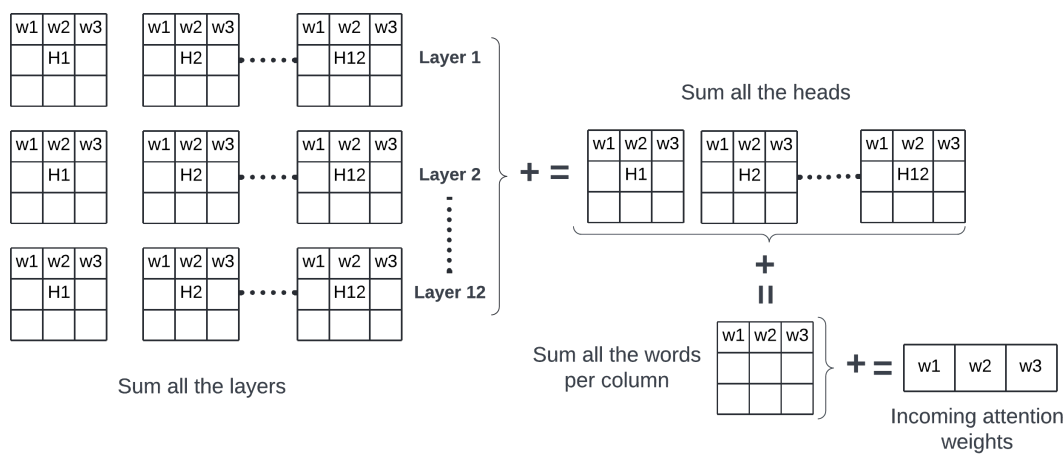

**Figure 10 Illustration of accumulating the incoming attention weights:** The figure shows the addition operations performed to get the incoming attention weights for each word in the sentence. There are 12 head blocks (H1, H2, ...., H12) per layer and there are 12 layers. The rows in a head block are words paying attention to w1, w2, and w3. This means the incoming attention to w1, w2, and w3 will be sum of their column.

With the addition of cognitive features, the expectation was the model would be able to learn brain activity and relevance patterns from the EEG and Eye tokens and "pay" better attention to the words in the sentence by assigning higher weights to them as compared to the vanilla model. Figure 11 shows attention weights comparison between five models for two sentences. The first row represents the weights for the baseline model, the second row for the model with the neural network's output added to the pooled output, and the third, fourth, and fifth show the weights for the models with augmentations in the BERT embedding function using EEG tokens, Eye tokens, and both respectively. The general trend of the models augmented with cognitive features shows that those models did pay higher attention to keywords as compared to the baseline model. (please see appendix B for all the attention plots).

The interesting part is how the model learns brain activity and/or word relevance from the augmentations. Consider the example as shown in figure 11 (A), the EEG token model in row 3 i.e. the model using EEG tokens in the embedding function assigns a very high weight to the word "the". This selective behavior for "the" word can be observed across the attention plots (please see appendix B). The reason the model does this is that the sum of EEG tokens that represents the brain activity for the word "the" in each sentence in the training corpus is the highest at 77971. However, from figure 11 (B) it can be seen that the same model did not assign high attention weights to "the" for this sentence. The reason was that the human subject did not have an associated brain activity for "the" as they did not pay attention to this word in this sentence so the model assigned high attention scores to other words that have associated brain activity and high EEG token sums like "moved". Hence, whenever the subject has an associated brain activity for "the" in a sentence, the model assigns high attention weight to it which can be observed from the attention plots. These insights inform about how the model might be working and learning from brain activity.

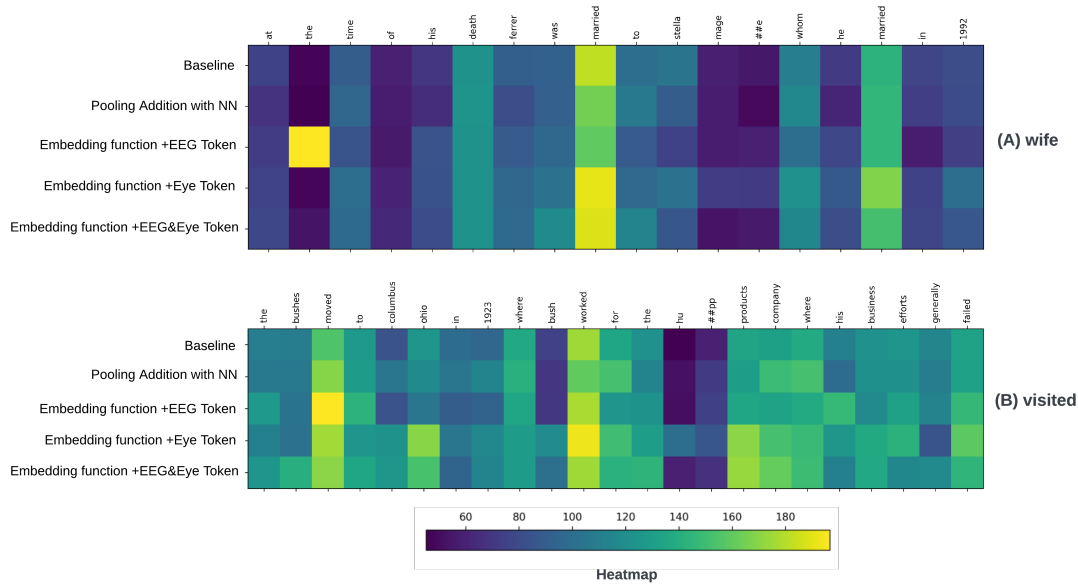

**Figure 11 Heatmap of accumulated attention weights for all models except the Cognition attention mask model:** The figure shows the incoming attention weights heat map for six models for two example sentences with relation types (A) wife and (B) visited. The attention weights were accumulated using the formula described in the analysis section.

The brain activity data for each word does not inform about its relevance as each word that was fixated upon, will have an associated fixation-related potential which individually cannot explain the relevance of a word in a sentence whereas, the Eye token value of each word in a sentence represents the overall relevance of the words for a sentence by combining five eye-tracking features in one value. According to the observations, the model tries to resolve the relevance pattern at the corpus level as it did with the EEG tokens, but with a difference in identifying the pattern. The EEG token model seemingly summed the EEG tokens at the corpus level to assign the attention weights to the highest token sum word that was fixated upon in a sentence. In contrast, to learn the relevance pattern for a word from the Eye tokens, the model seems to calculate the ratio of the number of fixations to the number of occurrences of a word in the corpus. For example, consider the same sentence in figure 11 (A), the EEG token model assigned high attention score to "the" for the reasons explained above whereas the Eye token model assigned the highest attention score to "married". If the EEG token model logic is followed for the Eye token and Eye token values for "the" and "married" are summed in the training corpus, "the" has a value of 1698 which is higher than "married" so the model should have assigned more attention weight to "the" in this scenario as well. Yet, this is not the case and the model assigns a high weight to "married" as out of 32 occurrences of the word in the training corpus, it was fixated upon 29 times by the human subject which is a logical way of informing about the relevance of the word. Whereas, out of 299 occurrences of "the" it was fixated upon 157 times which is about half hence, the model deemed "married" more relevant than "the". The method gives the model a pool of relevant words to assign more attention weights as compared to the ones that occur more in the corpus but have fewer fixations and if two or more words have similar ratios of fixation to the occurrence, the attention weights for those words are fine-tuned during the training process to decide which words will receive the highest values depending on the error calculation. This observation seems to be consistent across other examples.

A combination of both EEG and Eye tokens seems to have alleviated the issue of assigning high selective attention weights depending on the EEG tokens like the word "the". From the observation, it looks like the model is able to pay better attention to the keywords in many examples as compared to the features individually or better distribute the attention weights between words as compared to the vanilla model but further research is required into this model to gather more insights on how the model is combining both features.

Since the addition of sentence-level EEG features to the pooled output using a neural network did not change the architecture of the BERT model, it was difficult to analyze the change in attention heads as the neural network that handles the EEG features is used in combination with a downstream function. The observations show that in very few examples the sentence-level neural network model did pay higher attention to keywords as compared to other models but it is the same case with the vanilla model and it does not provide any substantial insights. This, however, was expected behavior as BERT's architecture was not affected directly.

The attention analysis for the cognitive attention mask is done separately as it answers an interesting question i.e. if BERT can be forced to pay human-like attention. The analysis says yes, at least partially. Consider the examples in figure 12 that compare baseline and cognitive attention mask models against human subject attention. It can be observed that the cognitive mask model ignores all the keywords that the human subject did not pay attention to in an exact match. This is consistent with all the examples (please see appendix C). However, the model failed to assign high attention scores to other keywords which hampered the performance. Nevertheless, it was very interesting to see the model ignore words like the human subject. Further experimentation with the self-attention function with higher quality eye-tracking data with different permutations and combinations of the eye-tracking features would be beneficial to try and get the model to exactly match human attention.

The models augmented with cognitive features seem to have gained more "confidence" while assigning attention weights as seen from the examples above but can the keywords that were assigned the most weights be used to explain the predictions? As per the observations, yes. The method used to accumulate the incoming attention for each word as described above seems like an effective way to explain model predictions and it also correlates well with LIME results with the exception of the model augmented with the cognitive attention mask.

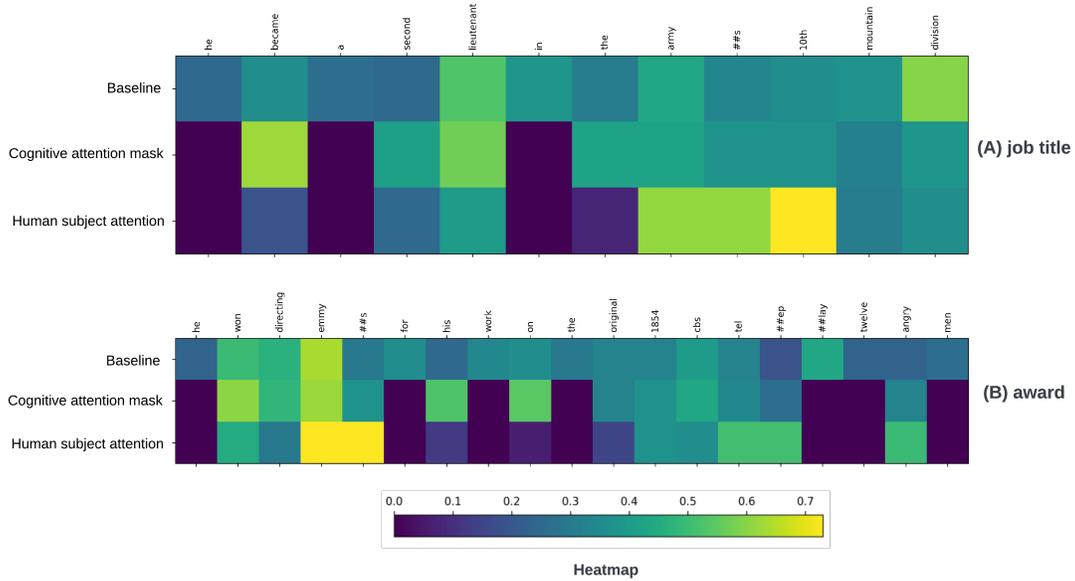

**Figure 12 Heatmap of accumulated attention weights for the Cognition attention mask model:** The figure shows the incoming attention weights heat map for the baseline model, the cognitive attention mask model, and the true human subject attention for two sentences having relation types (A) job title and (B) award. The cognitive attention mask model ignores words that were not attended to by the human subject represented in dark blue or 0 value.

Table 4 shows the predicted relation type and top five keywords selected by LIME and the attention weights for four test sentences with four relation types. There were inconsistencies in the correlation of the cognitive attention mask model with the LIME scores in many examples. The examples shown in table 4 are the ones where the keywords relate between both attention weights and LIME (please find the full result spreadsheet for the cognitive attention mask model in the supplementary files). As explained in the results section, a sentence can have multiple keywords relating to different class types so it is important that whatever class is chosen by the model, the selected keywords should intuitively match those classes.

| Prediction | Model | Self-Attention Keywords | Lime Keywords |
|---|---|---|---|
| **founder** | Vanilla | founded', 'magazine', 'stories', 'a', 'publishes' | founded', 'magazine', 'in', 'that', 'a' |
| | +EEG Token | founded', 'magazine', 'stories', 'publishes', 'a' | founded', 'in', 'that', '1997', 'magazine' |
| | +Eye Token | founded', 'magazine', 'publishes', 'stories', 'literary' | founded', '1997', 'in', 'that', 'a' |
| | +EEG&Eye Token | founded', 'magazine', 'a', 'publishes', 'stories' | founded', '1997', 'in', 'magazine', 'that' |
| | Addition with NN | founded', 'magazine', 'publishes', 'a', 'stories' | magazine', 'stories', 'literary', 'founded', 'that' |
| | Cognitive attention mask | founded', 'publishes', '1997', 'that', 'literary' | publishes', 'literary', '##y', 'magazine', 'zoe' |
| **wife** | Vanilla | met', 'wife', 'sweetheart', 'grew', 'lynne' | wife', 'his', 'and', 'met', 'future' |
| | +EEG Token | met', 'wife', 'sweetheart', 'grew', 'his' | wife', 'his', 'and', 'grew', 'up' |
| | +Eye Token | met', 'wife', 'sweetheart', 'grew', 'wyoming' | wife', 'future', 'and', 'vincent', 'his' |
| | +EEG&Eye Token | met', 'wife', 'sweetheart', 'grew', 'future' | wife', 'his', 'met', 'and', 'and' |
| | Addition with NN | met', 'wife', 'sweetheart', 'grew', 'future' | wife', 'vincent', 'his', 'future', 'and' |
| | Cognitive attention mask | 'grew', 'wife', 'sweetheart', 'and', 'age' | 'wife', 'his', 'vincent', 'lynne', 'met' |
| **education** | Vanilla | 'scholarship', 'graduated', 'harvard', 'from', 'college' | 'from', 'university', 'graduated', 'of', 'college' |
| | +EEG Token | 'graduated', 'scholarship', 'from', 'college', 'to' | 'university', 'school', 'from', 'graduated', 'college' |
| | +Eye Token | 'graduated', 'scholarship', 'college', 'from', 'harvard' | 'university', 'graduated', 'school', 'college', 'scholarship' |
| | +EEG&Eye Token | 'graduated', 'scholarship', 'harvard', 'university', 'college' | 'school', 'university', 'graduated', 'from', 'college' |
| | Addition with NN | 'graduated', 'scholarship', 'from', 'harvard', 'to' | 'school', 'from', 'law', 'harvard', 'graduated' |
| | Cognitive attention mask | 'graduated', 'received', 'scholarship', 'harvard', 'from' | 'university', 'school', 'college', 'graduated', 'scholarship' |
| **political affiliation** | Vanilla | 'republican', 'governor', 'arkansas', 'was', 'state' | 'republican', 'was', 'from', 'the', 'state' |
| | +EEG Token | 'the', 'republican', 'of', 'of', 'was' | 'republican', 'was', 'from', 'arkansas', 'the' |
| | +Eye Token | 'republican', 'governor', 'was', 'state', 'arkansas' | 'republican', 'us', 'state', 'arkansas', 'was' |
| | +EEG&Eye Token | 'republican', 'governor', 'state', 'arkansas', 'was' | 'republican', 'arkansas', 'state', 'was', 'us' |
| | Addition with NN | 'republican', 'governor', 'was', 'state', 'us' | 'republican', 'was', 'the', 'from', 'arkansas' |
| | Cognitive attention mask | 'governor', 'was', 'state', '2003', 'the' | 'republican', 'arkansas', 'state', 'was', '2003' |

**Table 4 Correlation between incoming attention weights and LIME scores:** The table shows the keywords selected by the highest incoming attention weights and LIME for 4 sentences. The first column is the predicted label by the models, the second column shows the keywords with the highest incoming self-attention values, and the third column contains the keywords selected by LIME. Top 5 keywords are shown for each example.

The first column shows block-wise predicted relation types and columns 2, 3, and 4 show the models, the keywords with the highest attention weights, and the keywords selected by LIME. This observation suggests that high attention weights were assigned to intuitive keywords by the self-attention function. Also, the LIME outputs are generally similar to the high attention weights for the intuitive keywords as can be seen from the table. For example, the word "founded" describes the relation type "founder" which both the methods captured correctly, similarly "republican" and "governor" intuitively describe "political affiliation" class and this trend is consistent for the other classes in the test sent. It would be difficult to say which of these is a better method but using the attention accumulation method has the advantage of providing scope to understand the model at a deeper level and potential architectural modifications according to the attention outputs.

# Chapter 6: Discussion

All the findings presented in this research suggest that cognitive features such as eye-tracking data, word-level, and sentence-level EEG data are valuable to NLP applications. This study provides an in-depth exploration of state-of-the-art transformer-based language models through architectural modulation and explainability as many complex tasks such as text classification can have multiple correct answers which makes it difficult to rely only on the evaluation metrics. The model might be correct in predicting other relation types present in the sentence however, the ground truth won't match with the prediction and the evaluation metric scores will decrease which is why explainability is a very important factor in judging the model output to make sure that whatever the model predicts should be intuitively correct.

To summarize what has been presented so far, cognitive features from the Zurich Cognitive Corpus (ZuCo) were used to perform novel augmentations on BERT at both architectural and downstream levels by keeping the training time on par with the vanilla BERT model in most of the cases and increasing the evaluation and explainability scores. In contrast, previous studies have shown the benefits of using BERT embeddings with an elaborate downstream training model using CNNs and bi-LSTMs which increases the model complexity and presumably, the training time. The architectural modifications with EEG tokens, Eye-tokens, and cognitive attention masks also provided a reference to analyze changes in the attention weights to generate insights on how the model might be learning the features during training. Three experiments were performed by modifying the embedding function of BERT with EEG tokes, Eye tokens, and both together. The model augmented with the EEG tokens improved the evaluation scores by 1.5% on average and reached the highest accuracy of 70% in 10 runs compared to 67% of the baseline model. The model with the Eye Tokens performed similarly to the vanilla model but showed consistent improvements in assigning high attention weights to keywords across the test set. Using both EEG and Eye tokens provided stability to the model in assigning high attention weights to the keywords with a slight average increase of 0.7% in the evaluation metrics as compared to the vanilla model. Overall, the combination of EEG and Eye tokens in the embedding function showed more consistent improvements in assigning attention weights to keywords as compared to other models. One experiment was performed on the self-attention function by augmenting it with the cognitive attention mask feature which compels the model to diminish the attention scores for the keywords ignored by the human subject. Both the evaluation metrics and accumulated attention weights showed substandard results as compared to the baseline and the other models. However, this was a very interesting experiment as the model successfully ignored the words like the human subject and though there were inconsistencies between keywords attention weights and LIME scores for this model, the best model out of the 5 runs with a 64% accuracy, was generally able to select the correct keywords as it can be seen from table 4 just not with a lot of "confidence", i.e. the sum of the incoming attention weights for the keywords was low. The final experiments were performed by incorporating the sentence-level EEG features in the Bert for sequence classification which is a downstream function that uses the pooled output from BERT as input to a classifier. Three methods were tested to integrate the EEG features with the pooled output before passing the result to the classifier. The first was divided into two sections, concatenating the 105-dimensional EEG features directly to the pooled output and passing the EEG features into a three-layered neural network with GELU activation and concatenating the output from the last layer of the neural network to the pooled output where both failed to give good results and achieved an accuracy of 24.7% and 24.3% respectively. The second method was multiplying and averaging each value of the 768-dimensional pooled output with the 105-dimensional EEG vector which also failed to perform well with an accuracy of 23.12% and increased the training time by a huge margin as compared to the baseline model. The final method performed the best which was adding the output from the last layer of the three-layered neural network to the pooled output before passing it to the classifier. The metrics on average increased by 2-3% and the best model achieved 70% accuracy out of the 10 runs but the "confidence" in assigning attention weights to the intuitive keywords did not increase as the architecture of the BERT model was not altered similar to the vanilla model.

To be fair to the models, the task was a bit easier for the human subjects as they were shown the relation type to look for before the sentences were presented so they knew what to search for whereas one sentence can have keywords relating to other classes which can greatly reduce the evaluation metric scores. It would be interesting to experiment further and try to mimic the exact experiment setup as the subjects had.

# Chapter 7: Future work

There are many interesting directions in which this research can be continued. One possible way which is a bit unconventional is to treat text classification as a machine translation task. BERT is an encoder that uses self-attention to learn how each word relates to every other word in a sentence but an encoder-decoder model like a transformer also learns how one sequence from the encoder, relates to other sequences from the decoder (Vaswani et al., 2017). This means there is cross-attention between the words in the encoder and the decoder. In other words, in the transformer architecture, each word from the encoder relates to every word in the decoder. For example, if a sentence from English "This is my coat" is to be translated into French "c'est mon manteau", the word "This" has to relate to every word in the French sequence through cross-attention to learn "c'est" is its translation. This, of course, depends on providing the model with multiple examples of "this" and "c'est" to derive the commonality between the words and assign high cross-attention weights between the two words. Machine translation has vastly improved using this paradigm and it quickly became the state-of-the-art model for the task (Wang et al., 2021). What if text classification is treated similarly? Consider two sentences, "She was awarded a Nobel prize" and "She was awarded the Fields medal" the goal would be to translate these two sentences into just one keyword which would be the relation type "award". Since both the sentences are required to "translate" into the same word, the model needs to find multiple common keywords from the encoder which are the two sequences to decode "award" as the output, and presumably the word "awarded" would be assigned the highest cross attention weight between encoder and the decoder since it is the only common keyword between the two sentences. This hypothesis was tested with a crud implementation of an encoder-decoder transformer with 1 layer and 4 attention heads coded from scratch. Figure 13 shows a few examples where the model tries to find common words in the training sequences to assign or "translate" them into a relation type.

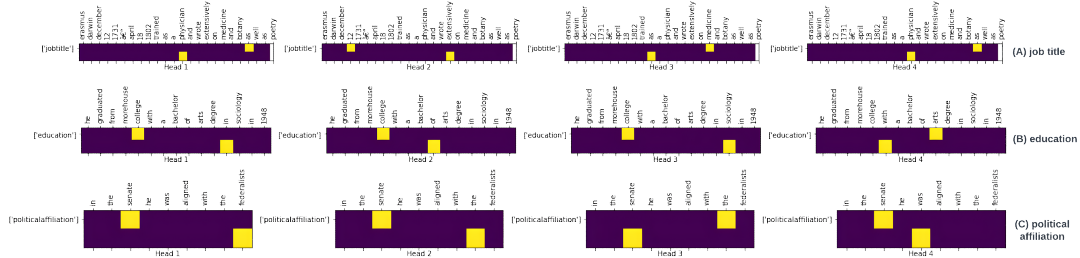

**Figure 13 Example of text classification as translation using an encoder-decoder transformer model:** The figure shows three example sentences in relation types (A) job title, (B) education, and (C) political affiliation. The yellow cells indicate high cross-attention between the encoder and the decoder. The predicted label is shown on the left of each head in the square brackets.

A point to note is that this implementation cannot be compared to any benchmark results and it is at the hypothesis testing state. This was done to check if there is any potential in following this approach. That being said, the few examples show that the model could assign high cross-attention weights to intuitive keywords. The main advantage of this setup is that the cross attention module in the middle layer of the decoder will provide a clear explanation of how the words in the sentence relate to the class label which is not possible using only the encoder i.e. BERT where the self-attention weights had to be accumulated. Since the model will need to find all the possible common keywords to relate them to a particular class, it will need lots of training examples to learn which words to assign high attention weights to. This is where cognitive features can be integrated with cross-attention. By informing the model what words are relevant, I speculate that it would need fewer training samples to learn the relation types which requires further research. Another direction is to try to train generative models to create synthetic EEG and eye-tracking data from the available datasets. Though EEG and eye-tracking hardware have become more accessible in recent years, they can't be used by everyone. The text saliency model (TSM) proposed by Sood et al. (2020) discussed in the introduction section can be used to generate synthetic gaze features which can be used to incorporate with state-of-the-art NLP models as they show that their gaze prediction model correlates well with true human gaze. This is an example of how synthetic cognitive features could improve NLP models.

# Chapter 8: Conclusion

This research presented novel experiments in integrating cognitive features like EEG and eye-tracking data at different levels of the architecture of BERT, a language encoder that uses a self-attention mechanism. The cognitive features were recorded during natural reading while performing a relation extraction task. In addition, model explainability was also studied using two methods, a model agnostic explanation framework called LIME and an approach of accumulating incoming attention weights for each word in a sentence which correlated well with the keywords selected by LIME. The latter can be used as a potential way of explaining classification tasks that use BERT. Augmenting the model with sentence-level EEG data, word-level EEG tokens, and word-level EEG and Eye tokens together, increased the average evaluation performance as compared to the baseline. Lastly, according to the observations, the cognitive models except for sentence-level EEG and cognitive attention mask, assigned more attention weights and LIME scores to the keywords that define relation types as compared to the vanilla BERT model. Further research in the experiments presented could be beneficial to see greater improvements. This study hopes to provide a foundation for exploring cognitive features in many different forms with NLP models. While the models might lack an actual understanding of language (Bishop, 2021; Dubova 2022) which has been the topic of an ongoing debate recently, experiments like the ones presented and other related works could potentially make the artificial models correlate better with human comprehension of language as they will be learning from the domain experts with the help of cognitive data.

# References

Aron Culotta, Andrew McCallum, and Jonathan Betz. (2006). Integrating ProbabilisticExtraction Models and Data Mining to Discover Relations and Patterns in Text.In Proceedings of the Human Language Technology Conference of the NAACL, MainConference, pages 296–303, New York City, USA. Association for ComputationalLinguistics.

Barrett, M.,Bingel, J., Keller, F., & Søgaard, A. (2016). Weakly SupervisedPart-of-speech Tagging Using Eye-tracking Data. Proceedings Of The 54ThAnnual Meeting Of The Association For Computational Linguistics (Volume 2:Short Papers). doi: 10.18653/v1/p16-2094

Bastiaansen,M., van Berkum, J., & Hagoort, P. (2002). Event-related theta powerincreases in the human EEG during online sentence processing. NeuroscienceLetters, 323(1), 13-16. doi: 10.1016/s0304-3940(01)02535-6

Benyou Wang,Lifeng Shang, Christina Lioma, Xin Jiang, Hao Yang, Qun Liu, & Jakob GrueSimonsen (2021). On Position Embeddings in BERT. In International Conference onLearning Representations.

Bishop JM(2021) Artificial Intelligence Is Stupid and Causal Reasoning Will Not FixIt. Front. Psychol. 11:513474. doi: 10.3389/fpsyg.2020.513474

Bojanowski,P., Grave, E., Joulin, A., & Mikolov, T. (2017). Enriching Word Vectorswith Subword Information. Transactions Of The Association ForComputational Linguistics, 5, 135-146. doi: 10.1162/tacl_a_00051

David M.Blei, Andrew Y. Ng, and Michael I. Jordan. 2003. Latent dirichlet allocation.J. Mach. Learn. Res. 3, null (3/1/2003), 993–1022

Devlin, J.,Chang, M., Lee, K., & Toutanova, K. (2019). BERT: Pre-training of DeepBidirectional Transformers for Language Understanding. Retrieved 28 August2022, from http://arxiv.org/abs/1810.04805

Do, T. H.,Nguyen, D. M., Tsiligianni, E., Cornelis, B., and Deligiannis, N. (2017). Multiviewdeep

learning for predicting Twitter users' location. arXiv preprint arXiv:1712.08091.

Dubova, M. (2022).Building human-like communicative intelligence: A grounded perspective. CognitiveSystems Research, 72, 63-79. doi: 10.1016/j.cogsys.2021.12.002

Dudschig, C.(2022). Language and non-linguistic cognition: Shared mechanisms and principlesreflected in the N400. Biological Psychology, 169,108282. doi: 10.1016/j.biopsycho.2022.108282

Foltz, P.(1996). Latent semantic analysis for text-based research. BehaviorResearch Methods, Instruments, &Amp; Computers, 28(2), 197-202.doi: 10.3758/bf03204765

Foulsham, T.(2014). Eye movements and their functions in everyday tasks. Eye, 29(2),196-199. doi: 10.1038/eye.2014.275.

Grabner, R., Brunner,C., Leeb, R., Neuper, C., & Pfurtscheller, G. (2007). Event-related EEGtheta and alpha band oscillatory responses during language translation. BrainResearch Bulletin, 72(1), 57-65. doi:10.1016/j.brainresbull.2007.01.001

Heilbron, M.,Armeni, K., Schoffelen, J., Hagoort, P., & de Lange, F. (2022). A hierarchyof linguistic predictions during natural language comprehension. ProceedingsOf The National Academy Of Sciences, 119(32). doi:10.1073/pnas.2201968119

Hogenboom, F. P.,Frasincar, F., & Kaymak, U. (2010). An overview of approaches toextract information from natural language corpora. In Proceedings ofthe 10th Dutch-Belgian Information Retrieval Workshop (DIR 2010), January 25,2010, Nijmegen, the Netherlands (pp. 69-70). Radboud UniversiteitNijmegen

Holcomb P. J.(1993). Semantic priming and stimulus degradation: implications for the role ofthe N400 in language processing. Psychophysiology, 30(1),47–61. https://doi.org/10.1111/j.1469-8986.1993.tb03204.x

Hollenstein,N., Barrett, M., Troendle, M., Bigiolli, F., Langer, N., & Zhang, Ce.(2019). Advancing NLP with Cognitive Language Processing Signals.

Hollenstein N,Renggli C, Glaus B, Barrett M, Troendle M, Langer N and Zhang C (2021) DecodingEEG Brain Activity for Multi-Modal Natural Language Processing. Front.Hum.

Neurosci. 15:659410. doi: 10.3389/fnhum.2021.659410

Hollenstein,N., Rotsztejn, J., Troendle, M., Pedroni, A., Zhang, C., & Langer, N.(2018). ZuCo, a
simultaneous EEG and eye-tracking resource for natural sentencereading. Scientific
Data, 5(1). doi:10.1038/sdata.2018.291

Klimesch, W.(2012). Alpha-band oscillations, attention, and controlled access to
storedinformation. Trends in Cognitive Sciences, 16, 606 - 617

Ling, S., Lee,A., Armstrong, B., & Nestor, A. (2019). How are visual words represented?Insights
from EEG-based visual word decoding, feature derivation and imagereconstruction.
Human Brain Mapping, 40(17), 5056-5068.doi: 10.1002/hbm.24757

Mikolov, Tomas &Chen, Kai & Corrado, G.s & Dean, Jeffrey. (2013). Efficient Estimationof Word
Representations in Vector Space. Proceedings of Workshop at ICLR. 2013

Pennington,J., Socher, R., & Manning, C. (2014). Glove: Global Vectors for WordRepresentation.
Proceedings Of The 2014 Conference On Empirical MethodsIn Natural Language
Processing (EMNLP). doi: 10.3115/v1/d14-1162

Prystauka, Y.,& Lewis, A. (2019). The power of neural oscillations to inform
sentencecomprehension: A linguistic perspective. Language And LinguisticsCompass,
13(9). doi: 10.1111/lnc3.12347

Ramachandram,D., & Taylor, G.W. (2017). Deep Multimodal Learning: A Survey on
RecentAdvances and Trends. IEEE Signal Processing Magazine, 34, 96-108.

Reynolds, D.(2009). Gaussian Mixture Models. Encyclopedia Of Biometrics,659-663. doi:
10.1007/978-0-387-73003-5_196

Ribeiro, M., Singh,S., & Guestrin, C. (2016). "Why Should I Trust You?": Explainingthe Predictions
of Any Classifier. Proceedings of the 22nd ACM SIGKDDInternational Conference on
Knowledge Discovery and Data Mining.

Sereno, S.(2003). Measuring word recognition in reading: eye movements and event-
relatedpotentials. Trends In Cognitive Sciences, 7(11),489-493. doi:
10.1016/j.tics.2003.09.010

Sherstinsky,A. (2020). Fundamentals of Recurrent Neural Network (RNN) and Long Short-
TermMemory (LSTM) network. Physica D: Nonlinear Phenomena, 404,132306. doi:
10.1016/j.physd.2019.132306

Sood, E.,Tannert, S., Mueller, P., & Bulling, A. (2020). Improving Natural LanguageProcessing
Tasks with Human Gaze-Guided Neural Attention.

Szczepanski,M., Pawlicki, M., Kozik, R., & Choras, M. (2021). New explainability methodfor
BERT-based model in fake news detection. Scientific Reports, 11(1).doi:
10.1038/s41598-021-03100-6

Szegedy, C.,Liu, W., Jia, Y., Sermanet, P., Reed, S., Anguelov, D., et al. (2015). "Goingdeeper with
convolutions," in Proceedings of the IEEE Conference on ComputerVision and Pattern
Recognition, 1–9. doi: 10.1109/CVPR.2015.7298594

Uther, W.,Mladenic, D., Ciaramita, M., Berendt, B., Kolcz, A., & Grobelnik, M. et al.(2011). TF–IDF.
Encyclopedia Of Machine Learning, 986-987. doi:10.1007/978-0-387-30164-8_832

Vaswani,A., Shazeer, N.M., Parmar, N., Uszkoreit, J., Jones, L., Gomez, A.N., Kaiser,L., & Polosukhin,
I. (2017). Attention is All you Need. ArXiv,abs/1706.03762.

Wang, Haifeng& Wu, Hua & He, Zhongjun & Huang, Liang & Church, Kenneth.(2021). Progress in
Machine Translation. Engineering.10.1016/j.eng.2021.03.023.

Weiss, S., &Mueller, H. (2012). "Too Many betas do not Spoil the Broth": The Role of BetaBrain
Oscillations in Language Processing. Frontiers In Psychology, 3.doi:
10.3389/fpsyg.2012.00201

Williams, C.,Kappen, M., Hassall, C., Wright, B., & Krigolson, O. (2019). Thinking thetaand alpha:
Mechanisms of intuitive and analytical reasoning. Neuroimage, 189,574-580. doi:
10.1016/j.neuroimage.2019.01.048

Winkler, I., Haufe,S., & Tangermann, M. (2011). Automatic Classification of ArtifactualICA-
Components for Artifact Removal in EEG Signals. Behavioral AndBrain Functions, 7(1),
30. doi: 10.1186/1744-9081-7-30.

Xiao, R.,Shida-Tokeshi, J., Vanderbilt, D., & Smith, B. (2018). Electroencephalographypower and coherence changes with age and motor skill development across thefirst half year of life.

PLOS ONE, 13(1), e0190276.doi: 10.1371/journal.pone.0190276

Zeng, H.,& Song, A. (2014). Removal of EOG Artifacts from EEG Recordings UsingStationary

Subspace Analysis. The Scientific World Journal, 2014,1-9. doi: 10.1155/2014/259121

# Appendix A: LIME Scores heat maps for the test corpus

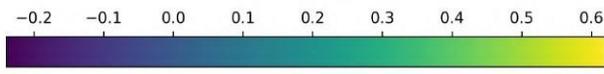

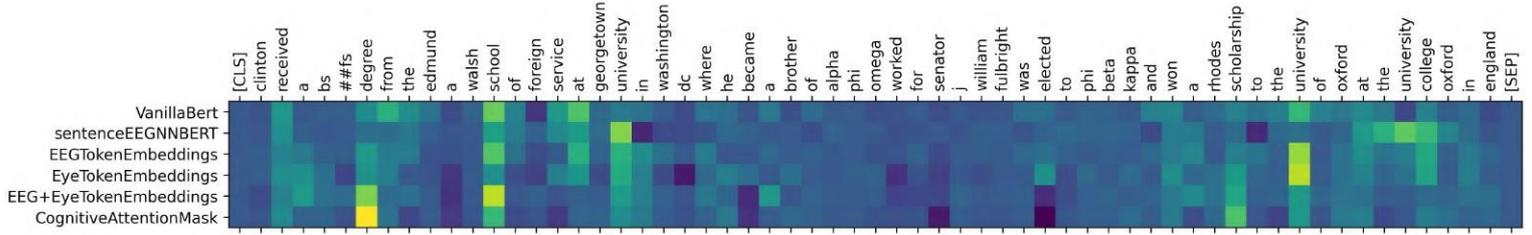

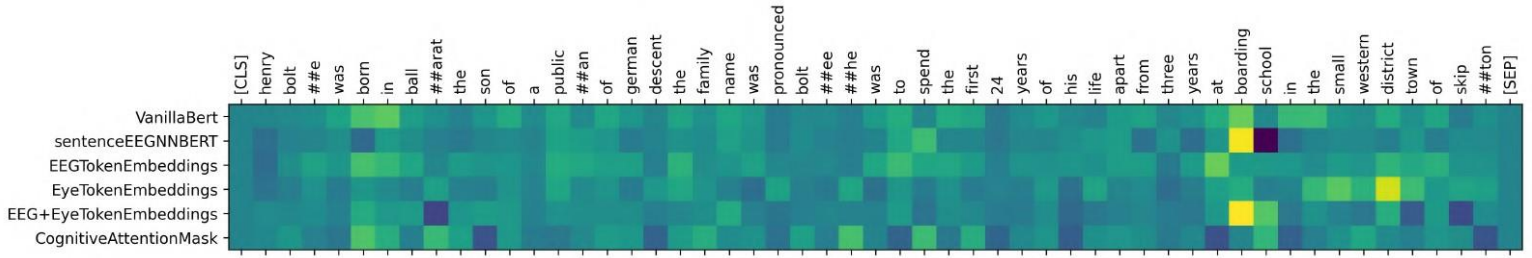

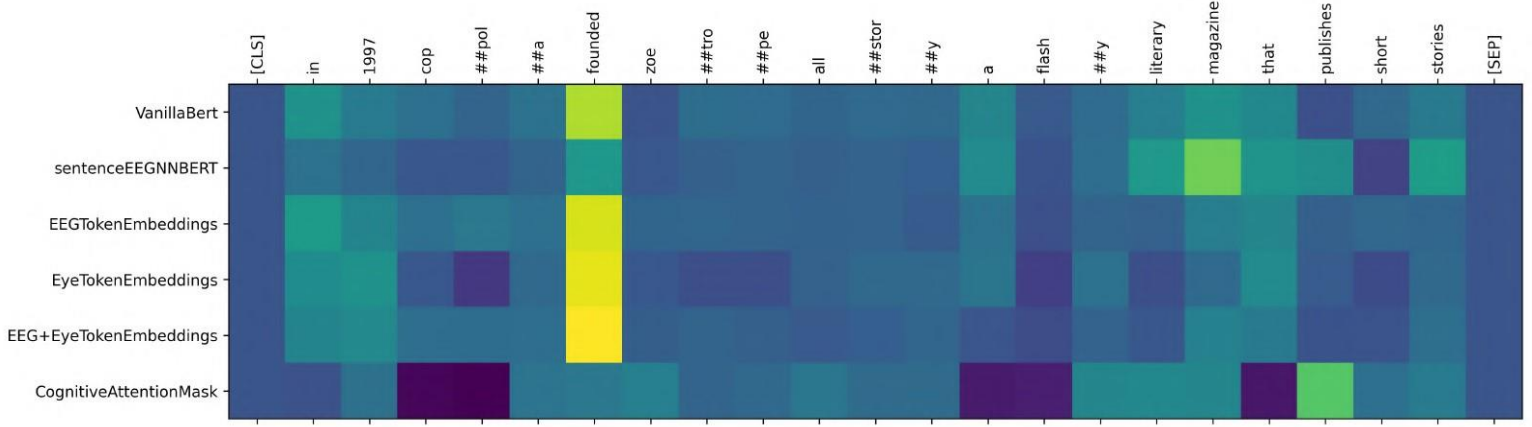

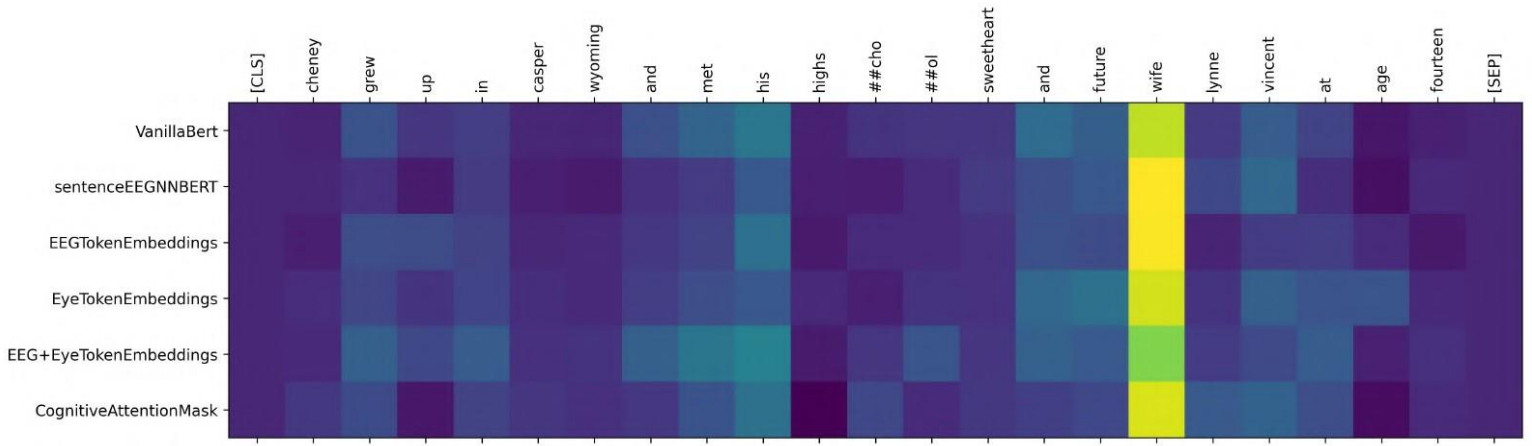

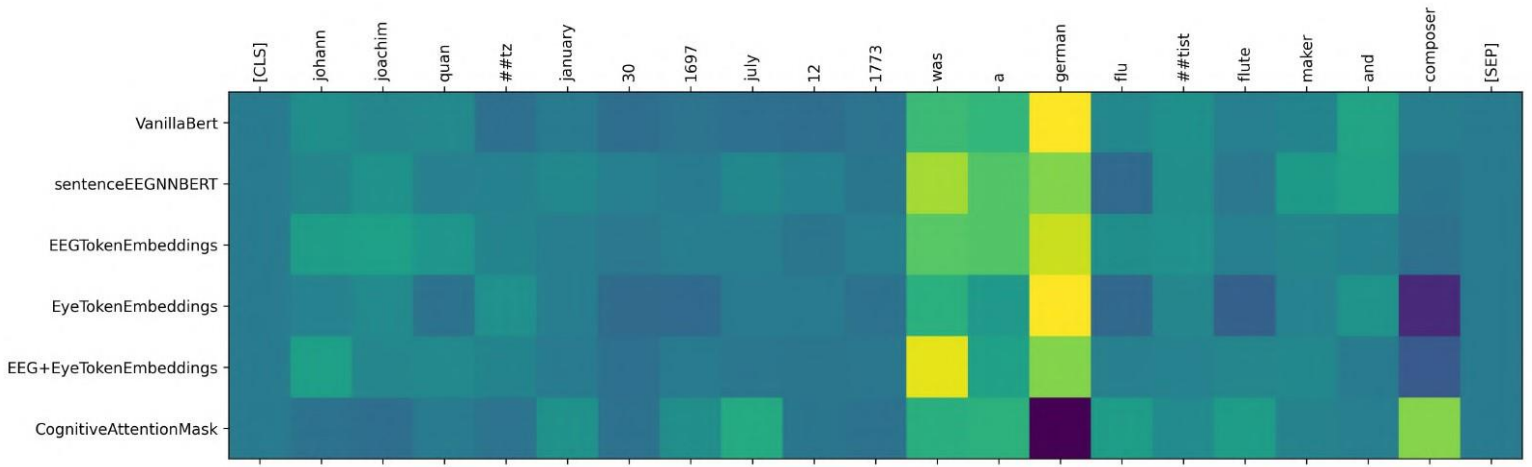

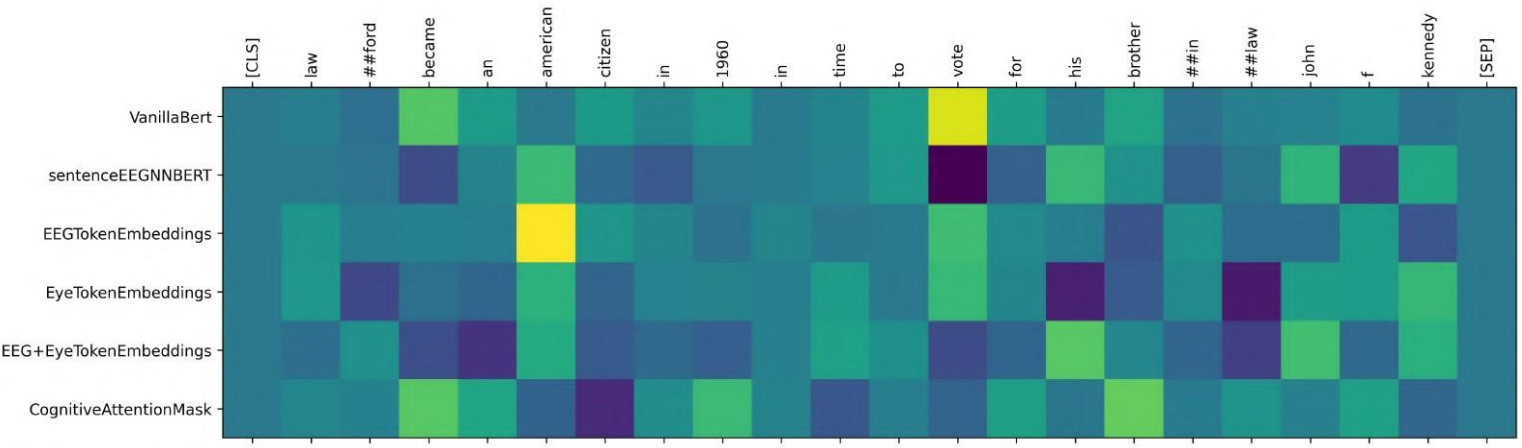

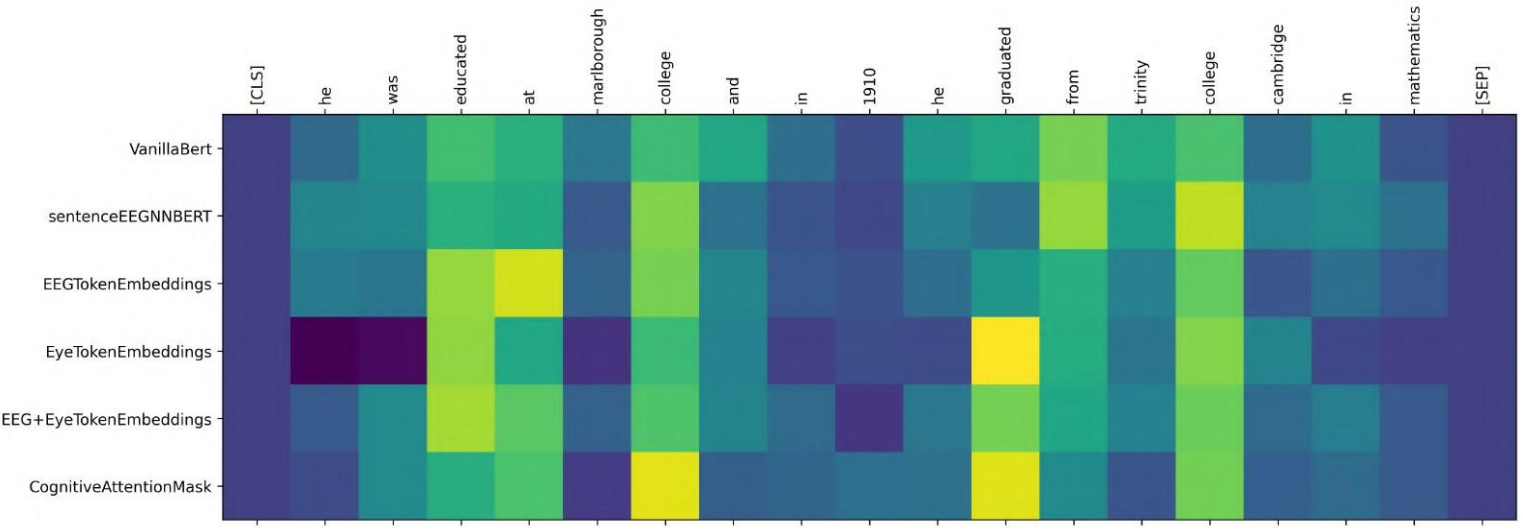

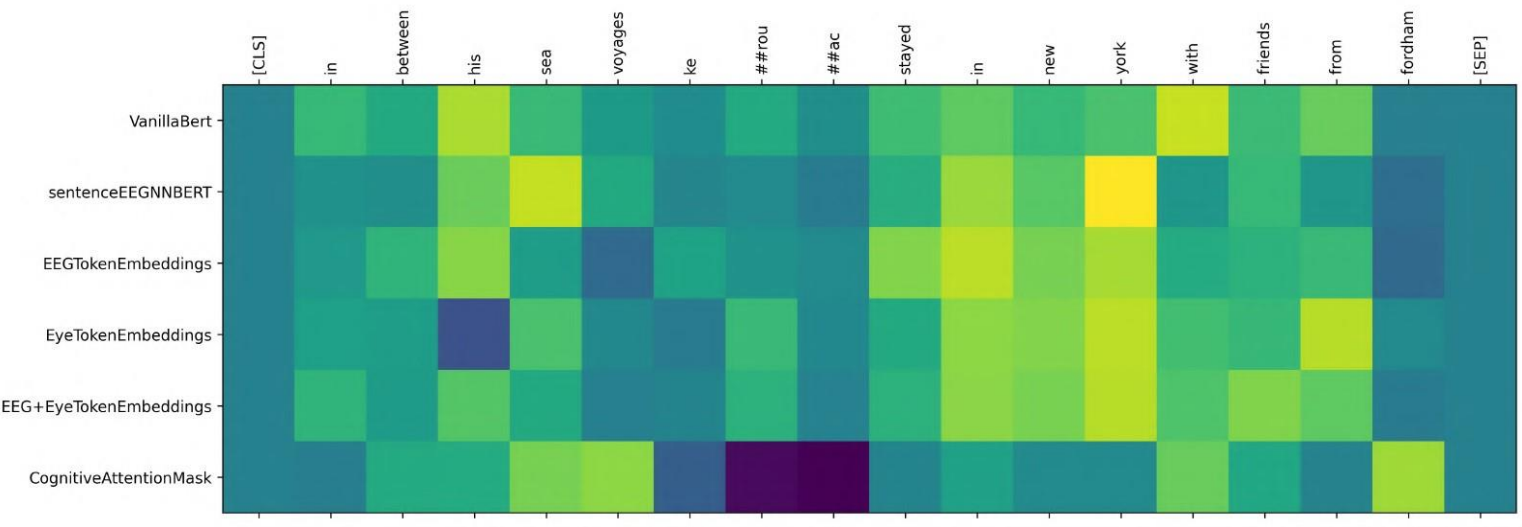

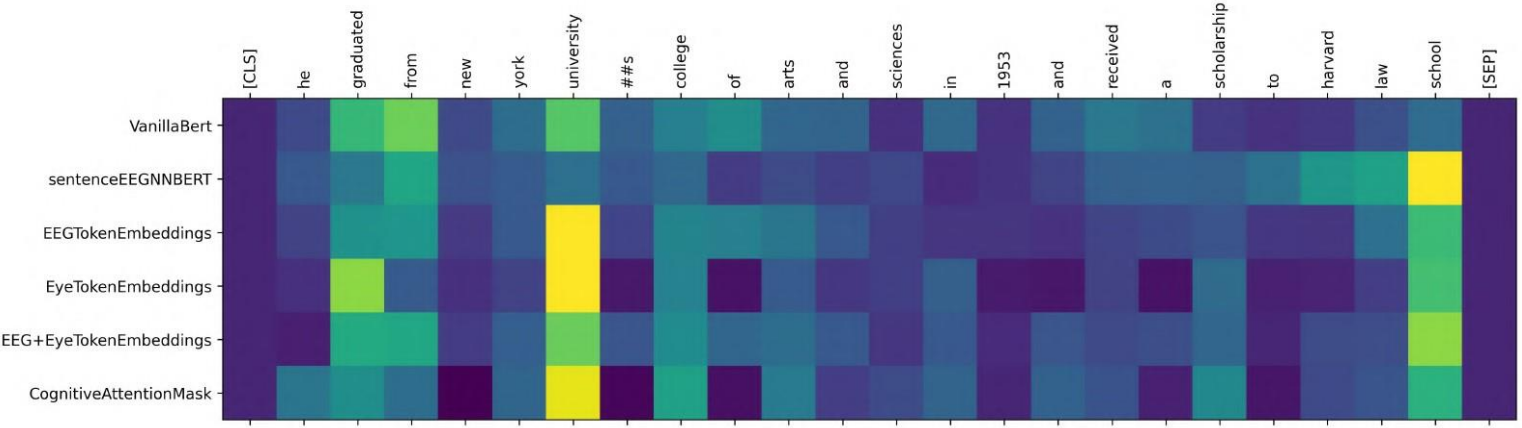

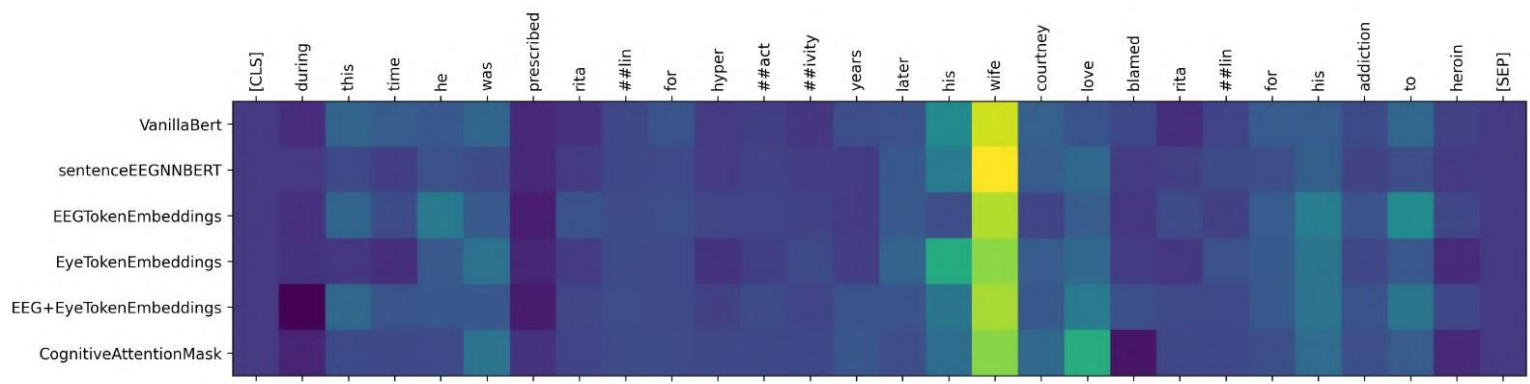

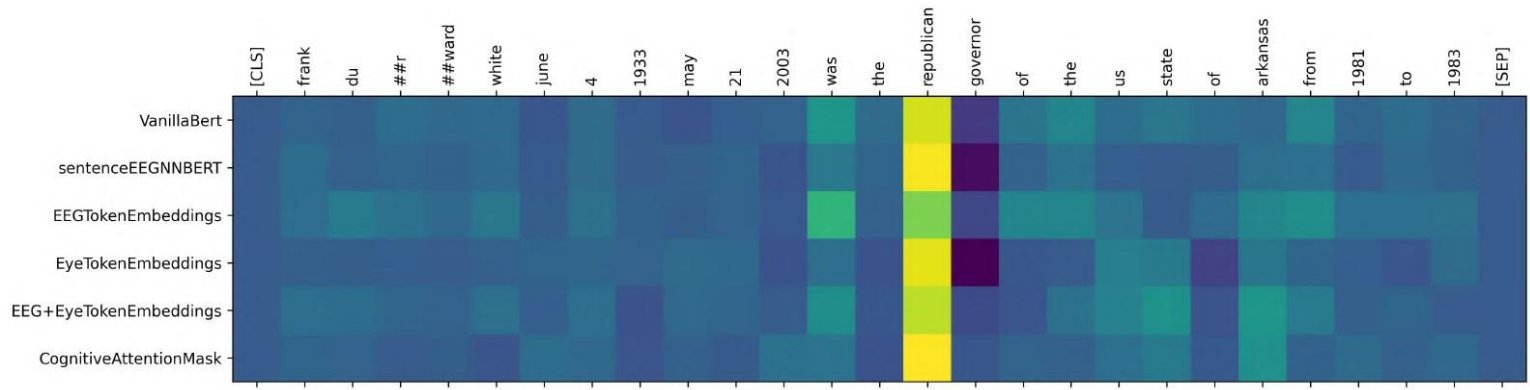

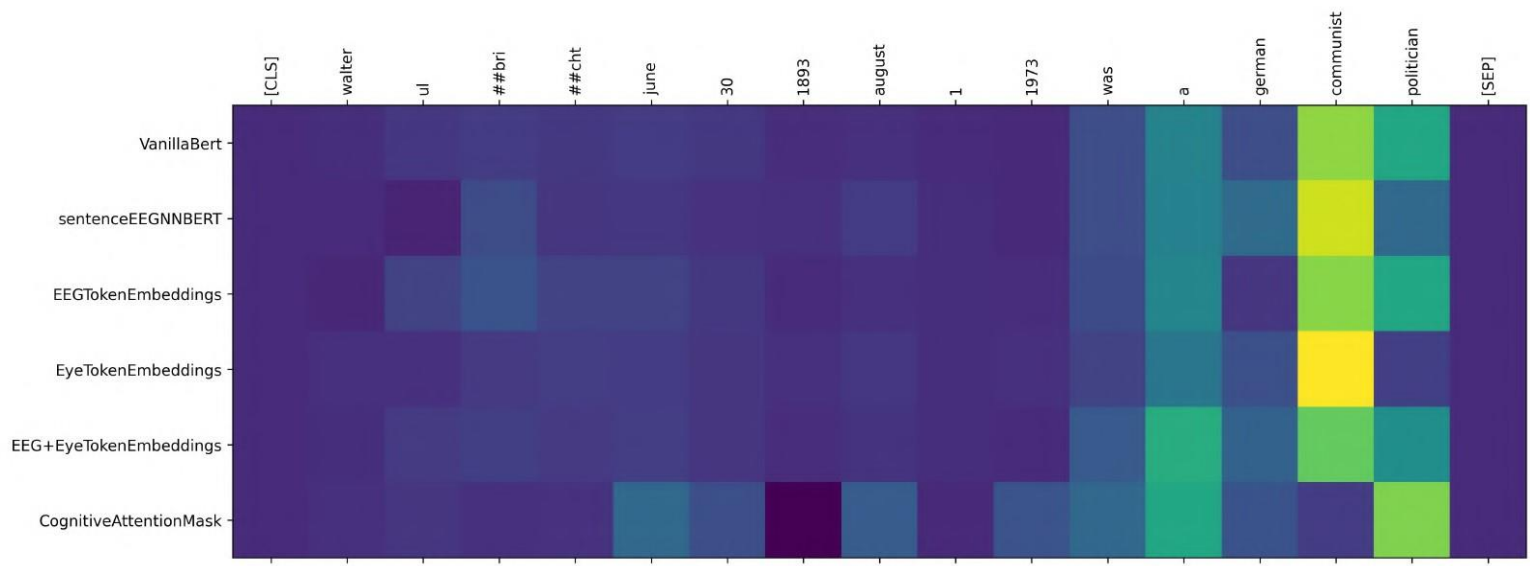

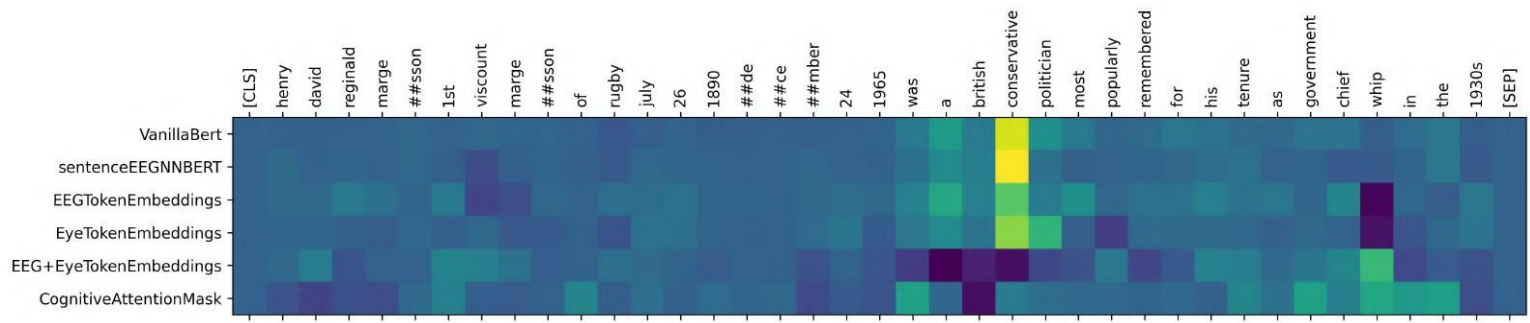

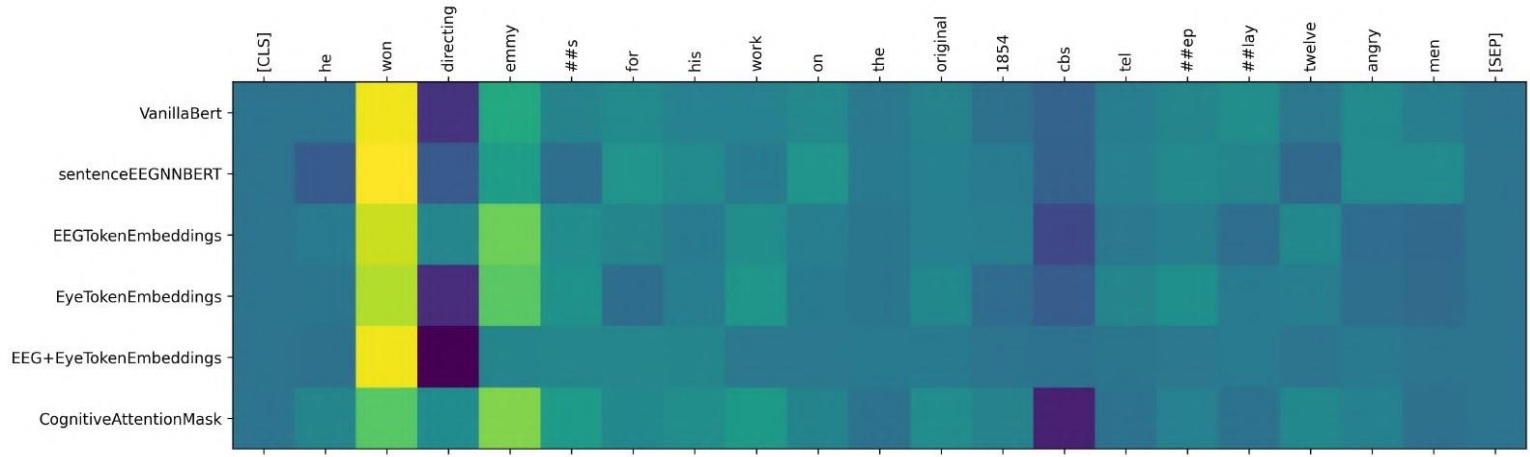

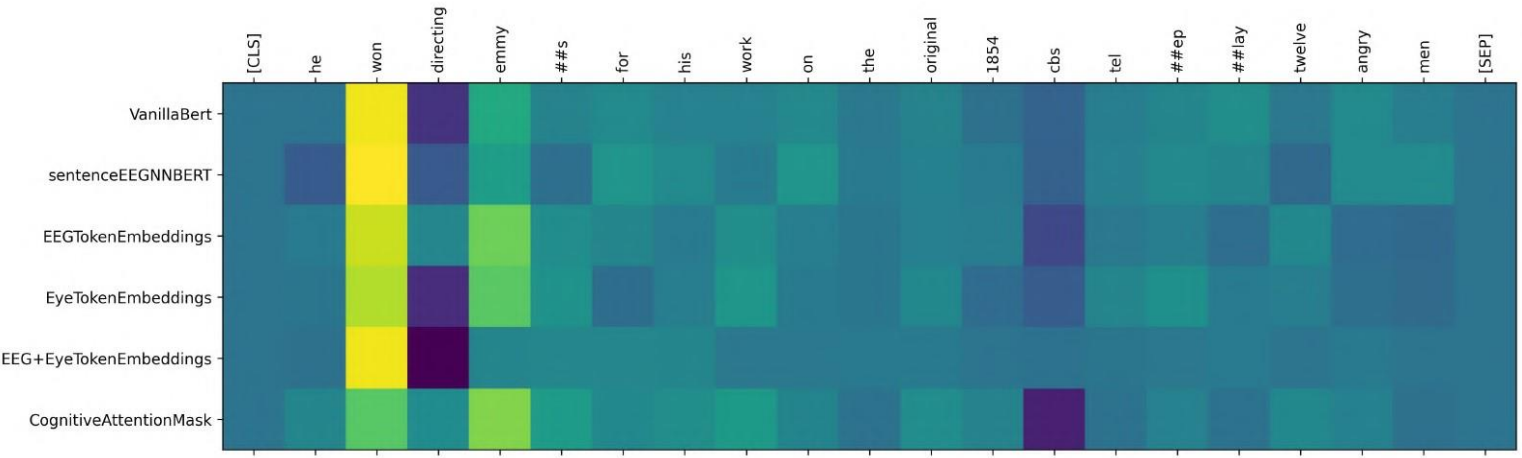

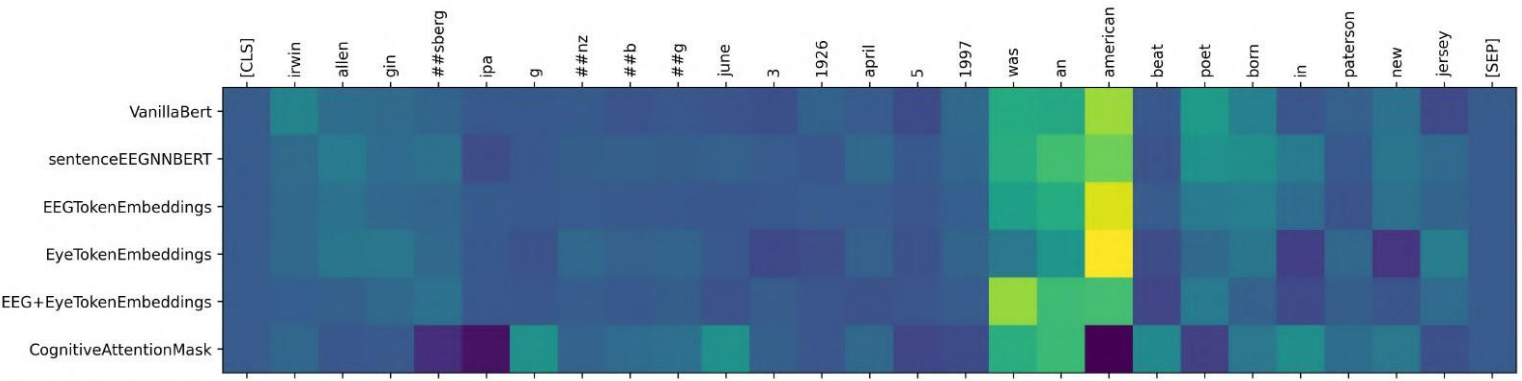

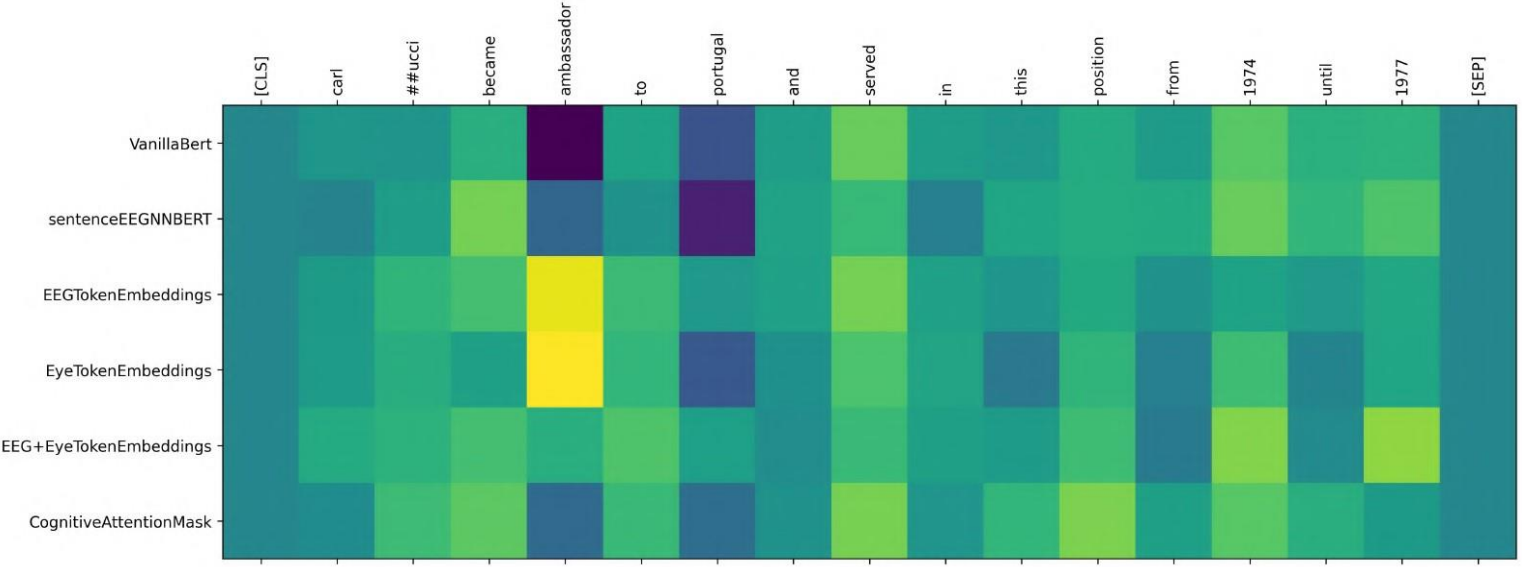

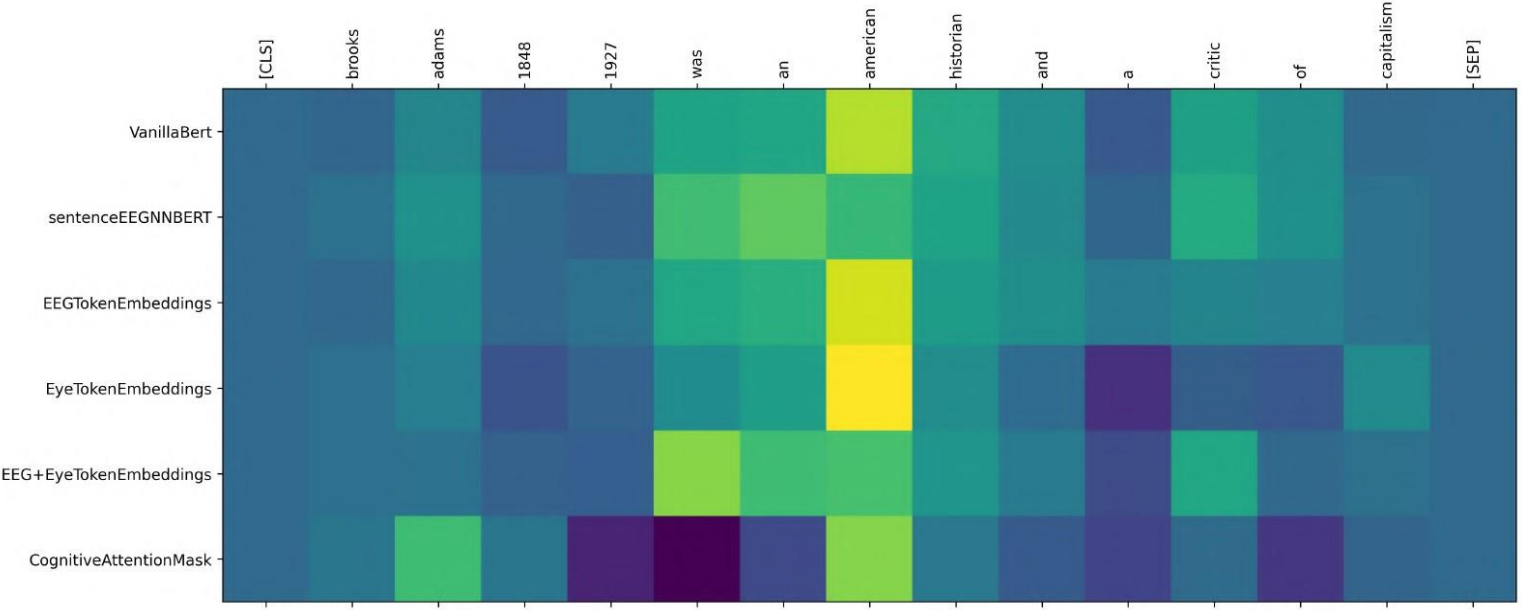

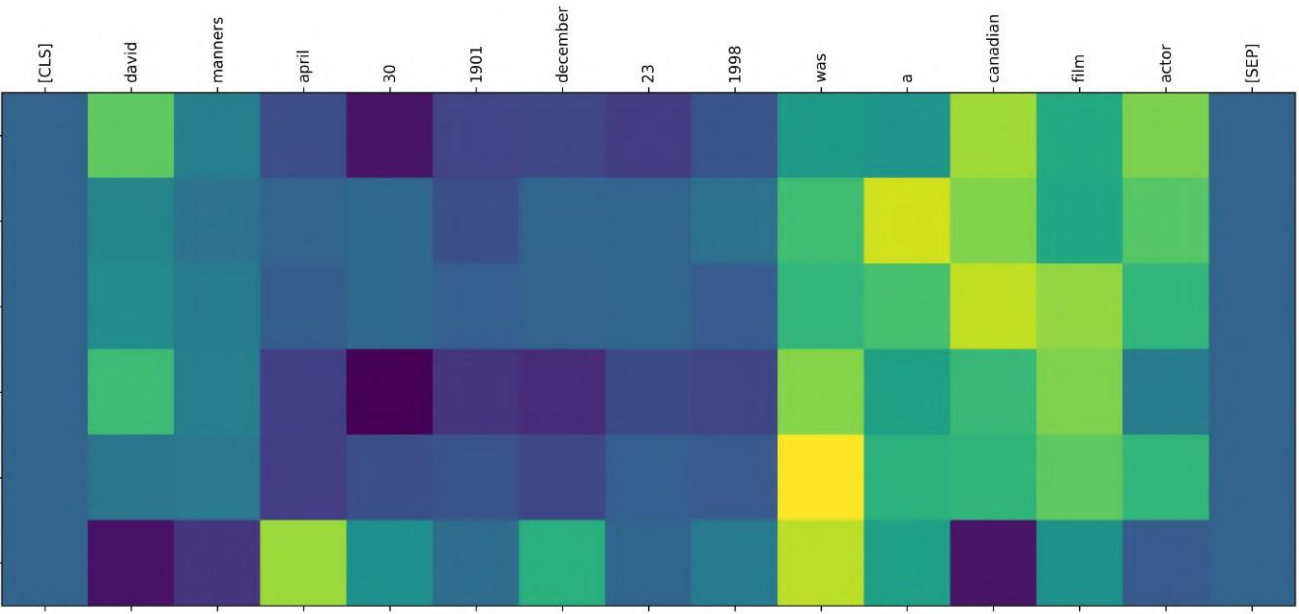

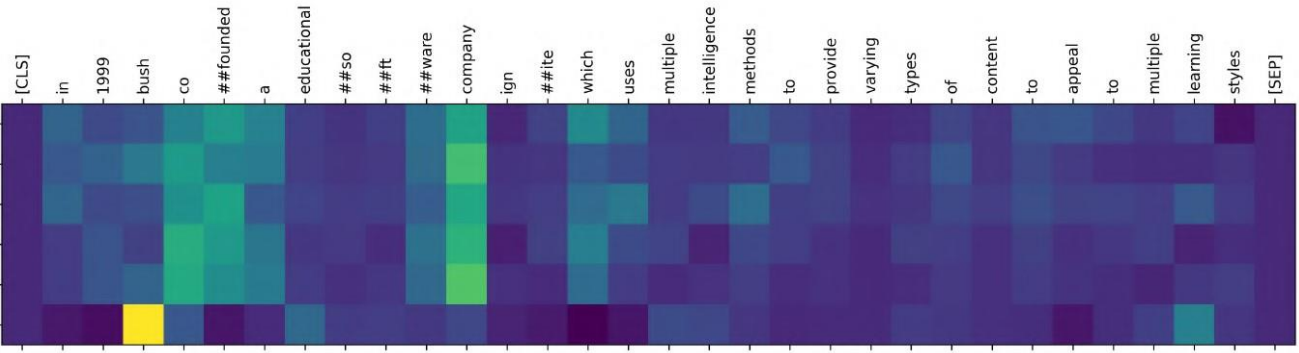

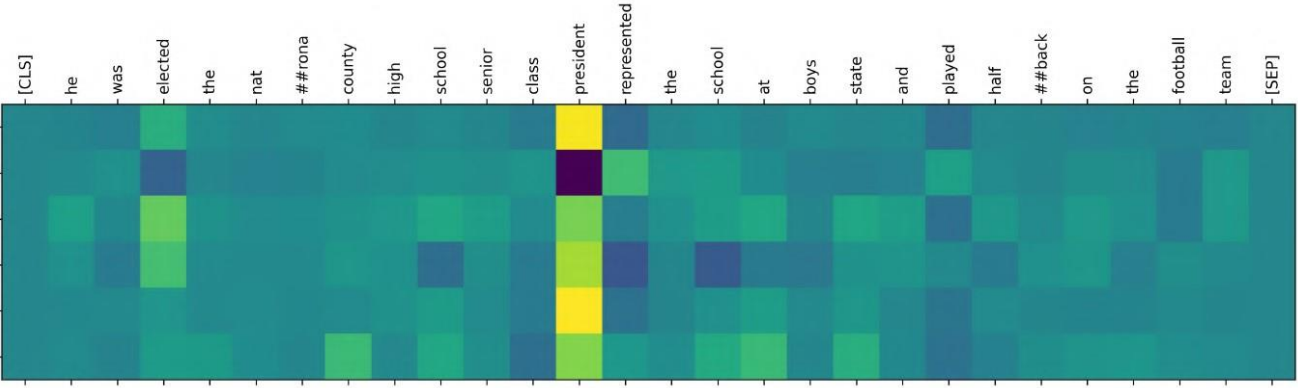

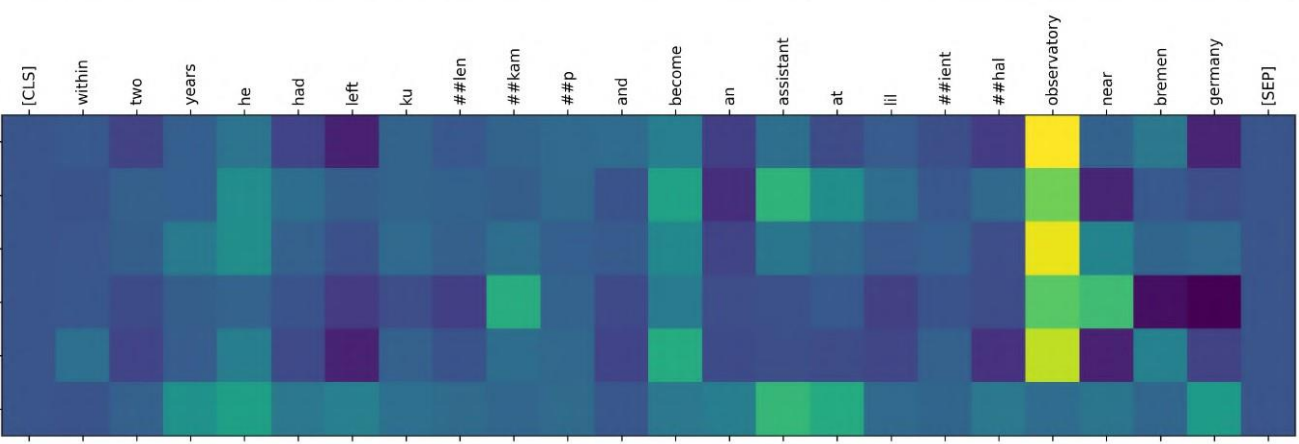

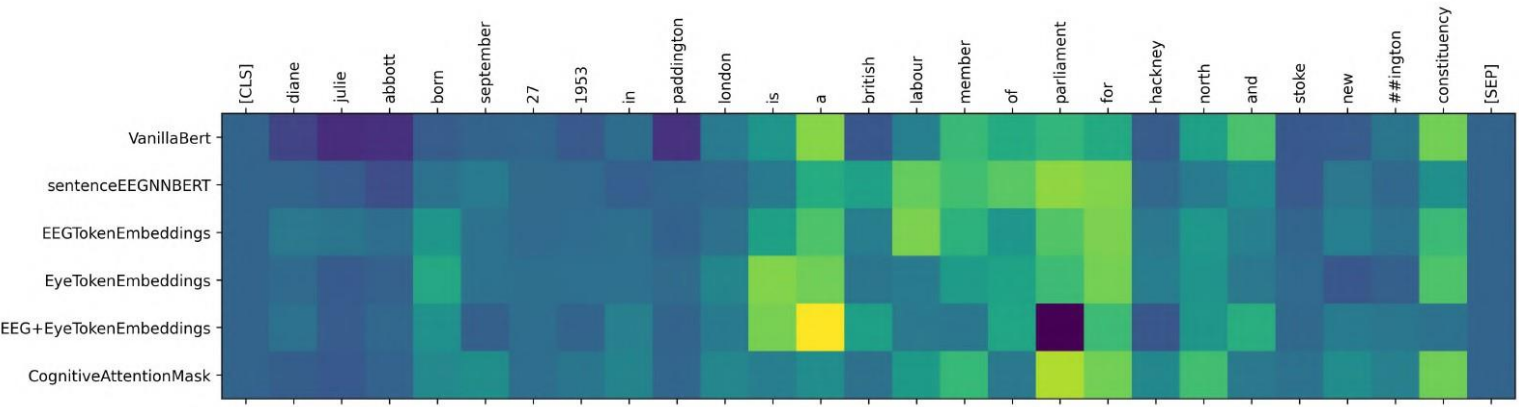

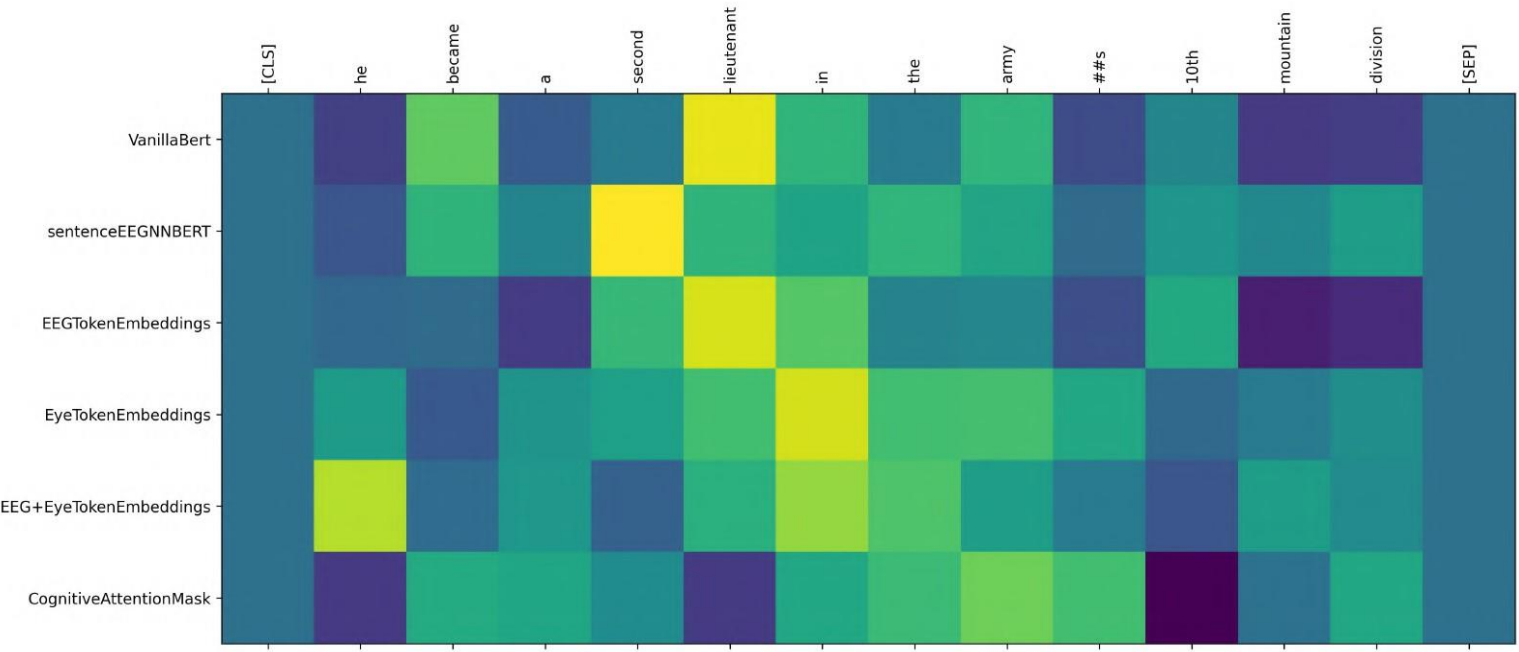

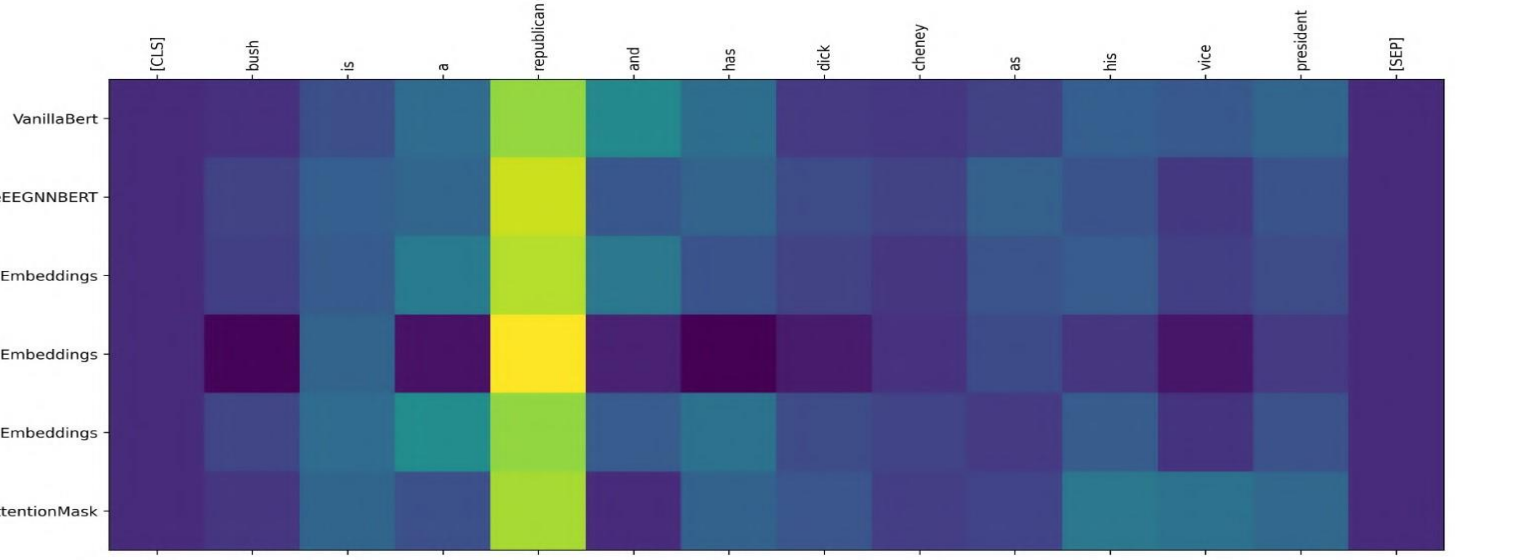

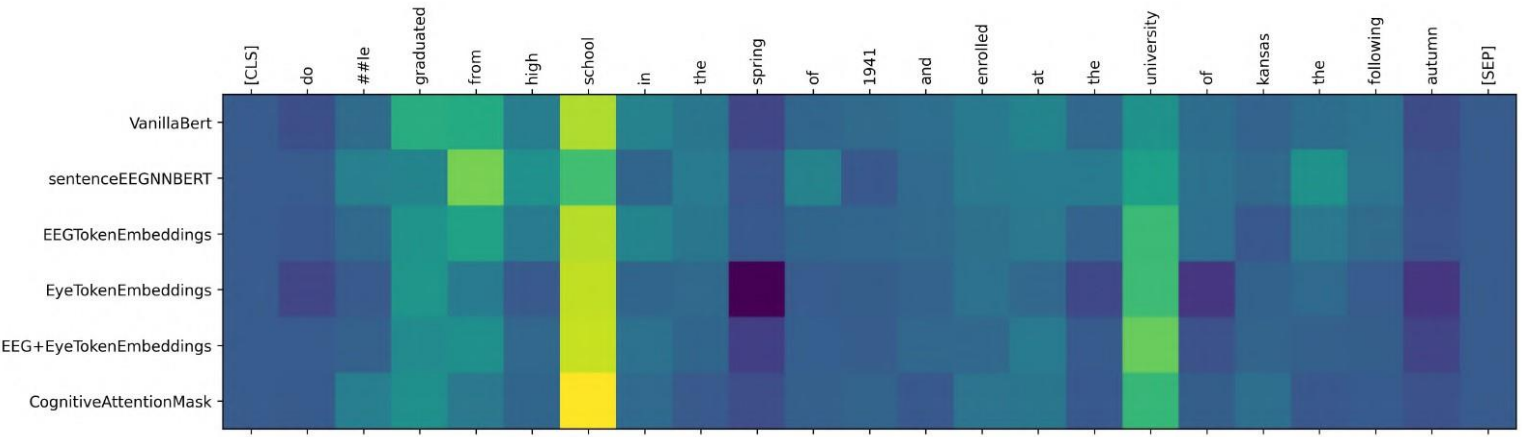

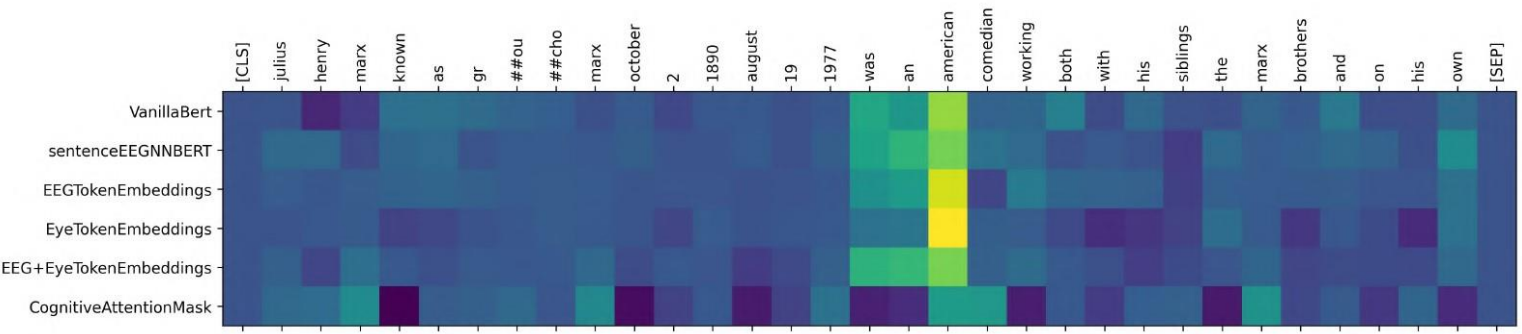

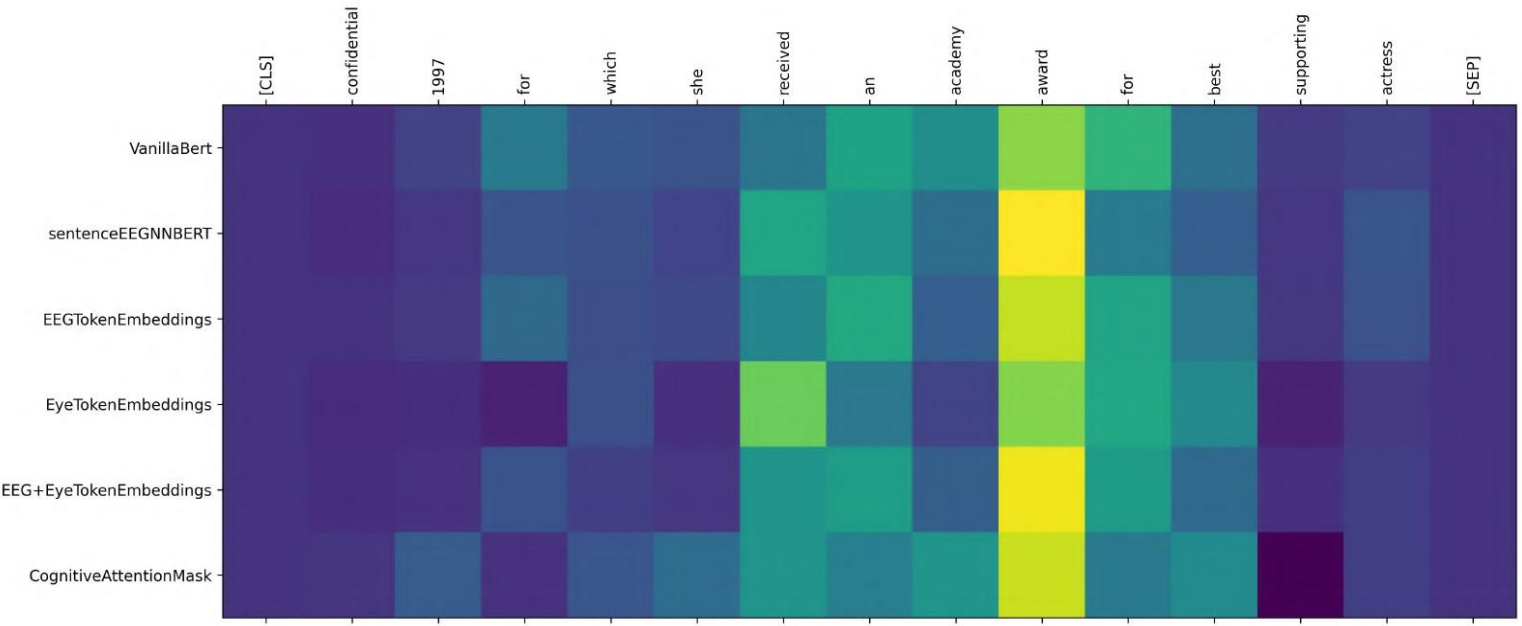

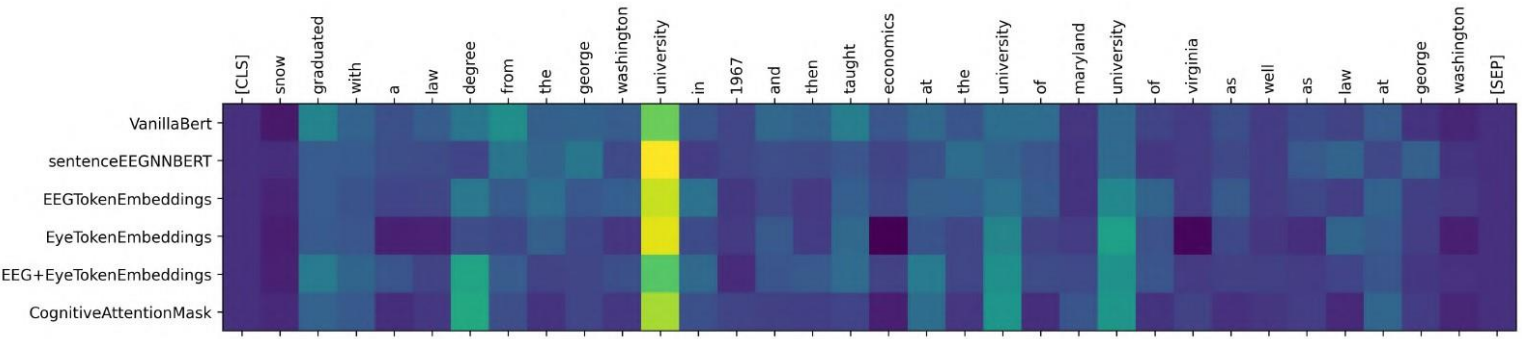

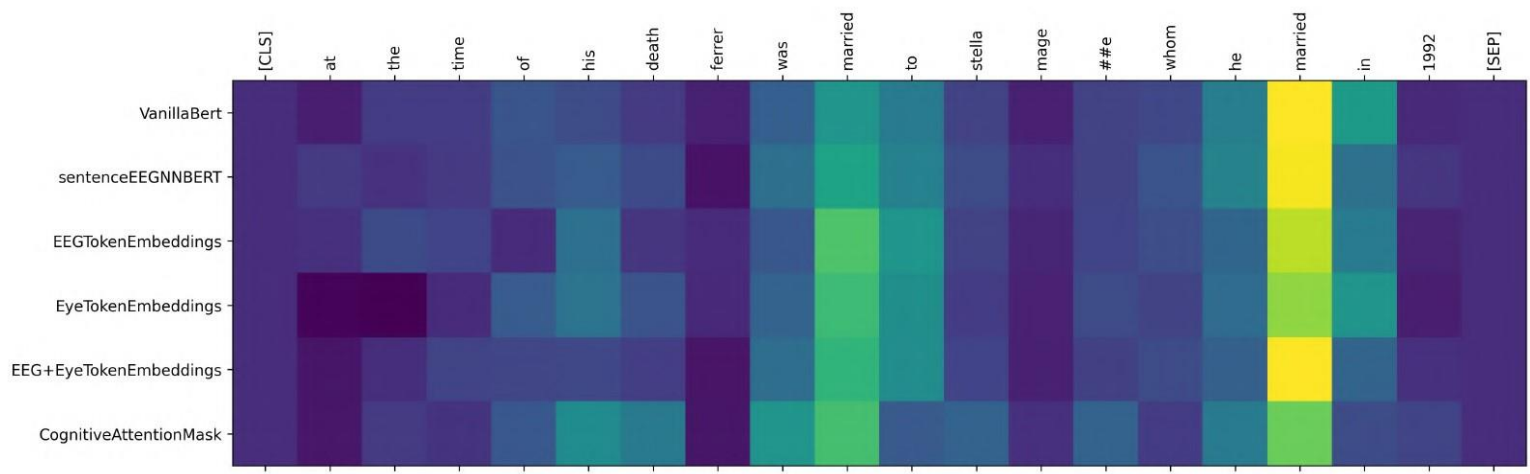

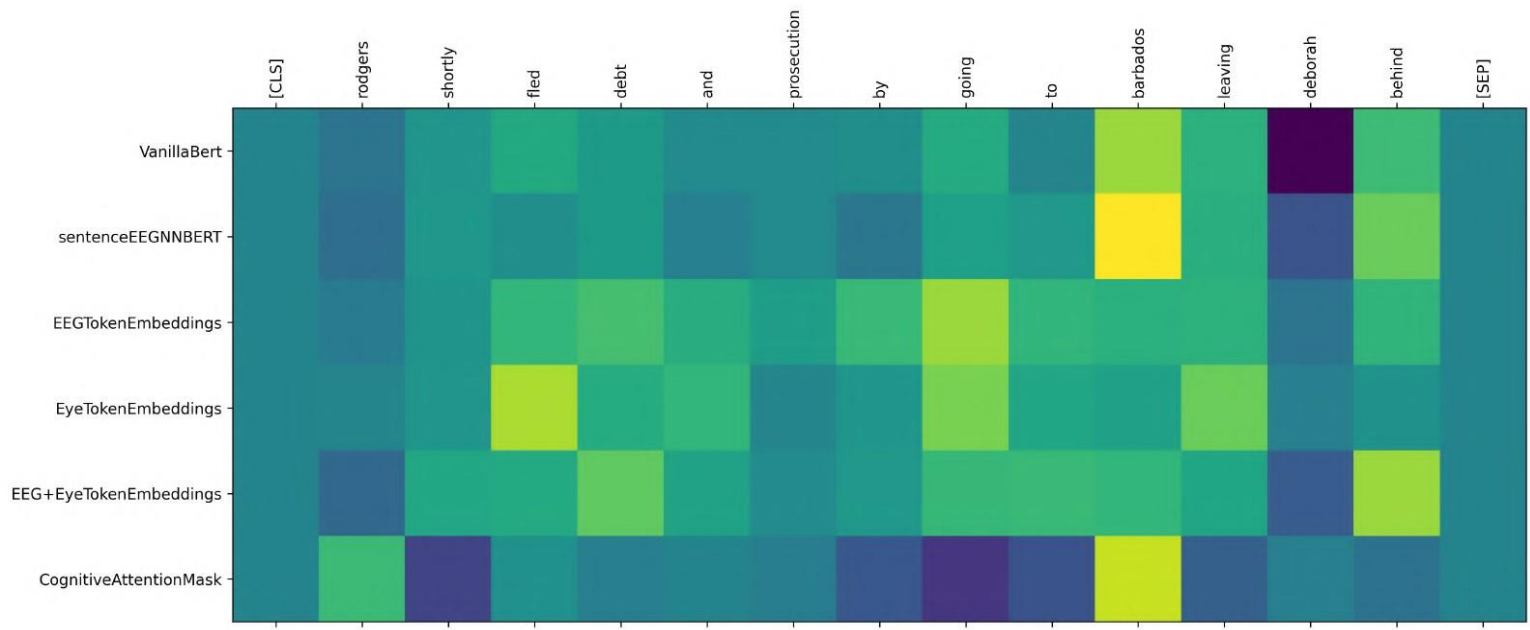

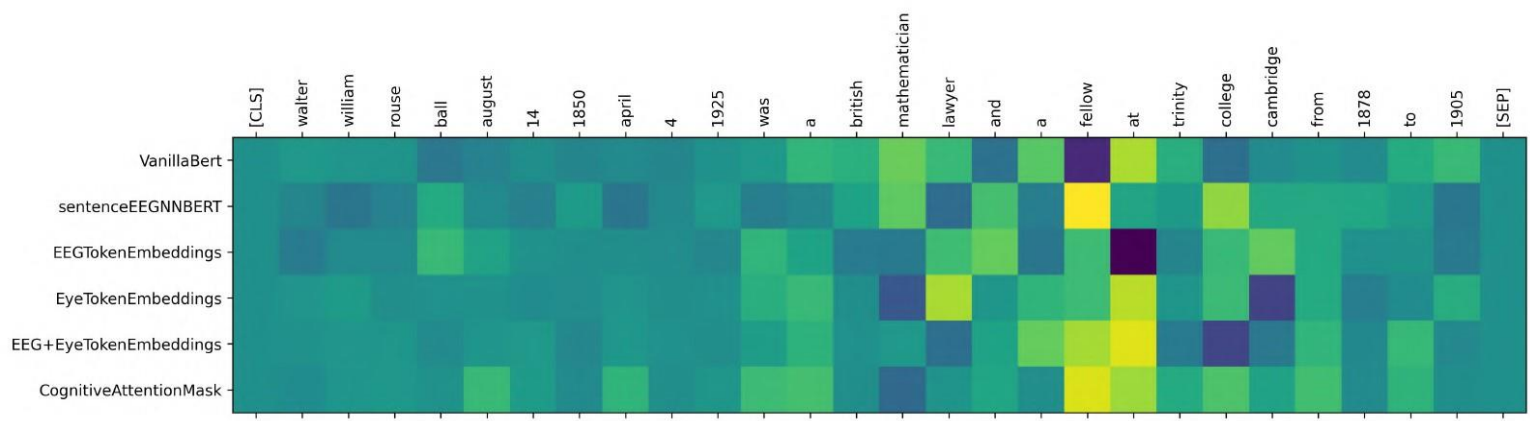

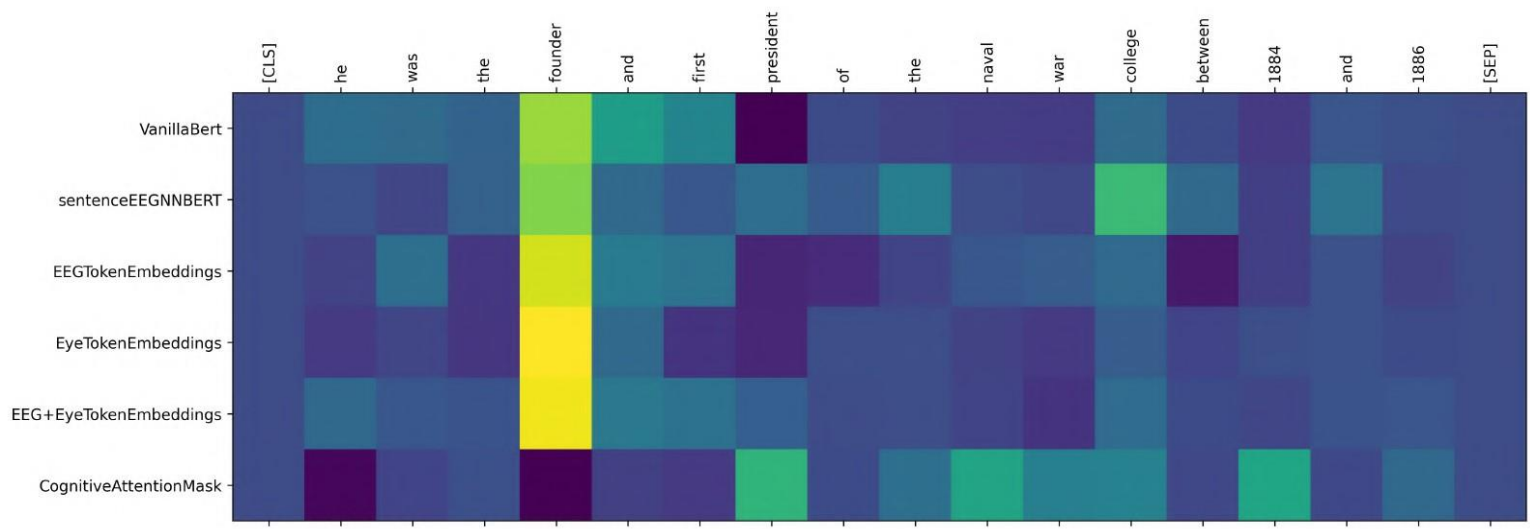

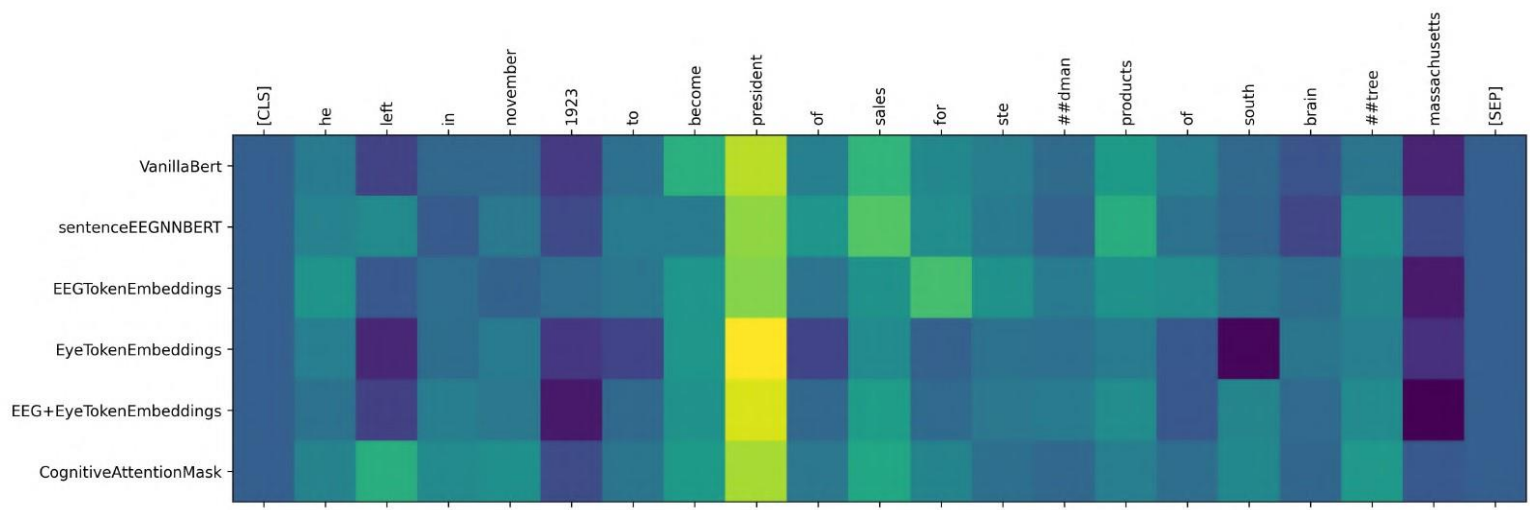

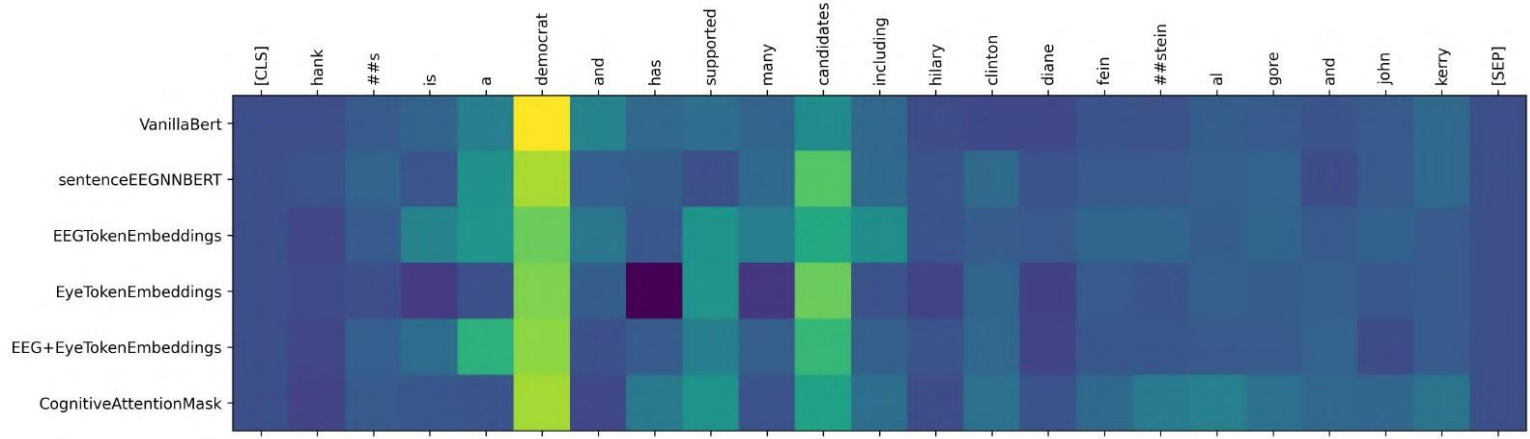

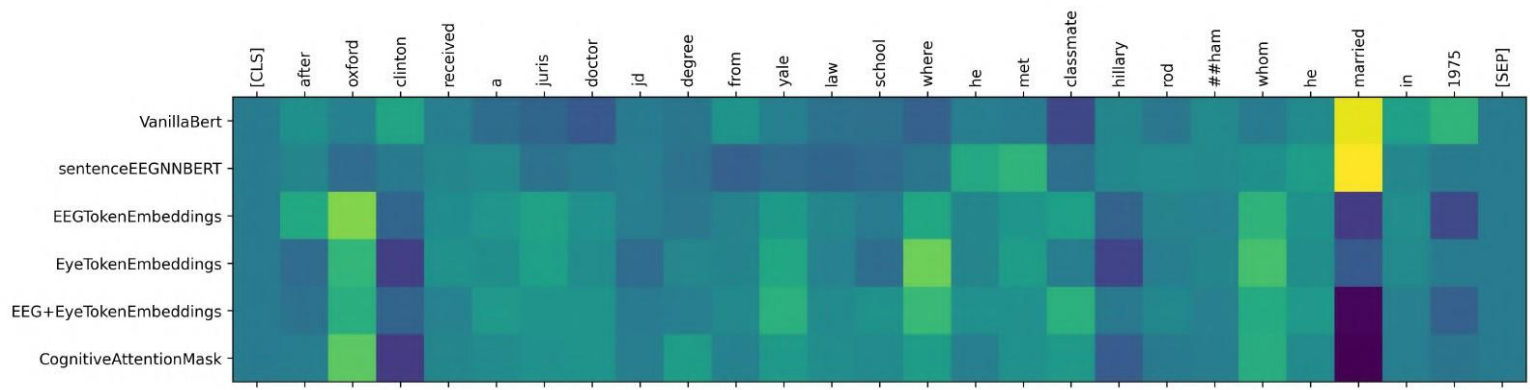

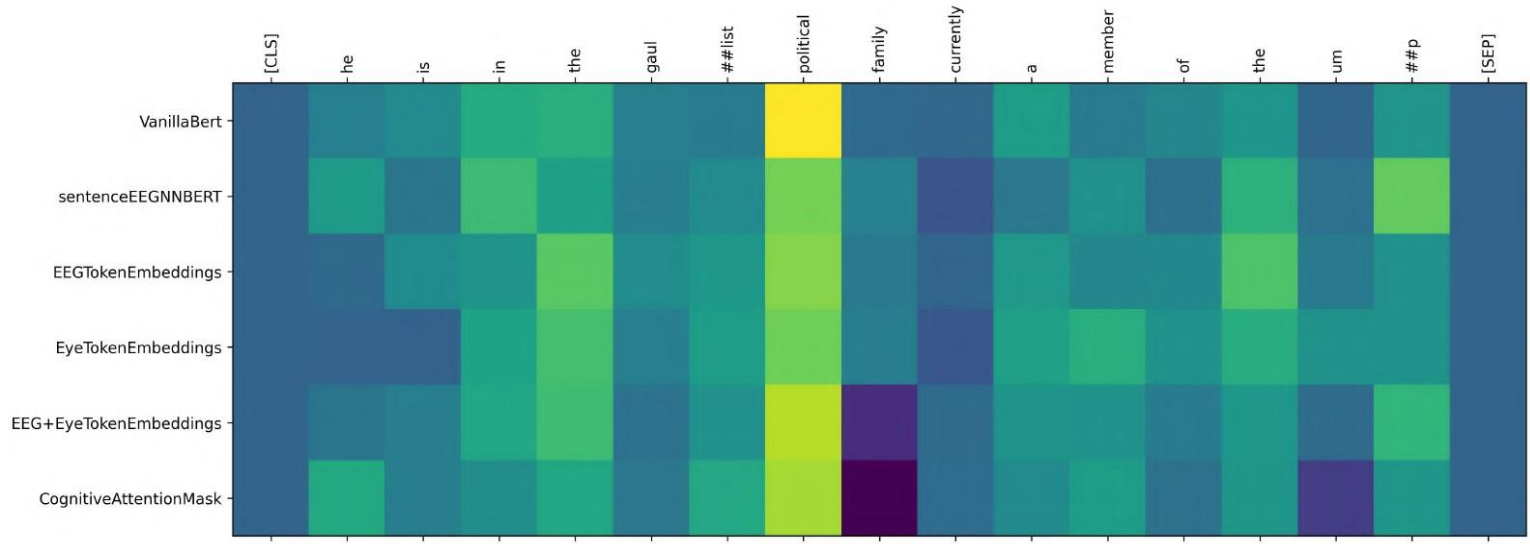

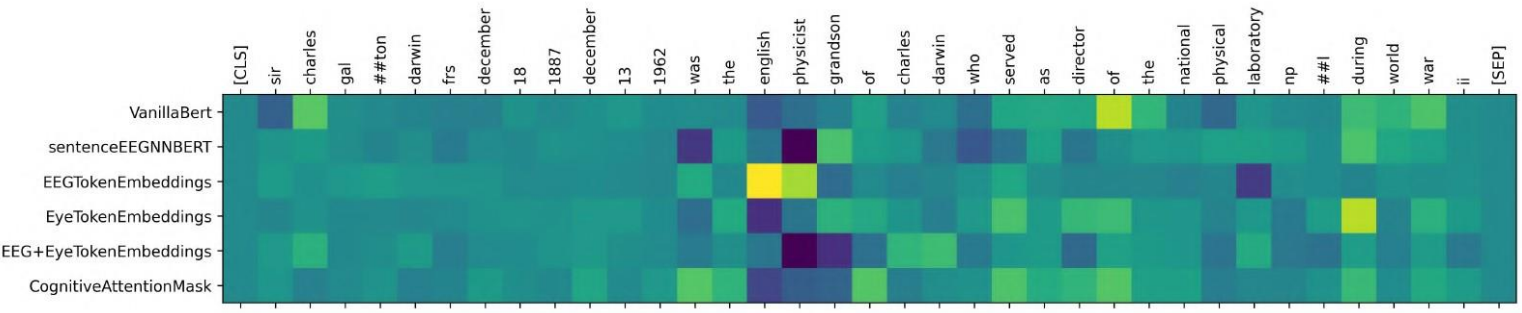

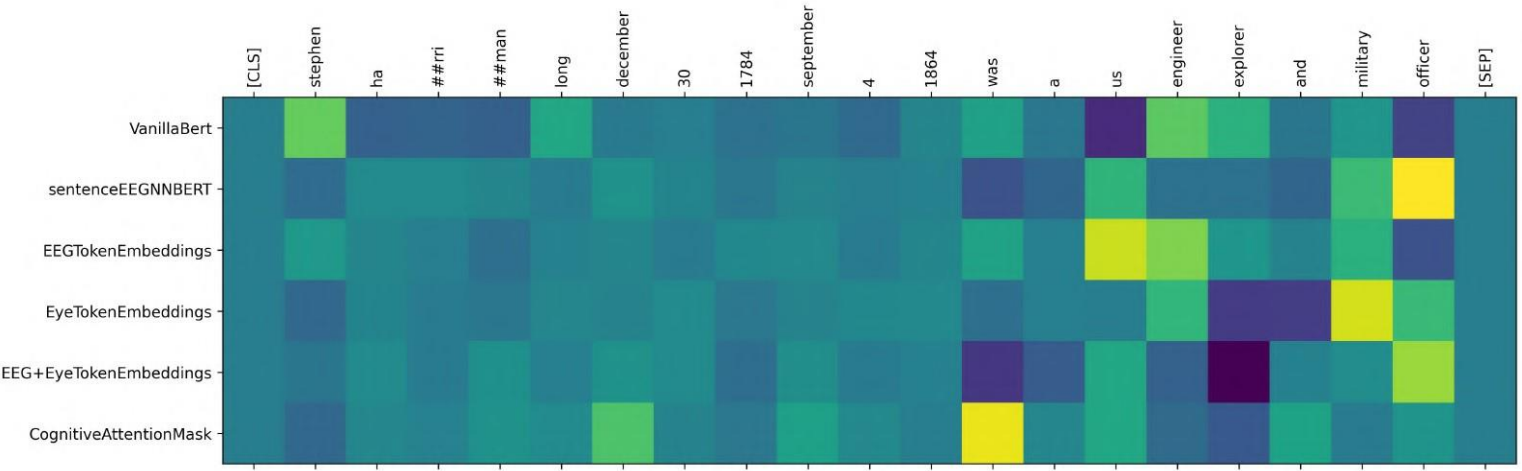

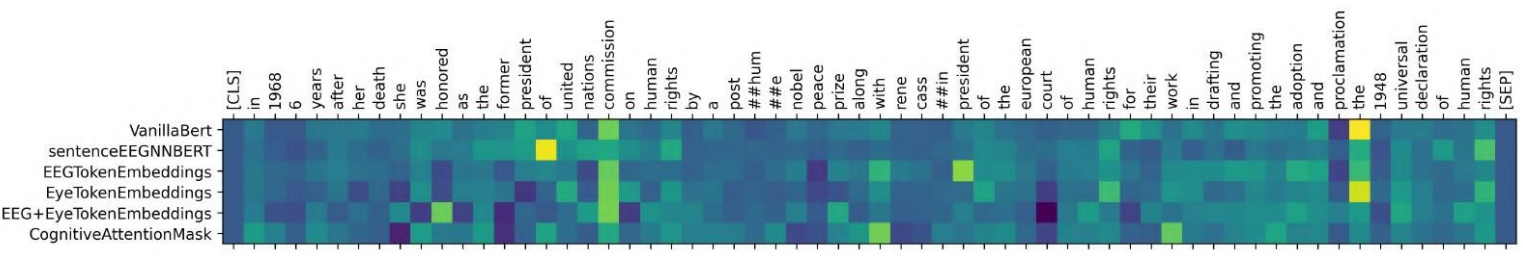

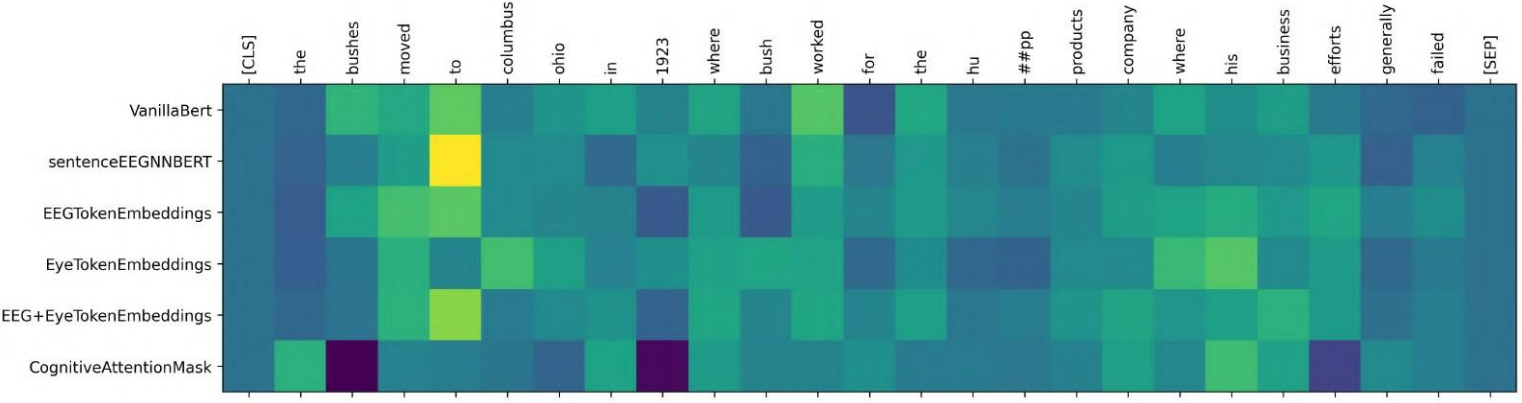

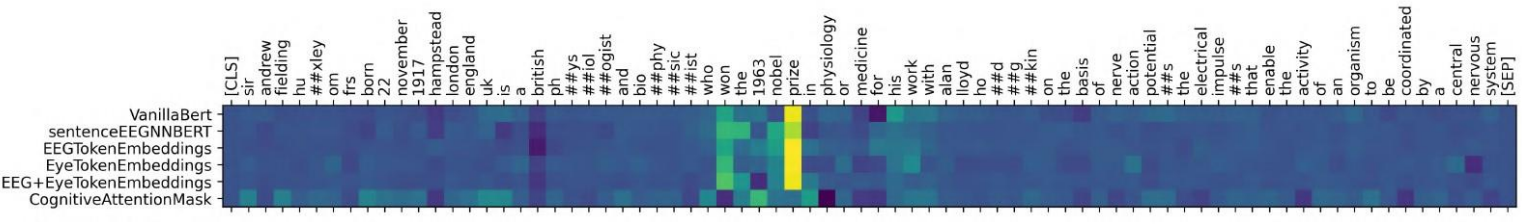

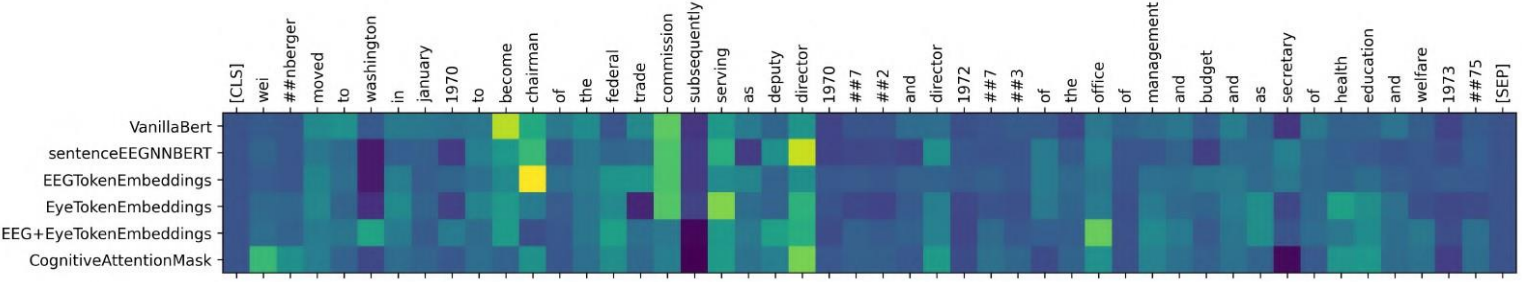

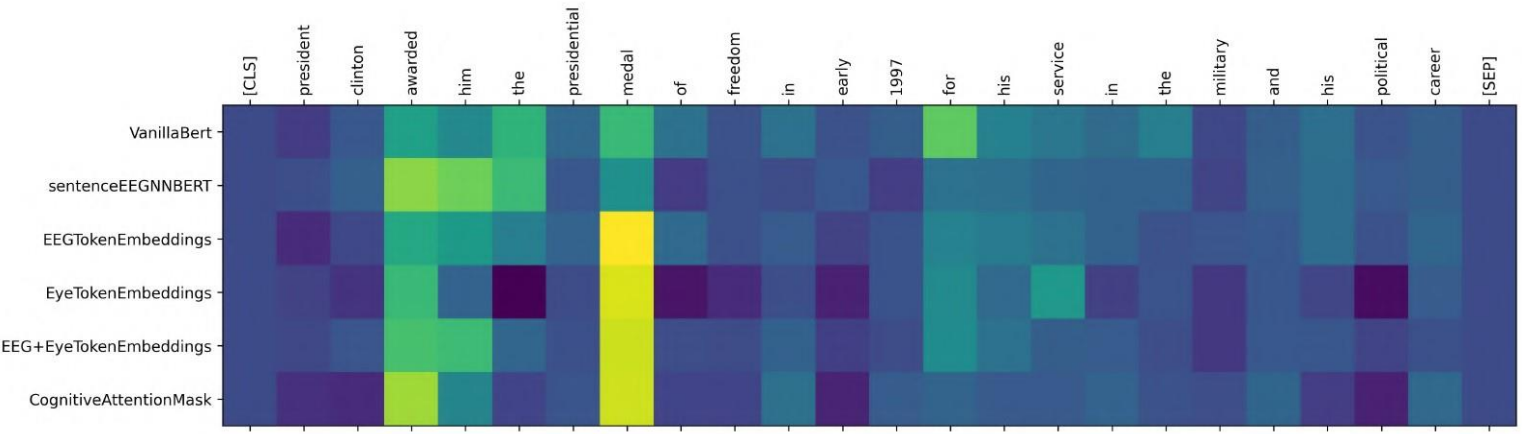

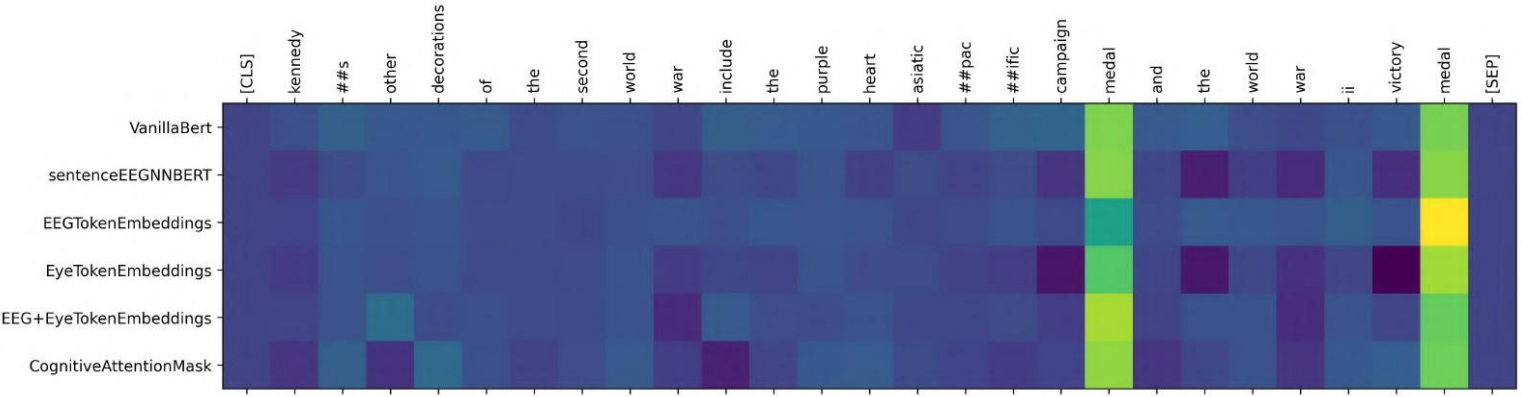

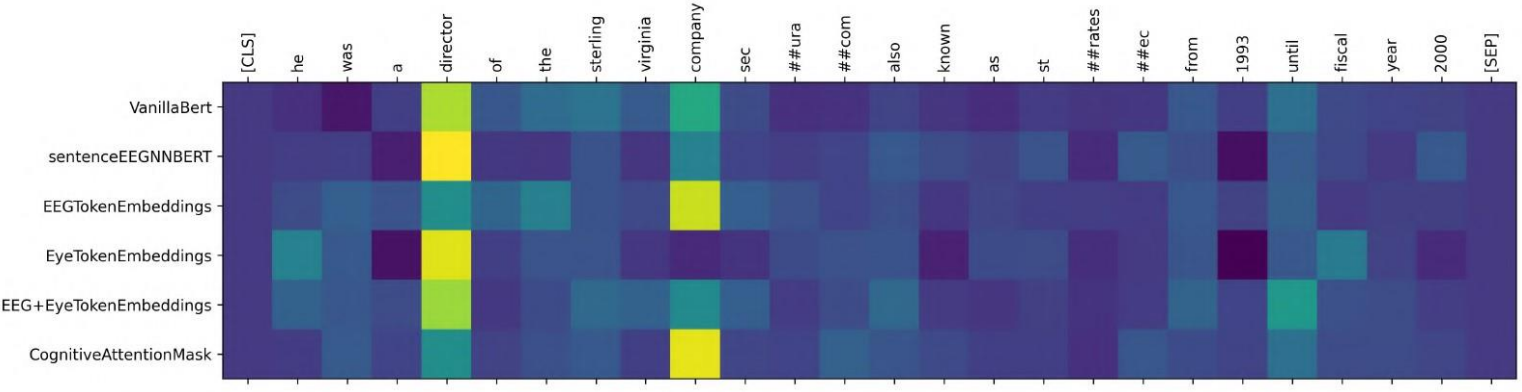

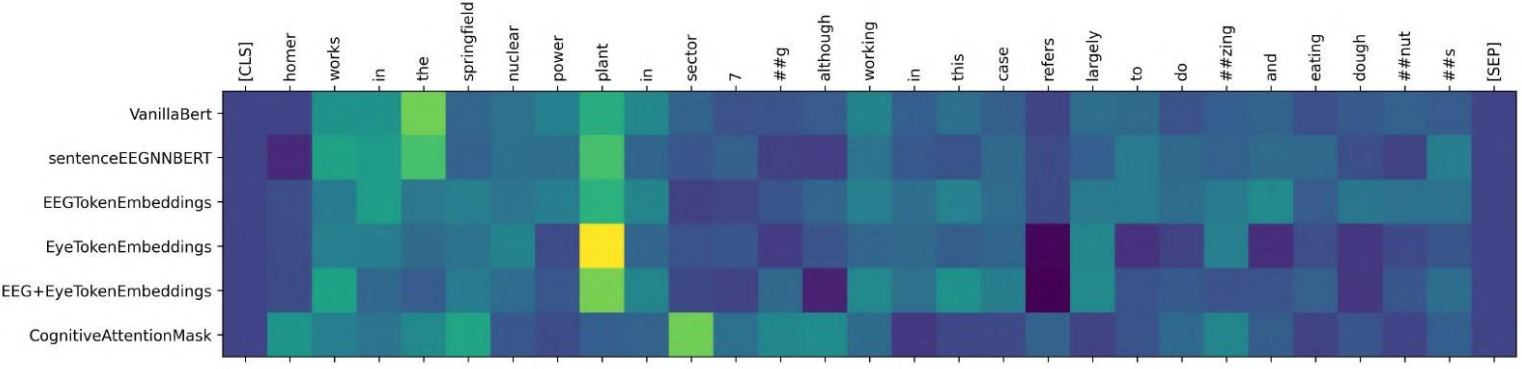

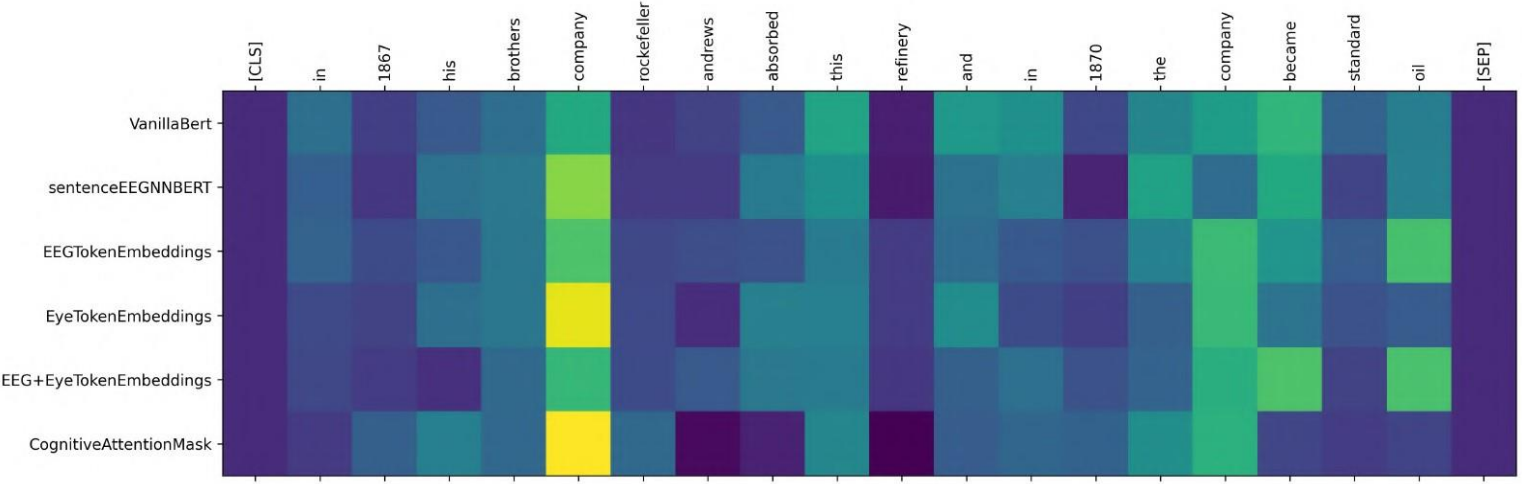

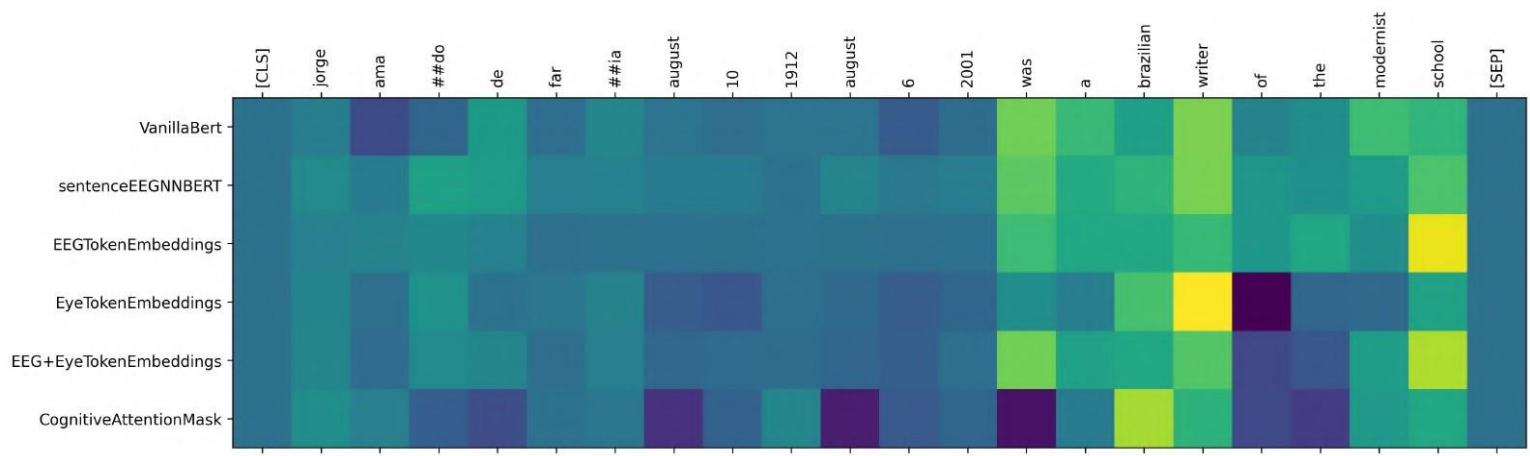

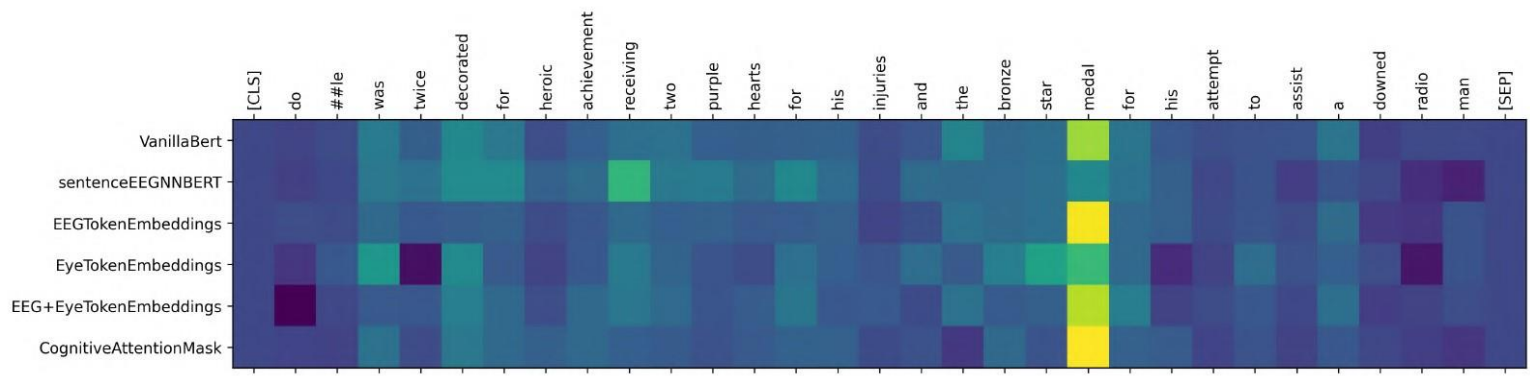

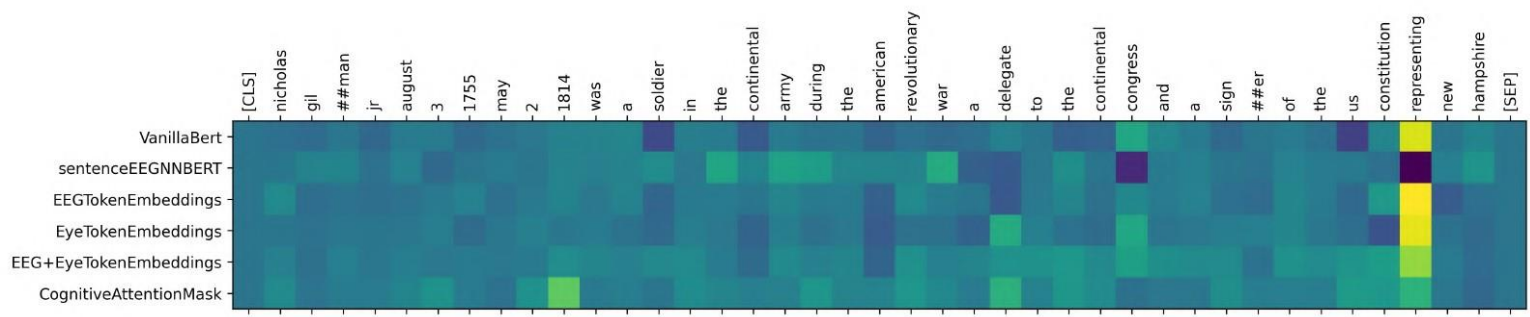

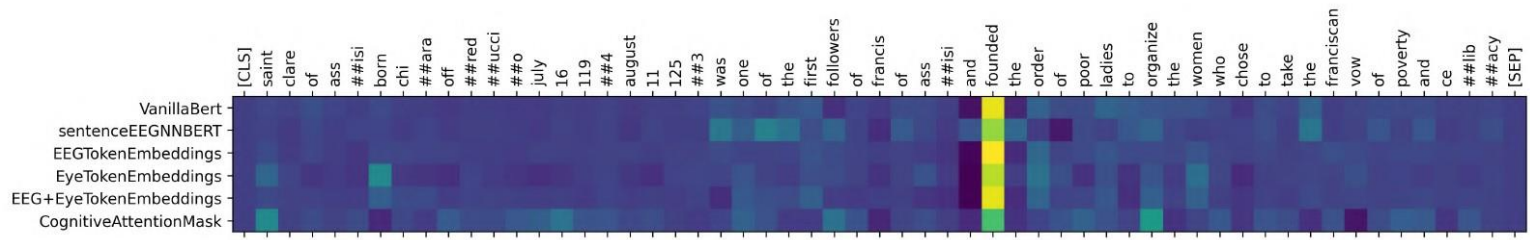

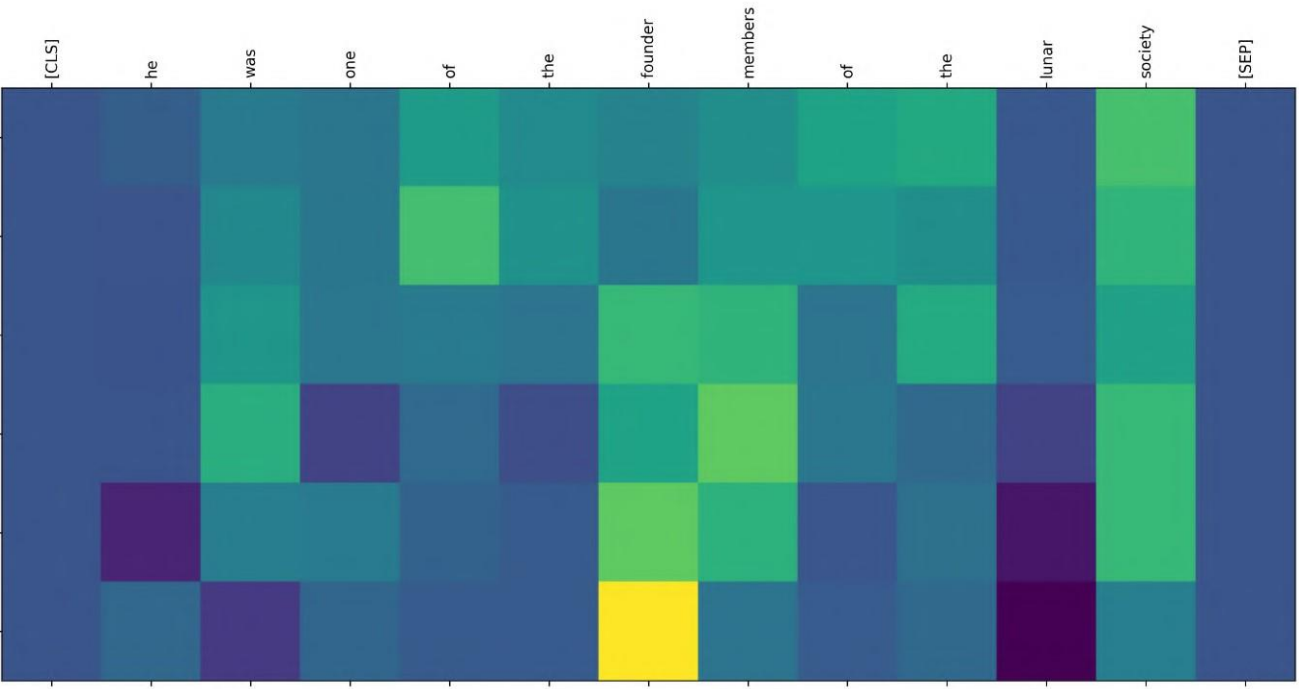

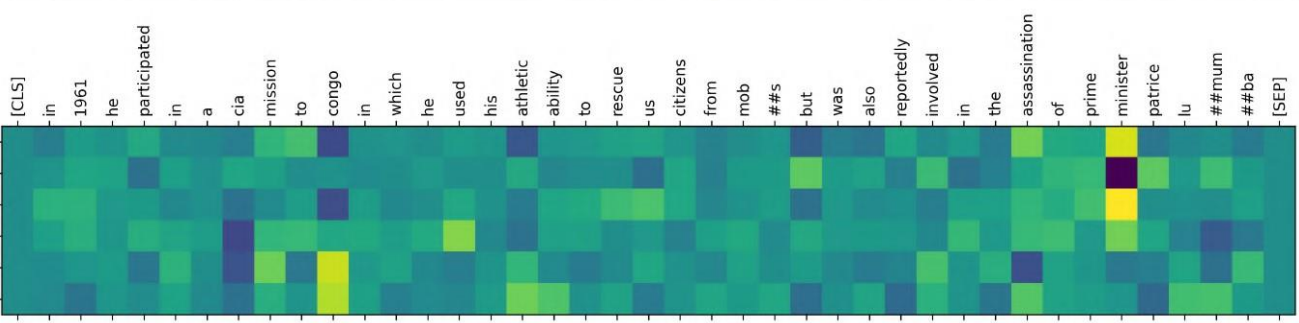

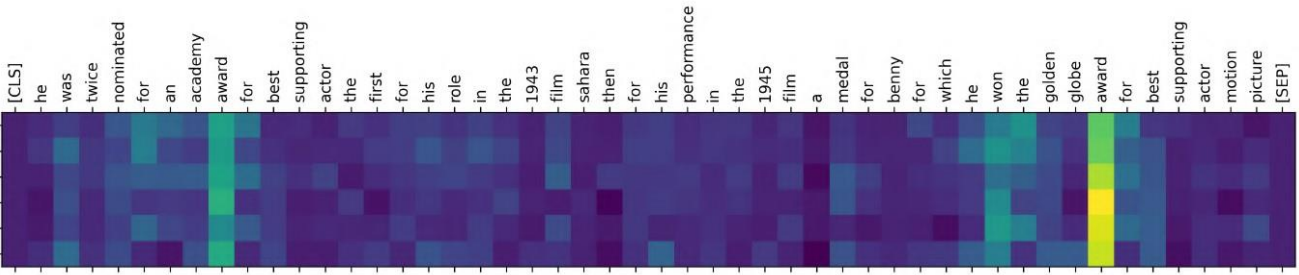

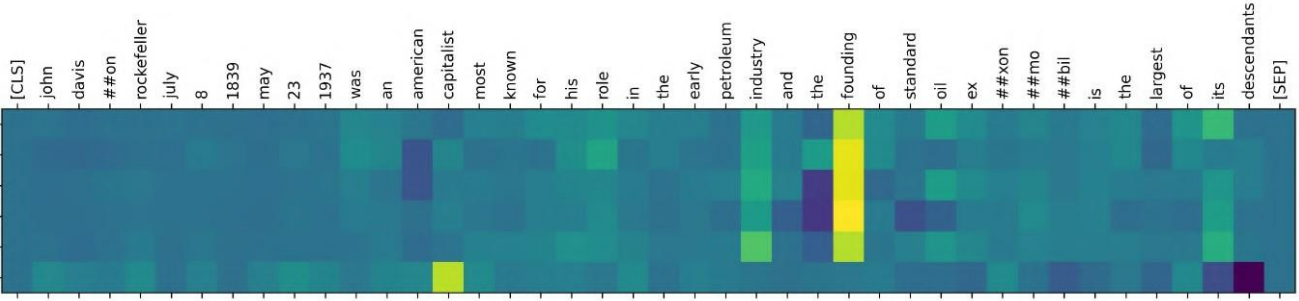

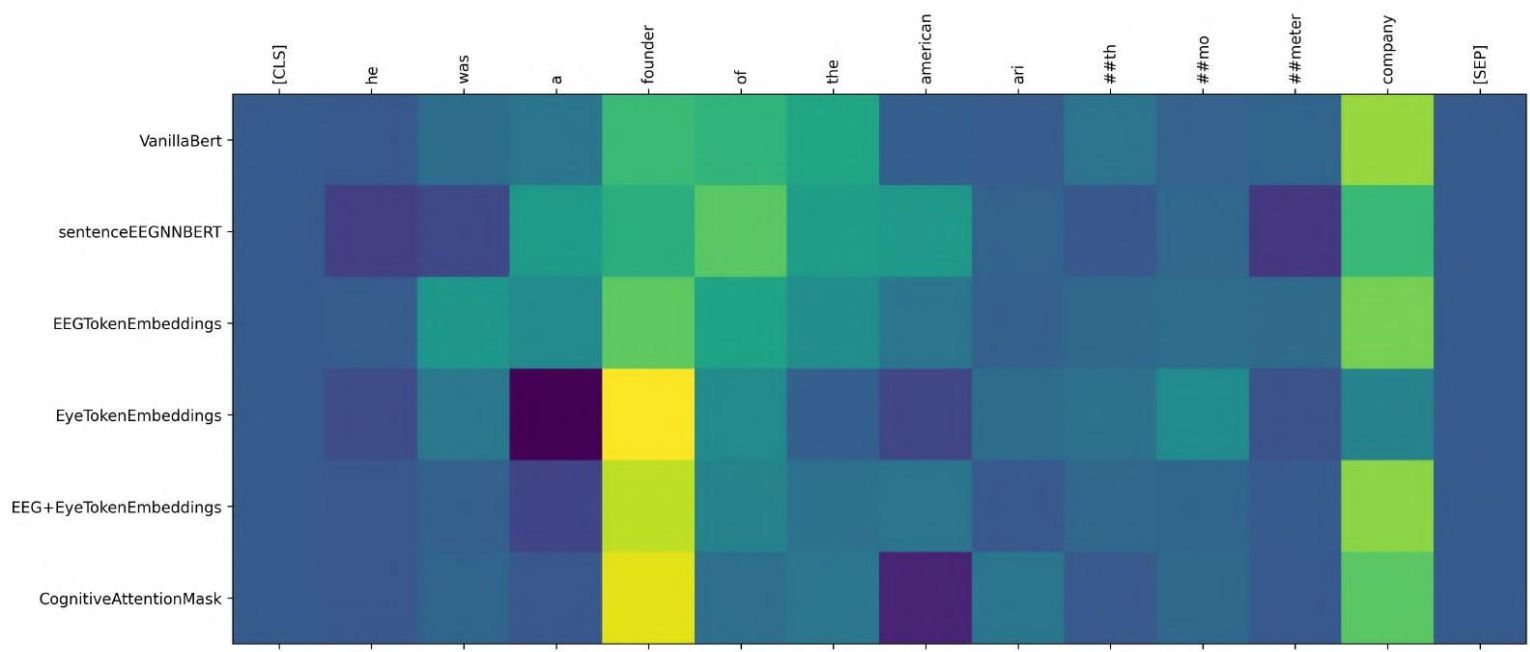

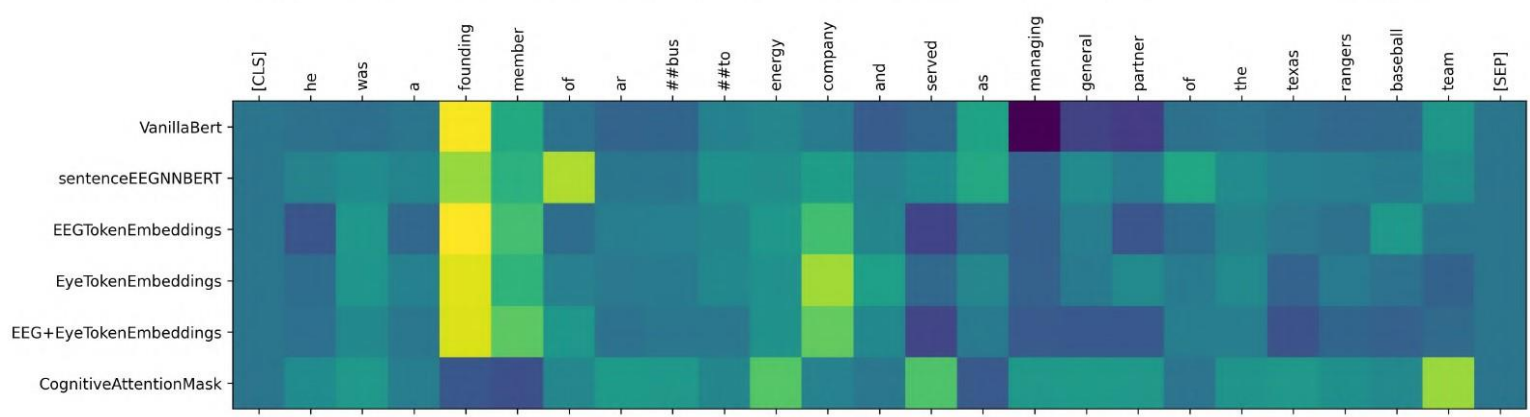

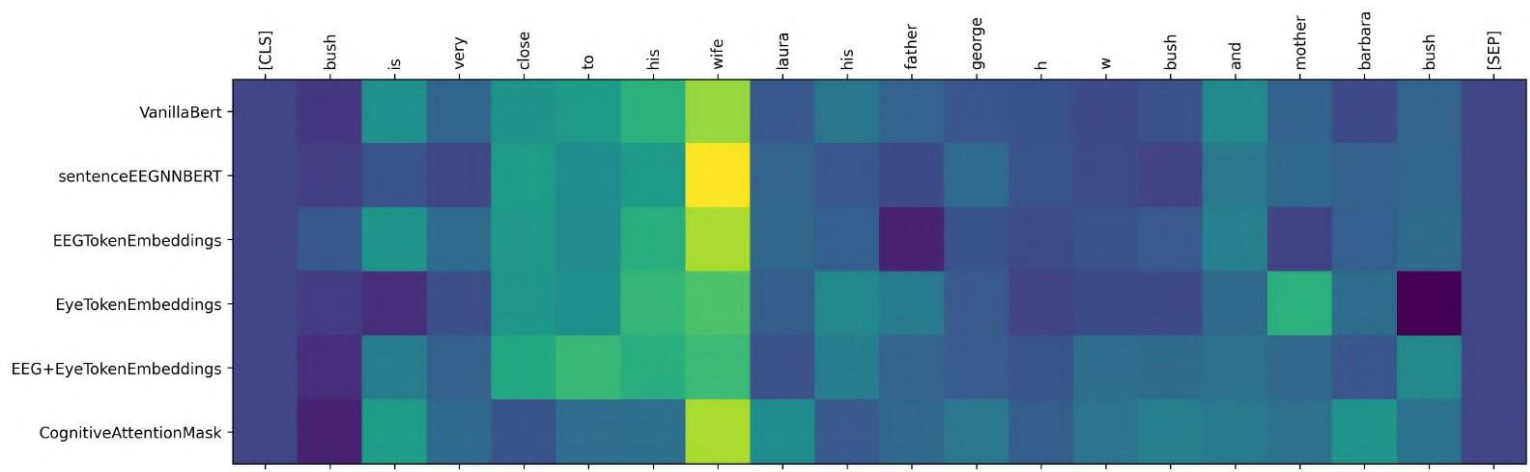

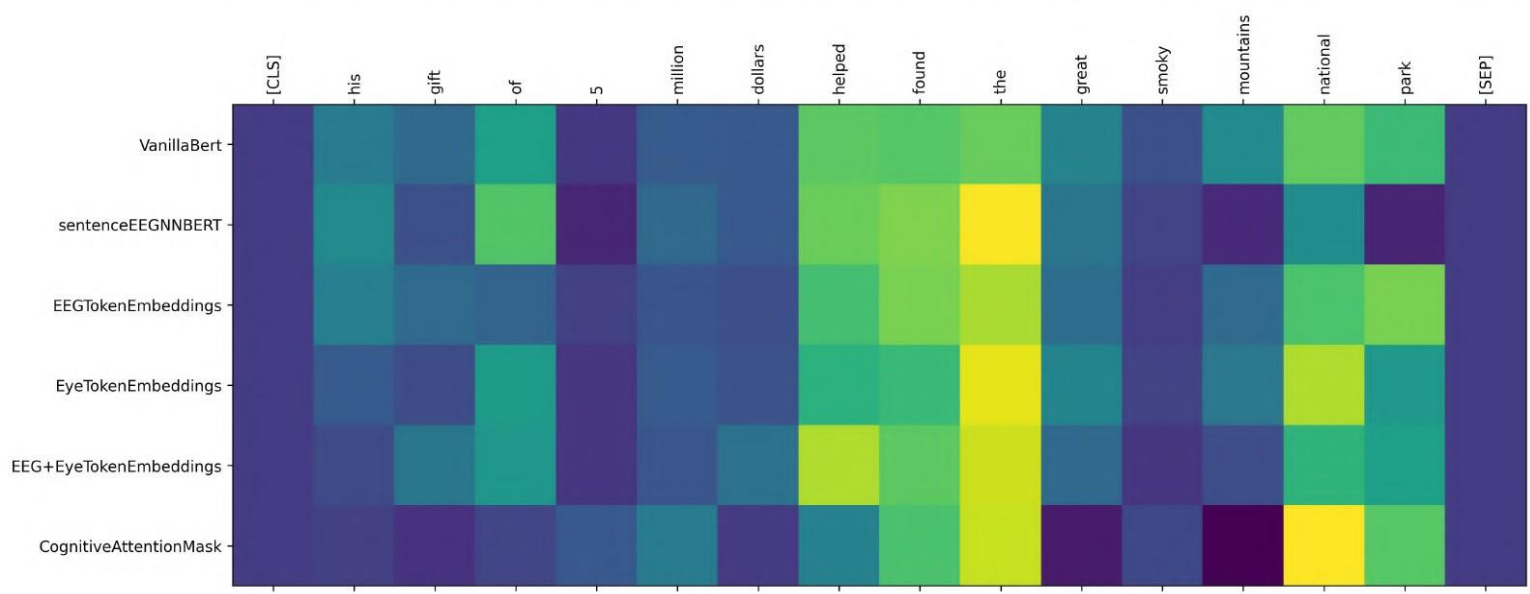

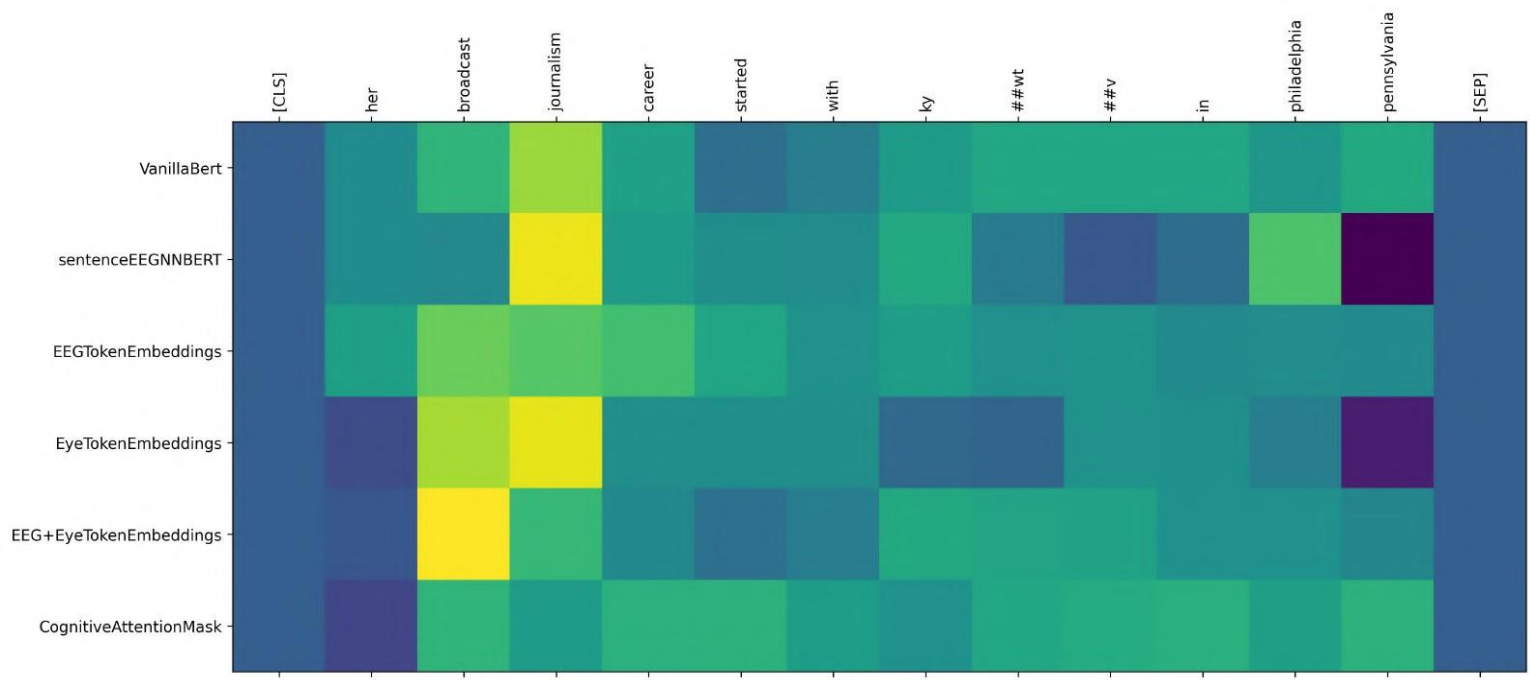

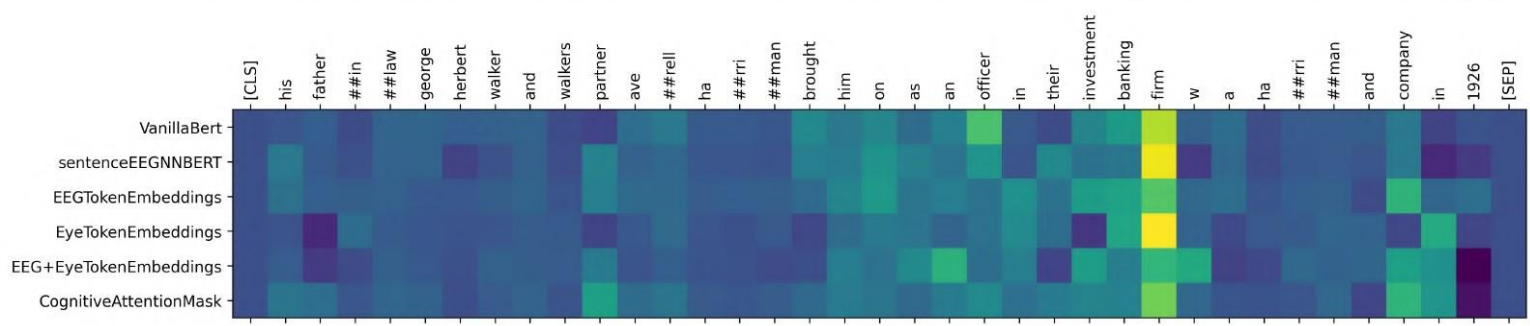

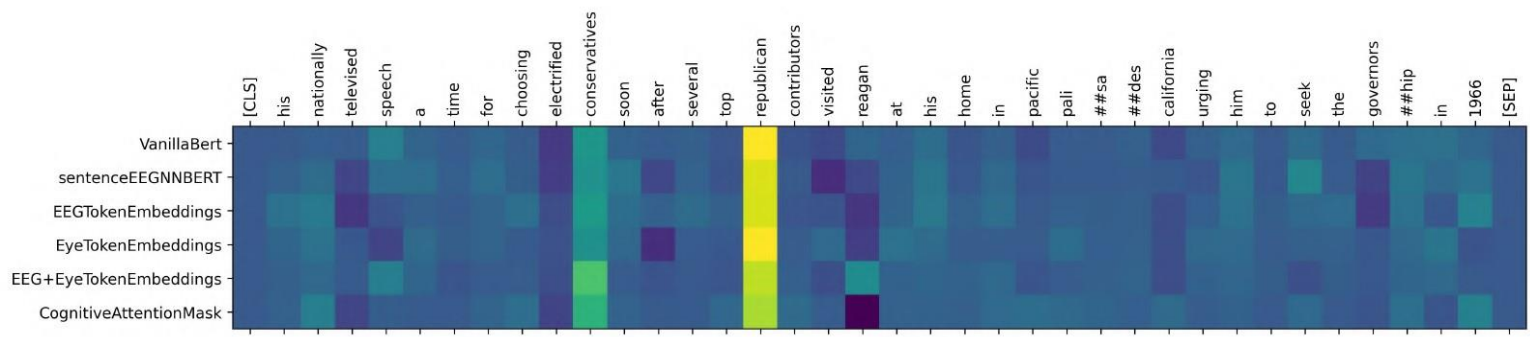

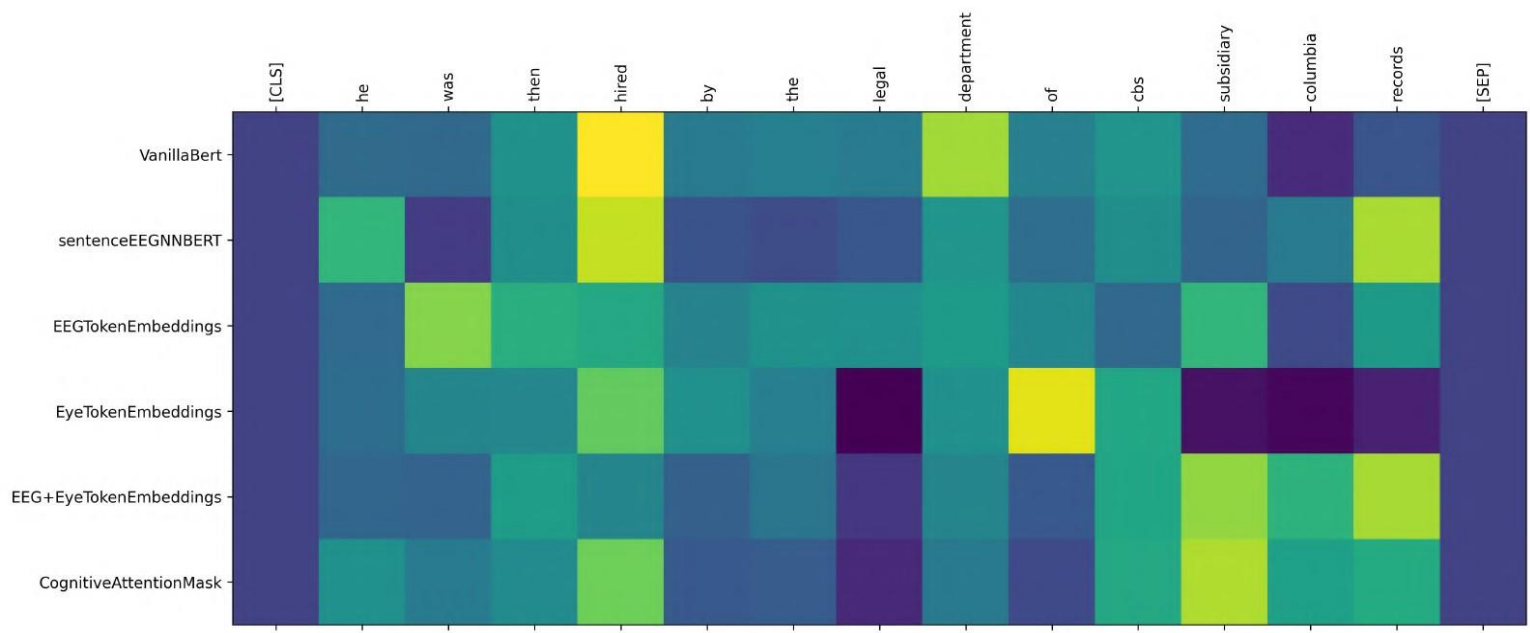

# Appendix B: Heat maps for incoming self-attention weights for the test corpus

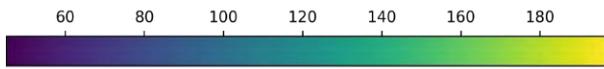

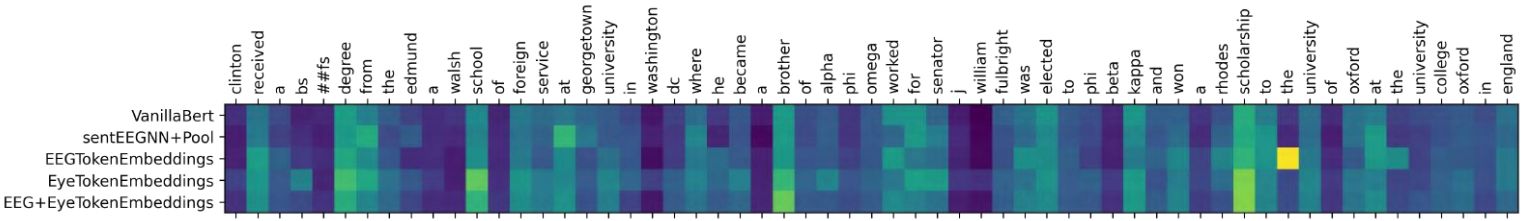

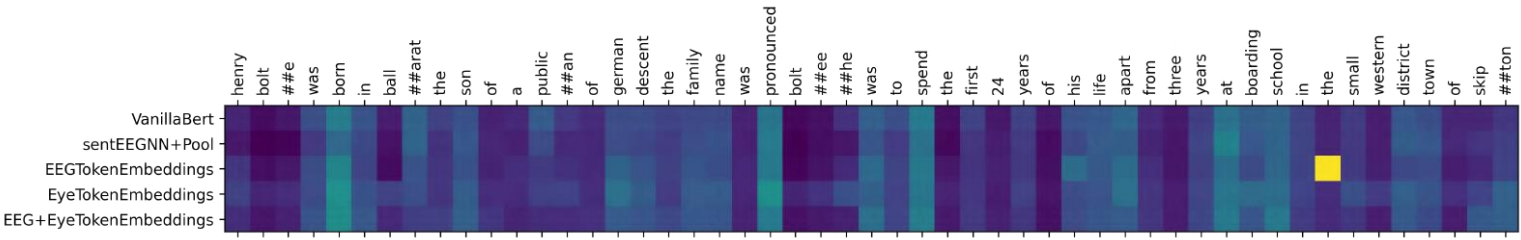

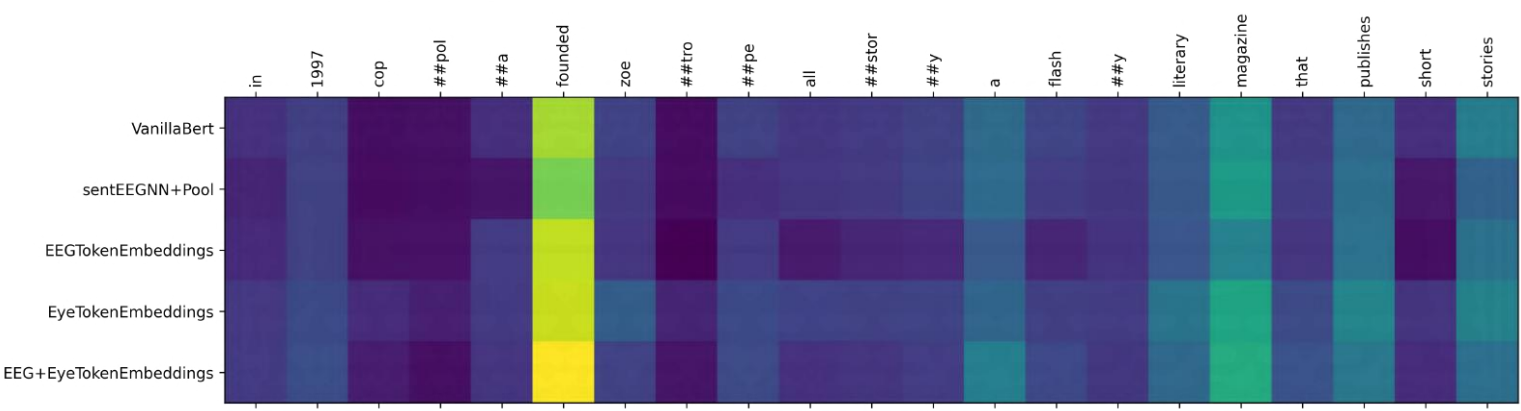

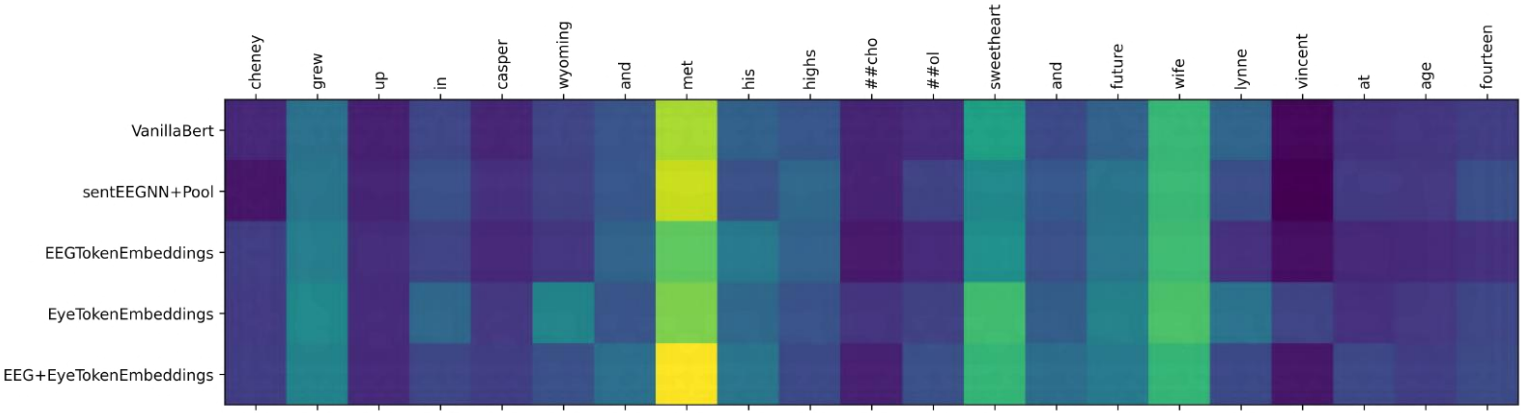

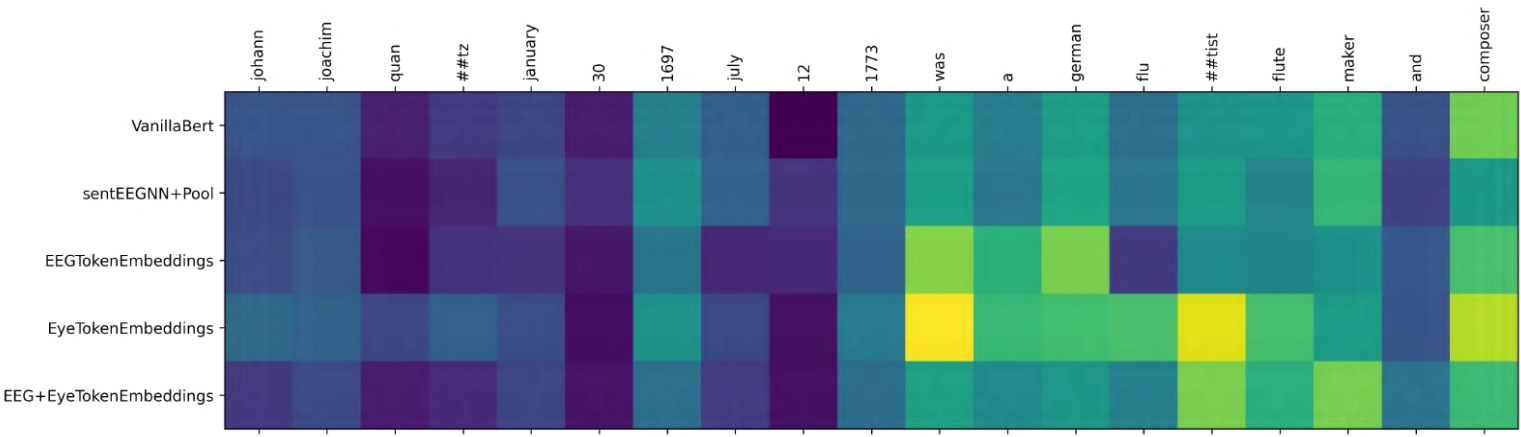

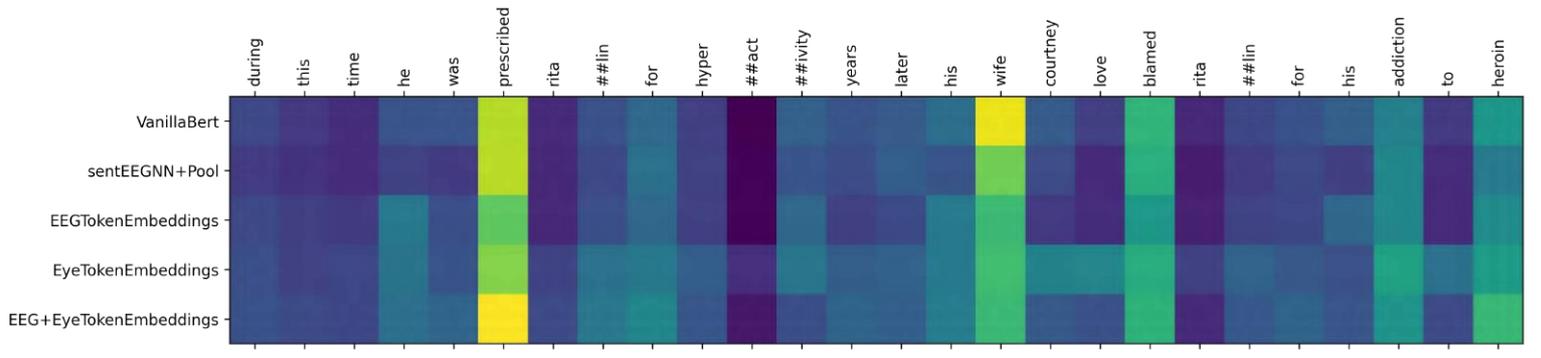

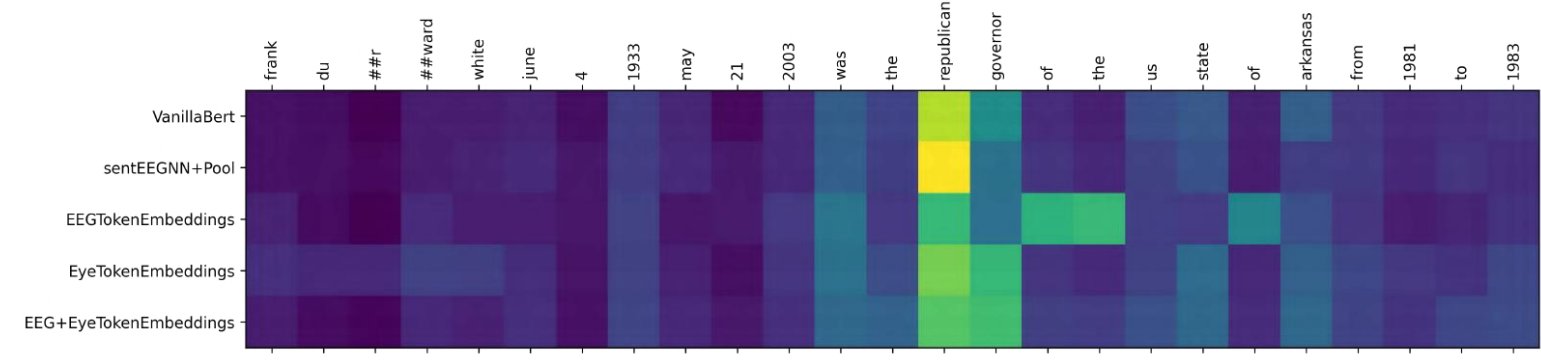

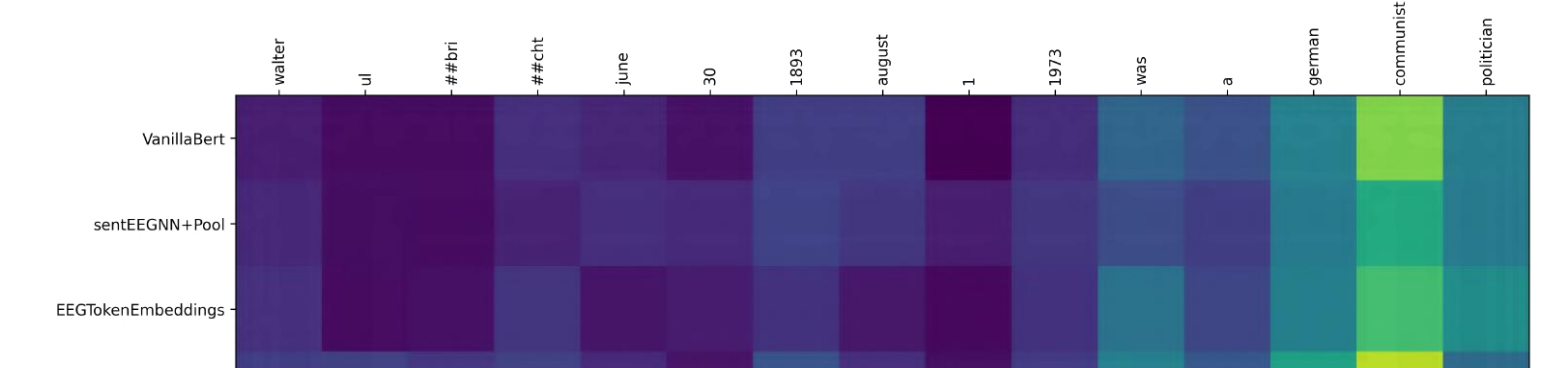

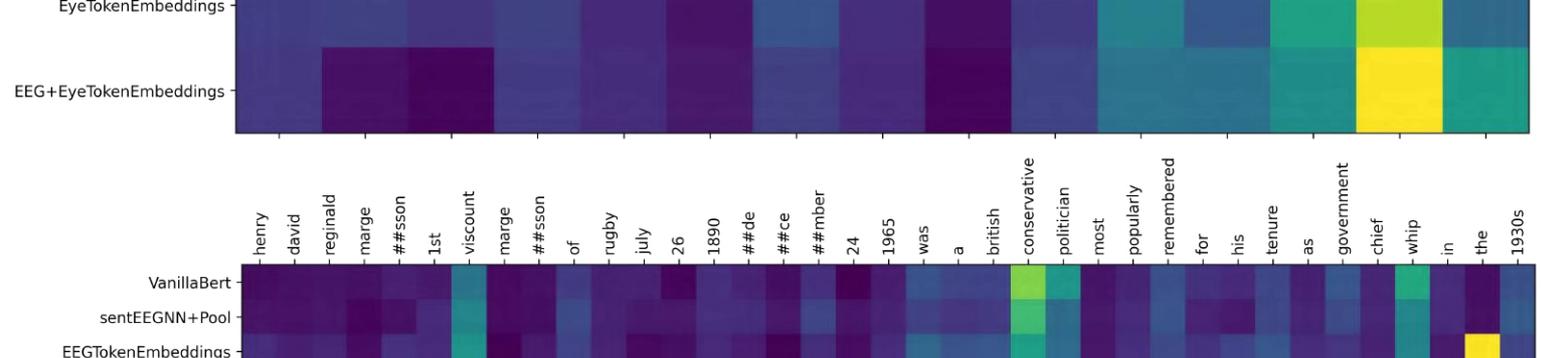

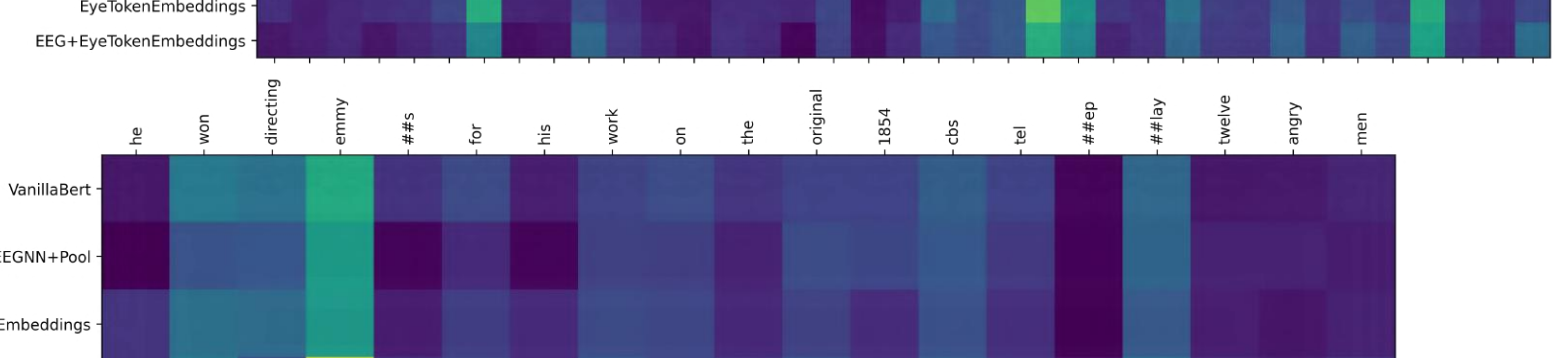

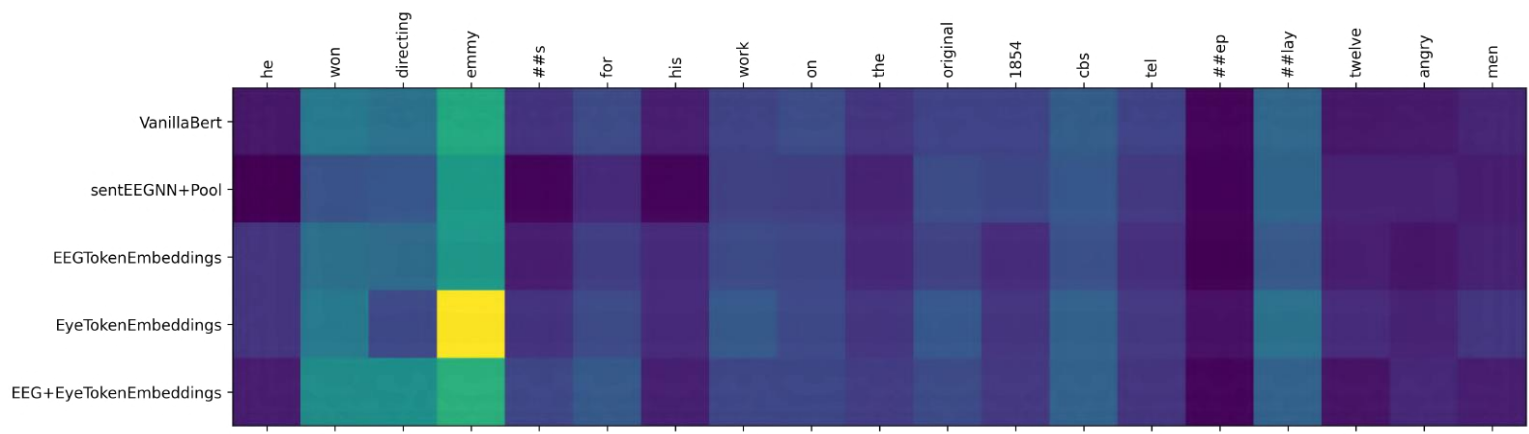

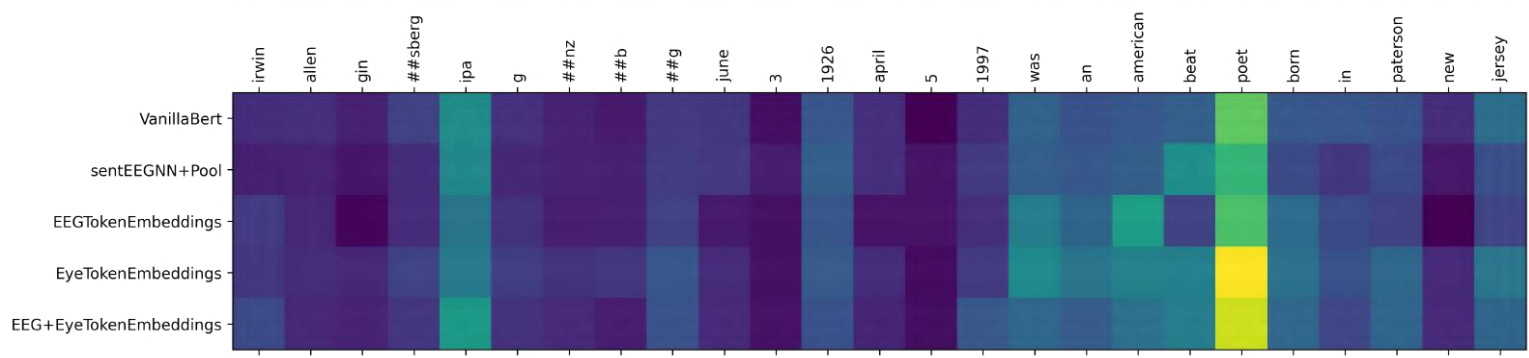

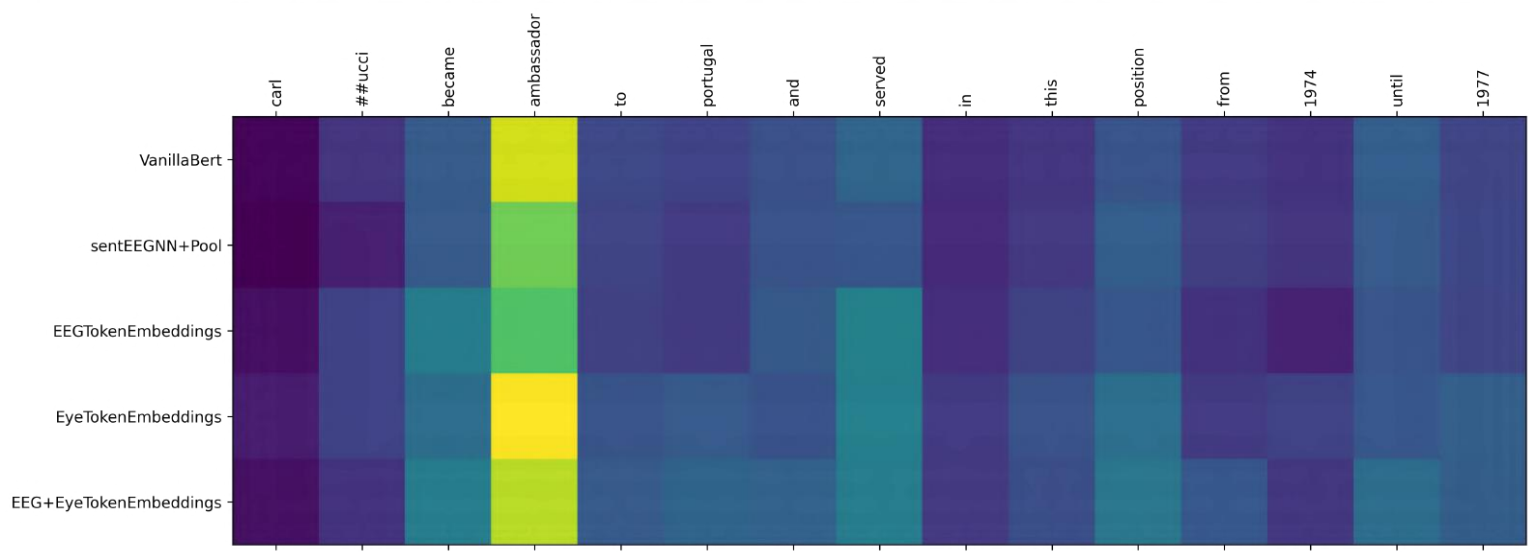

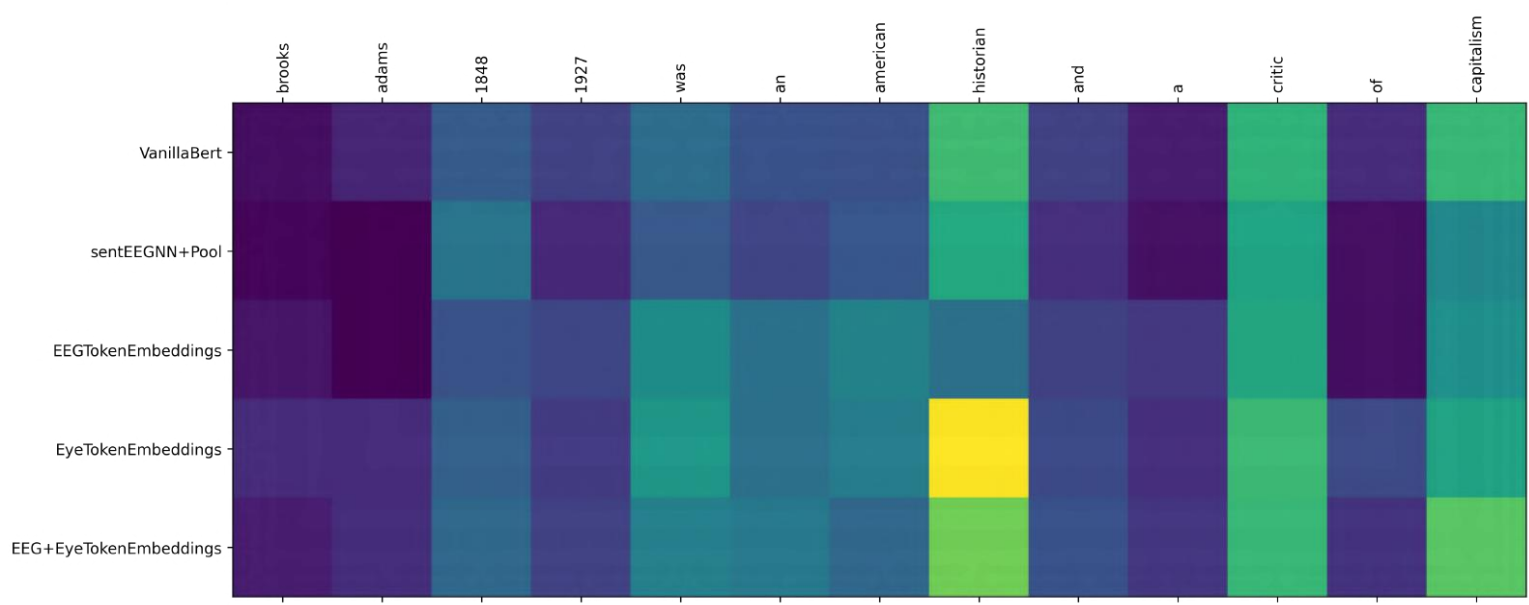

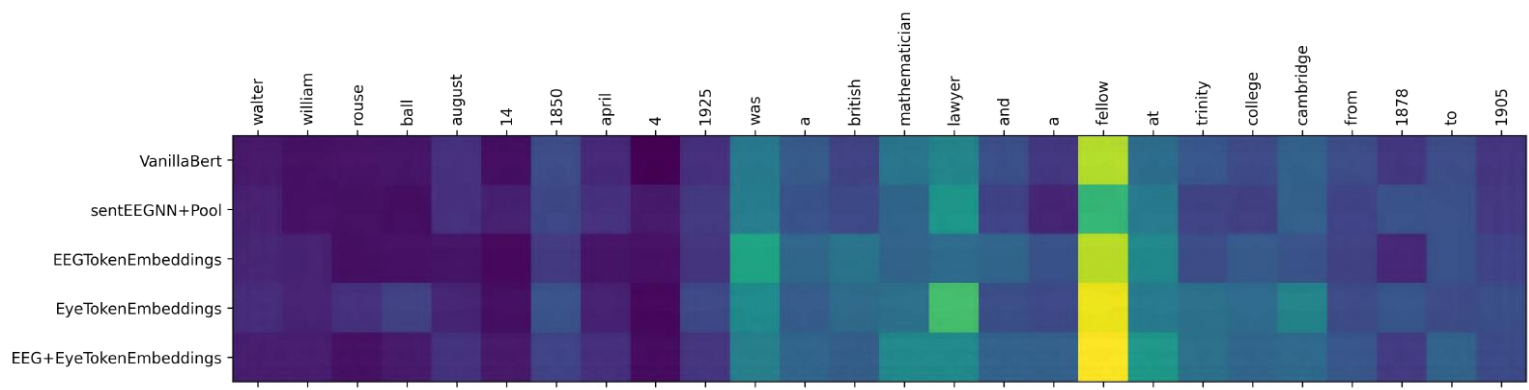

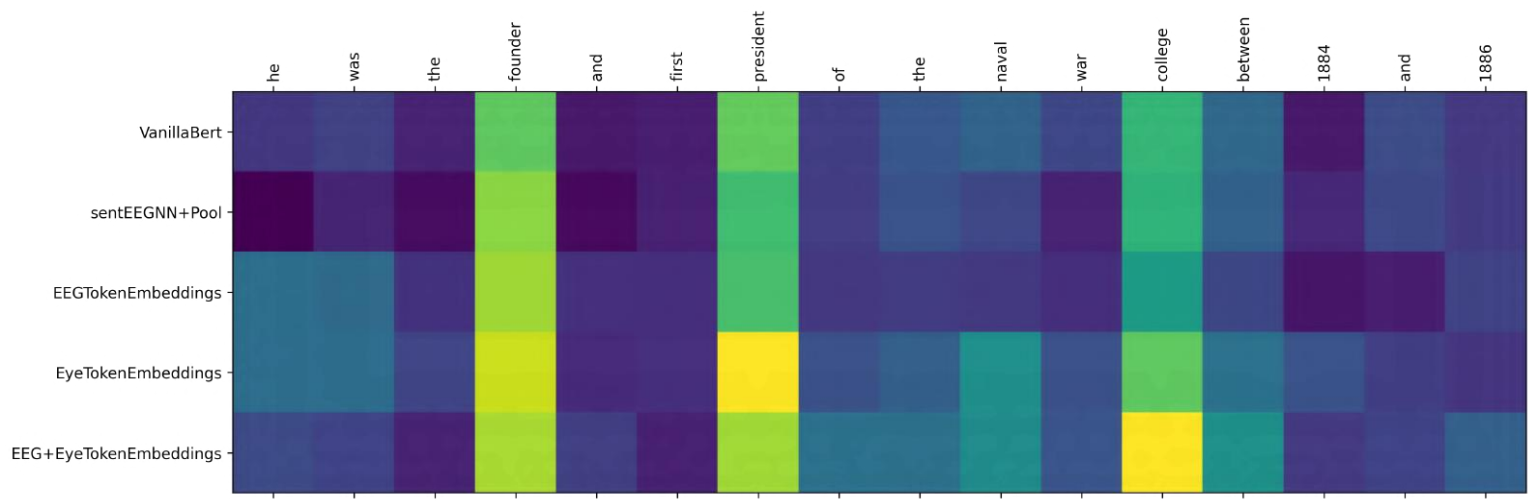

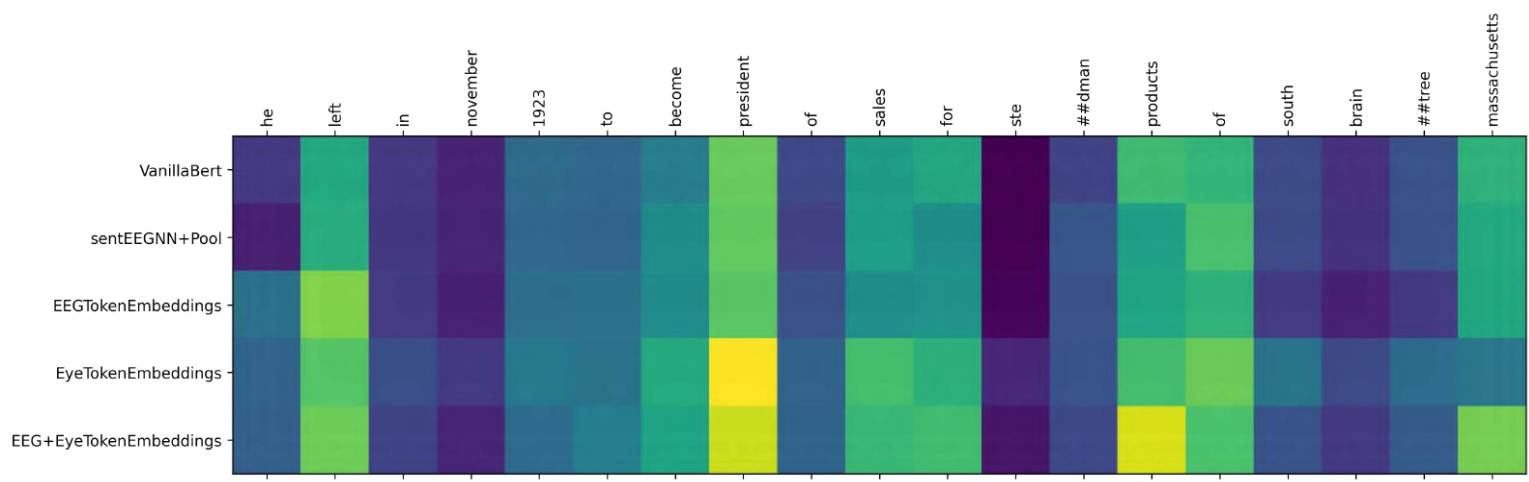

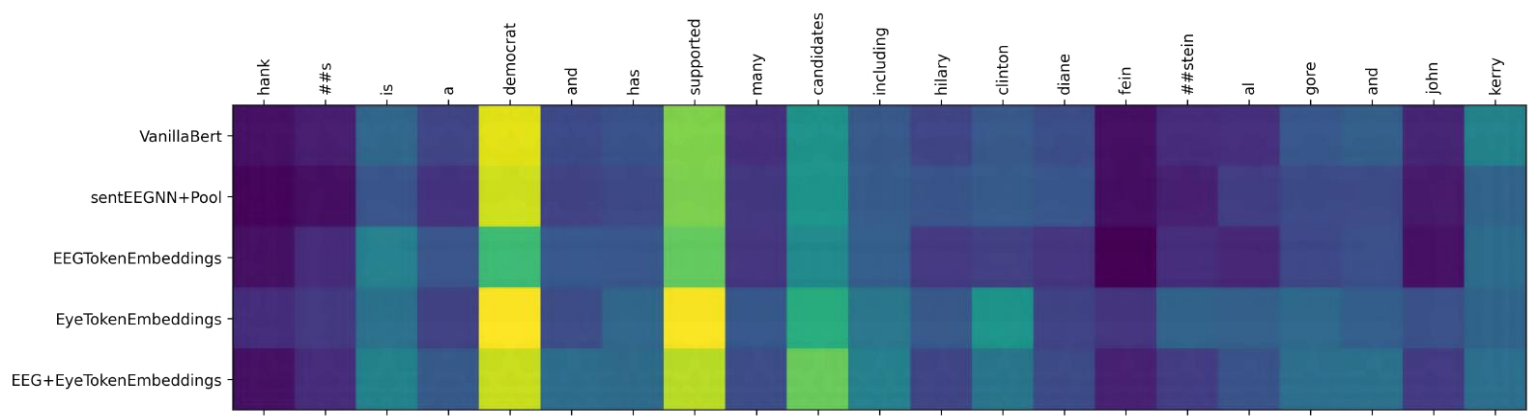

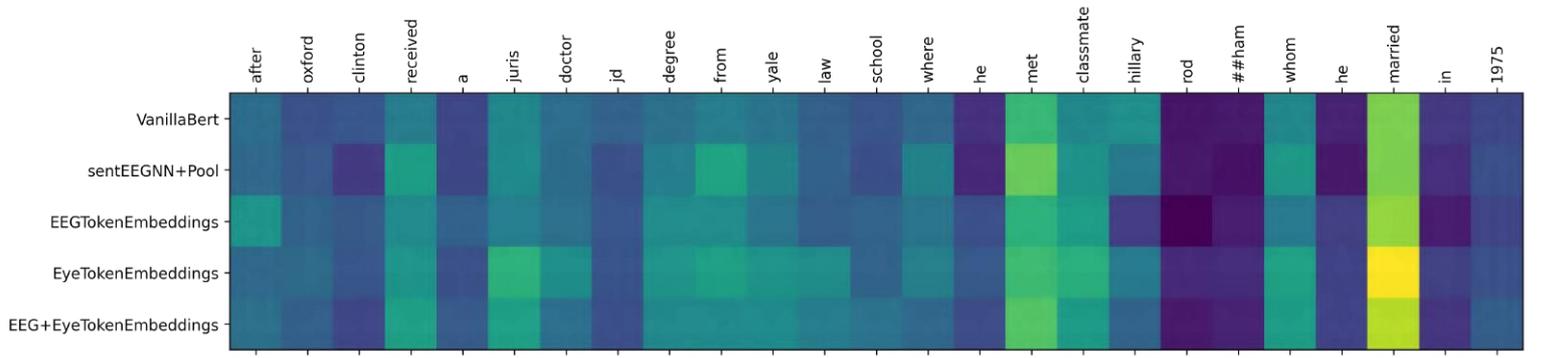

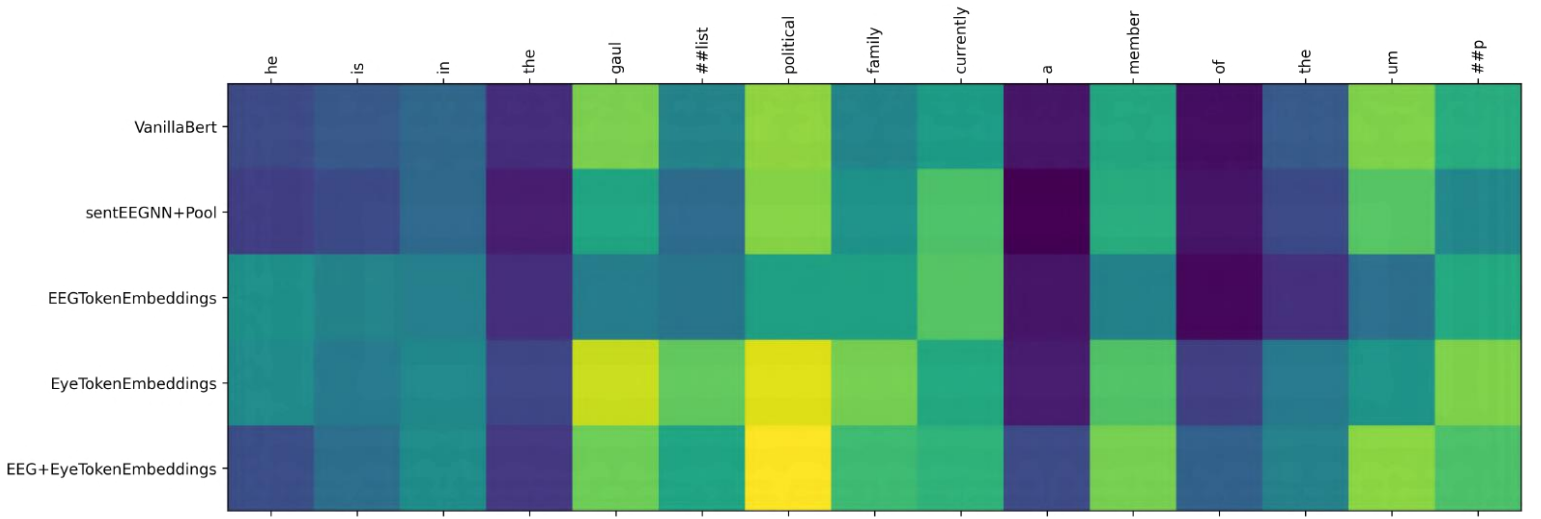

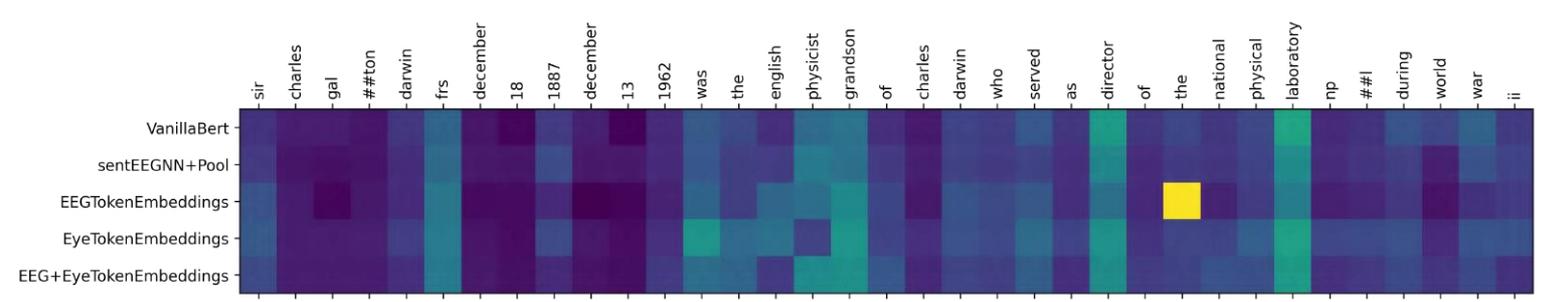

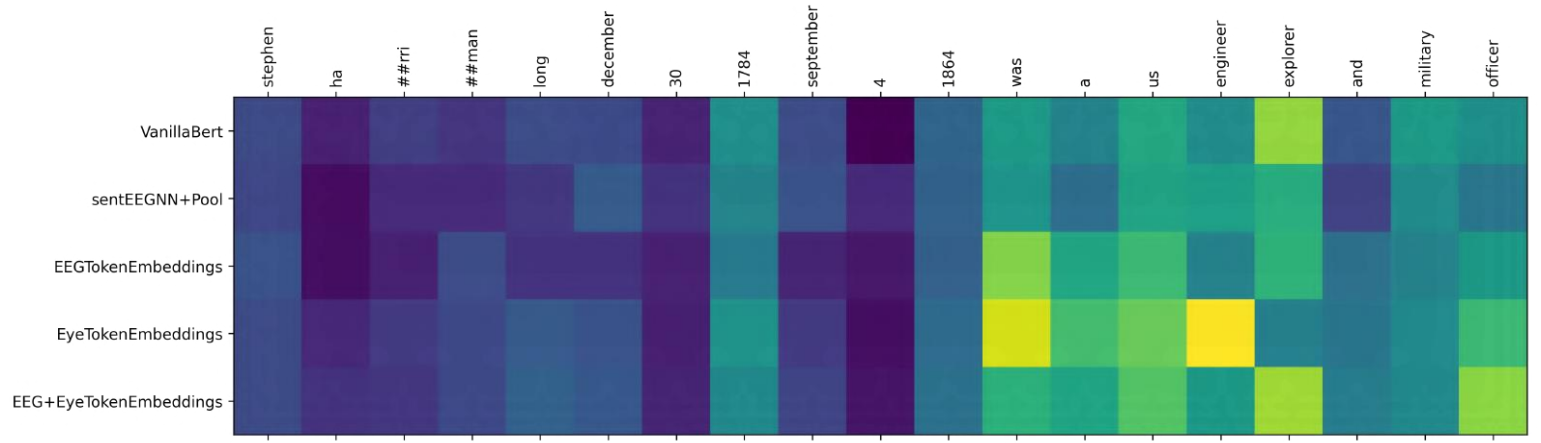

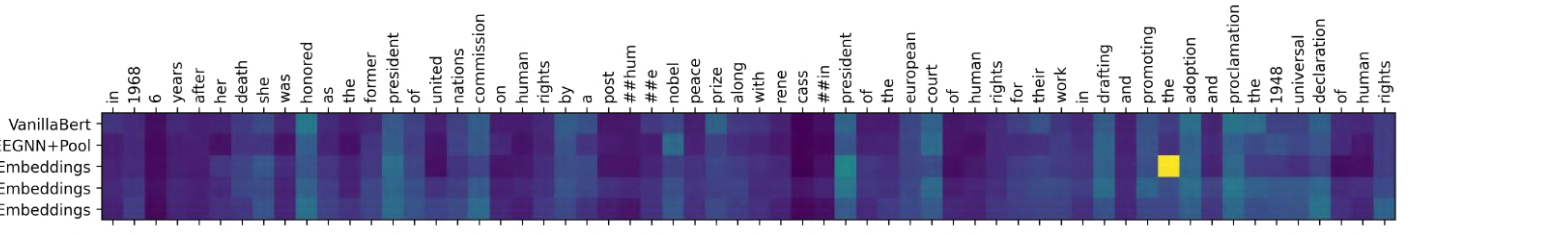

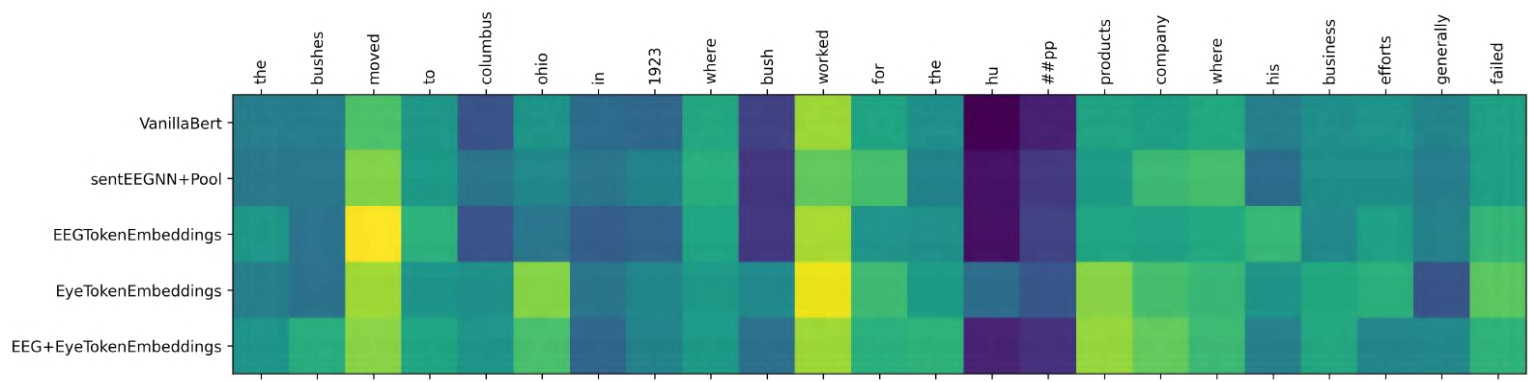

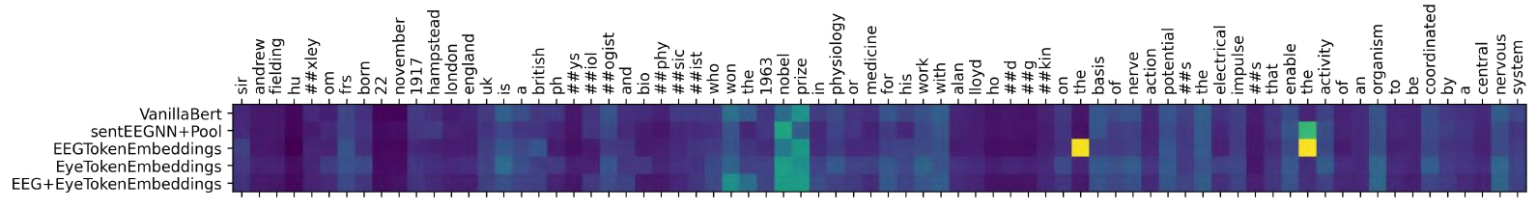

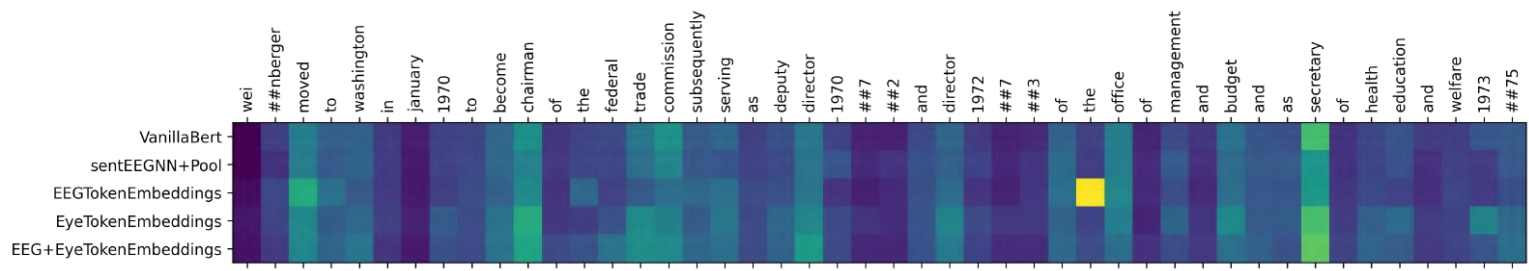

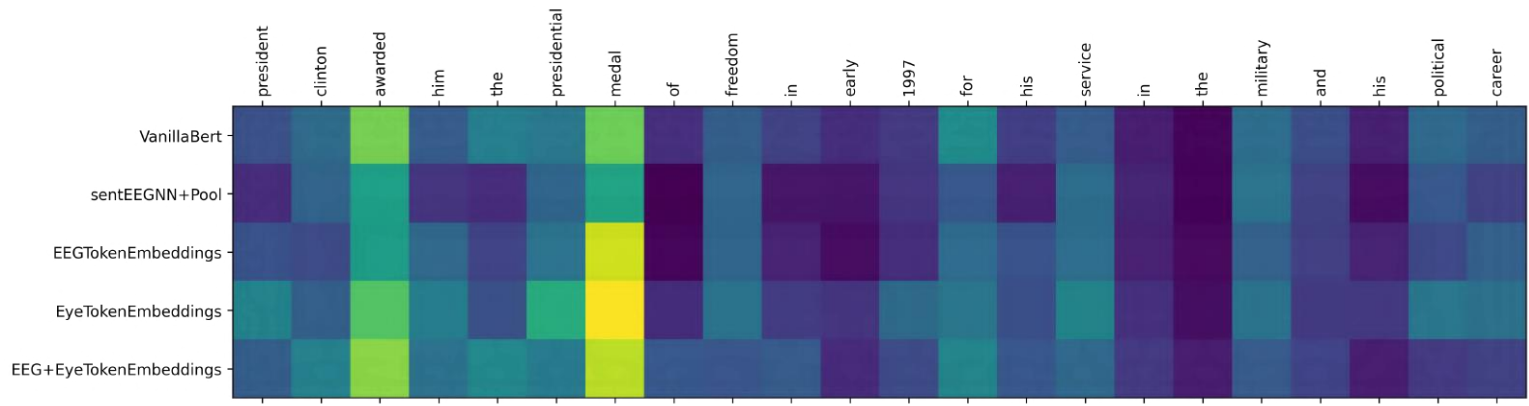

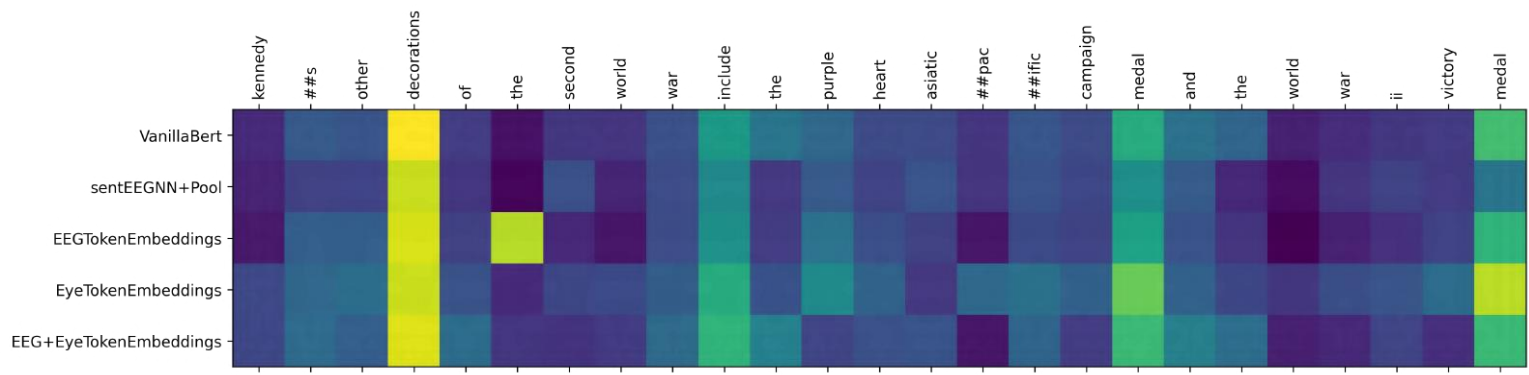

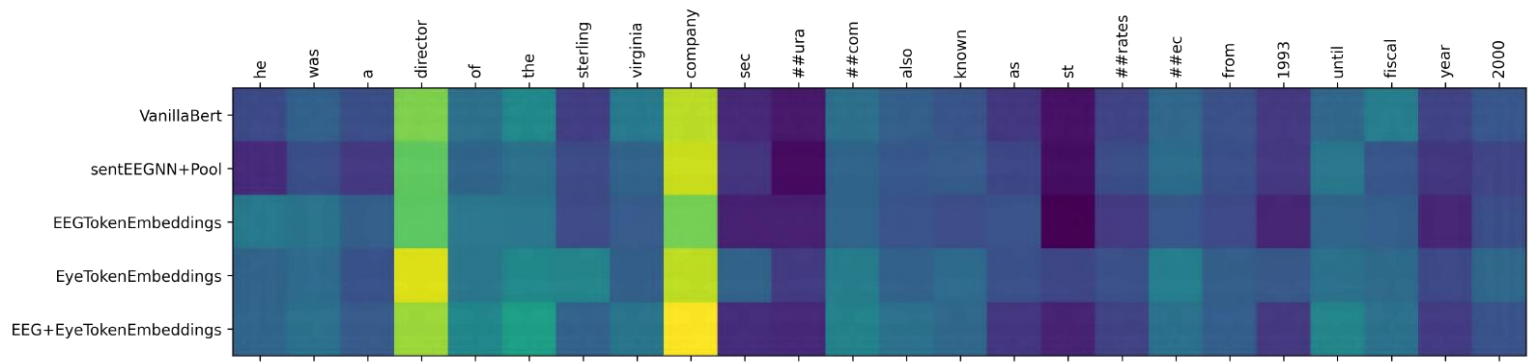

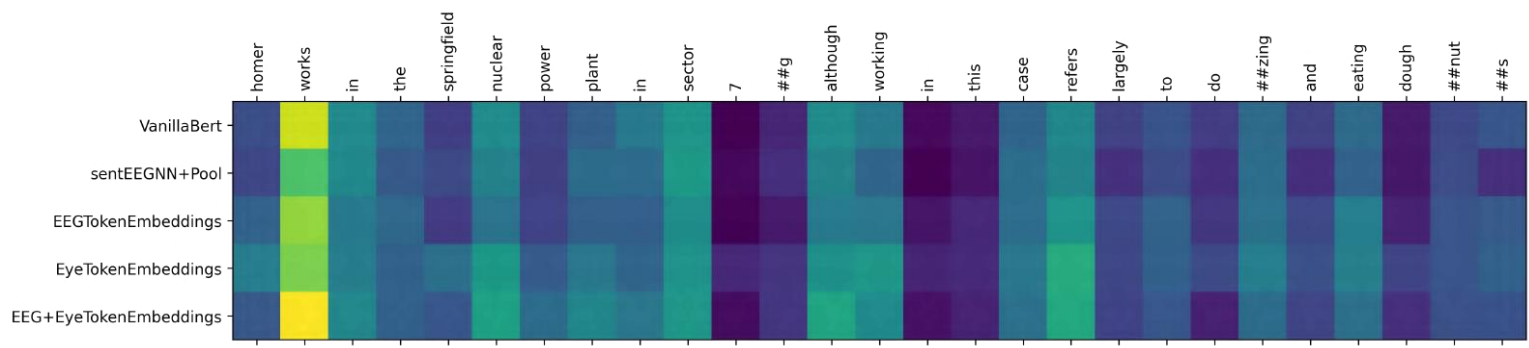
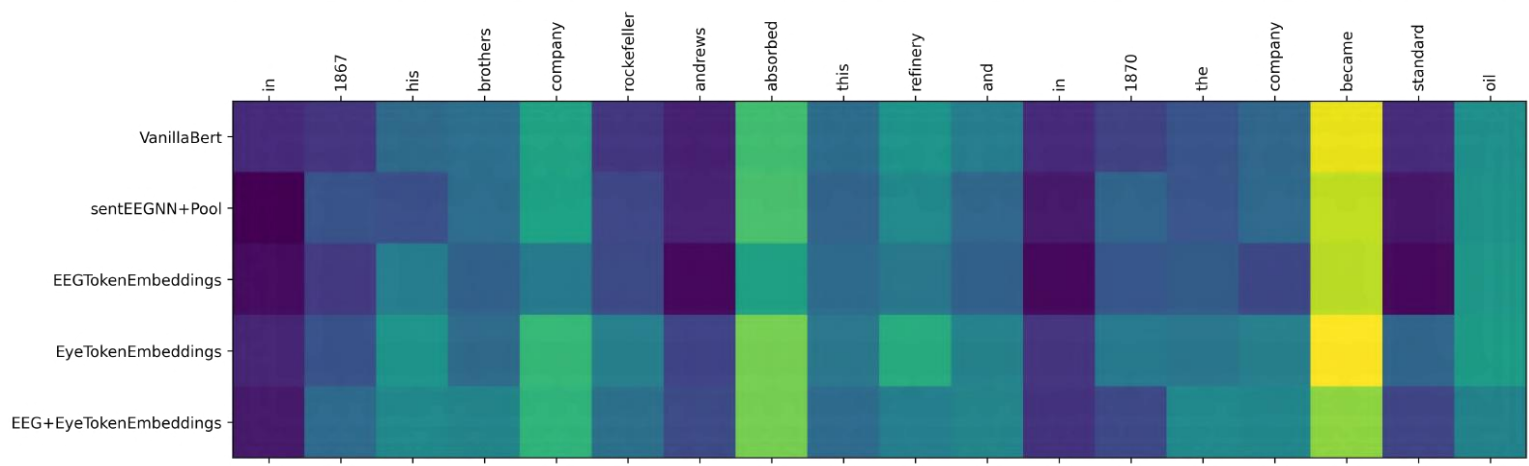
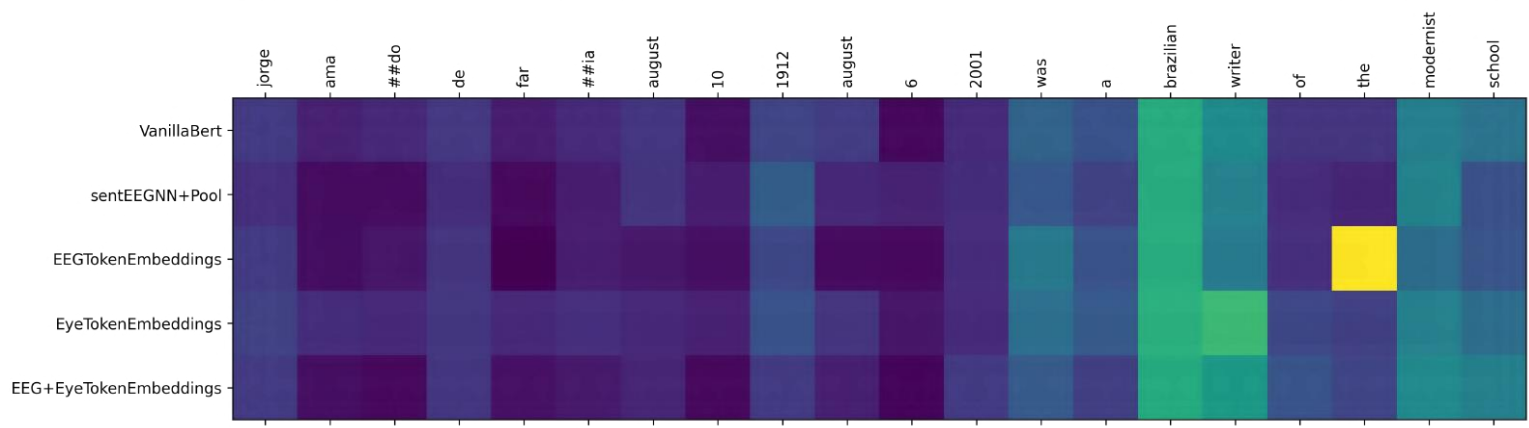
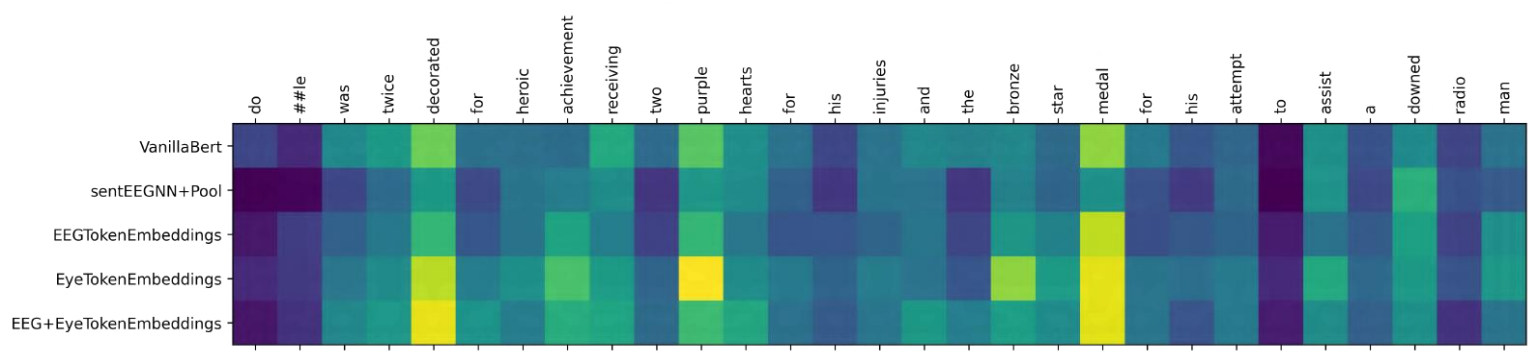
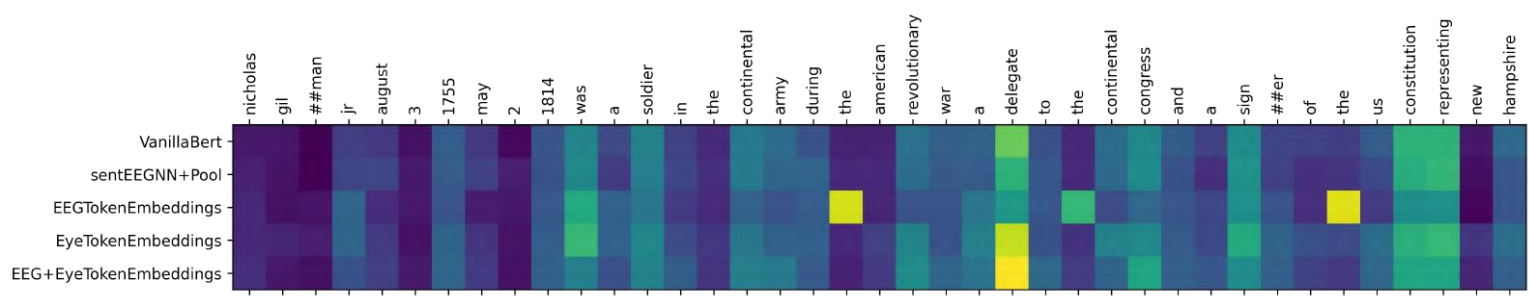
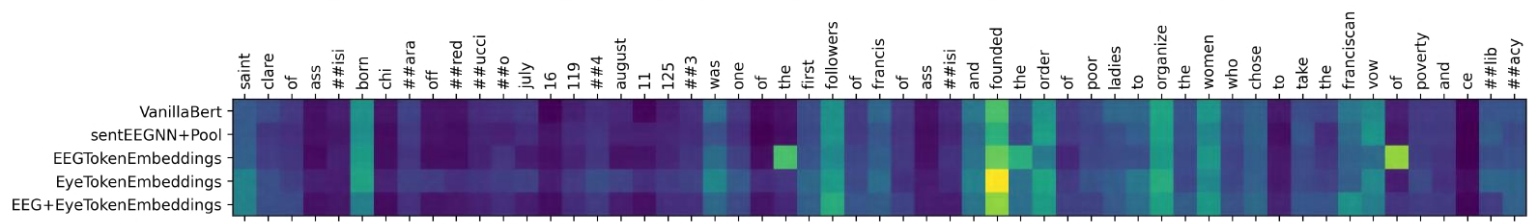

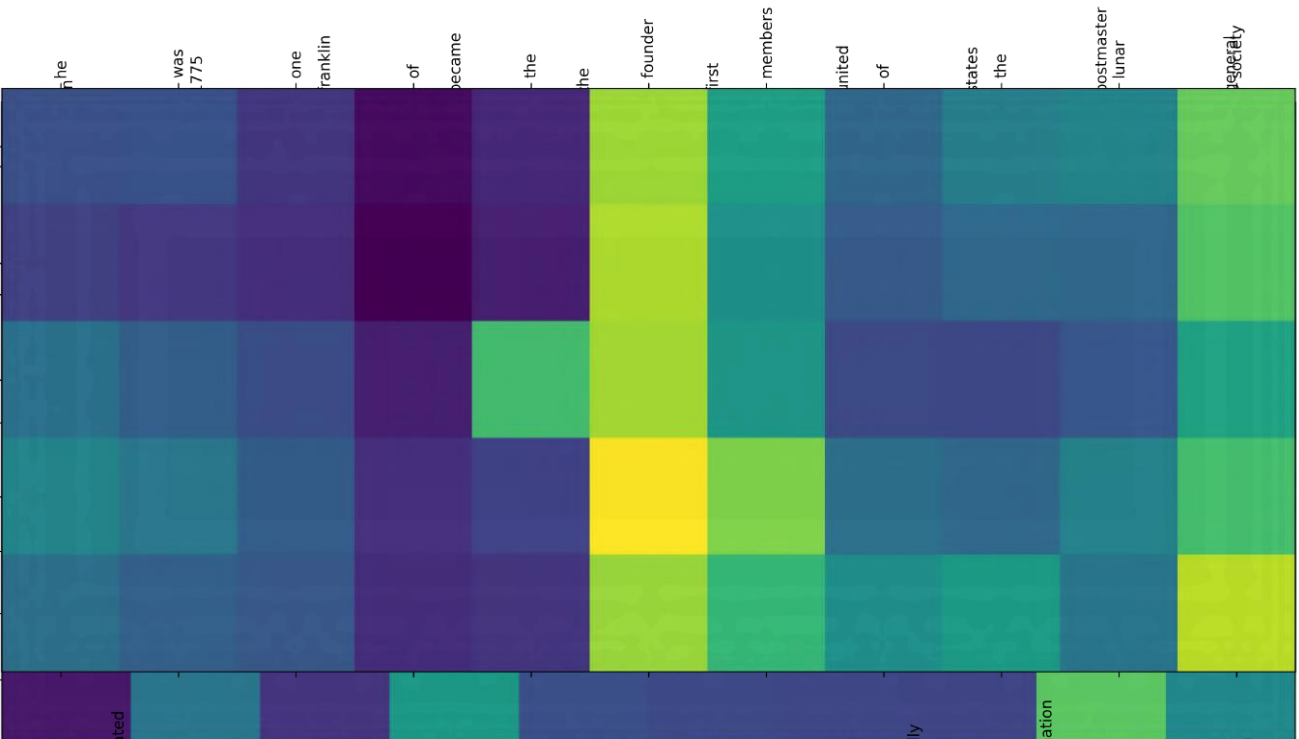

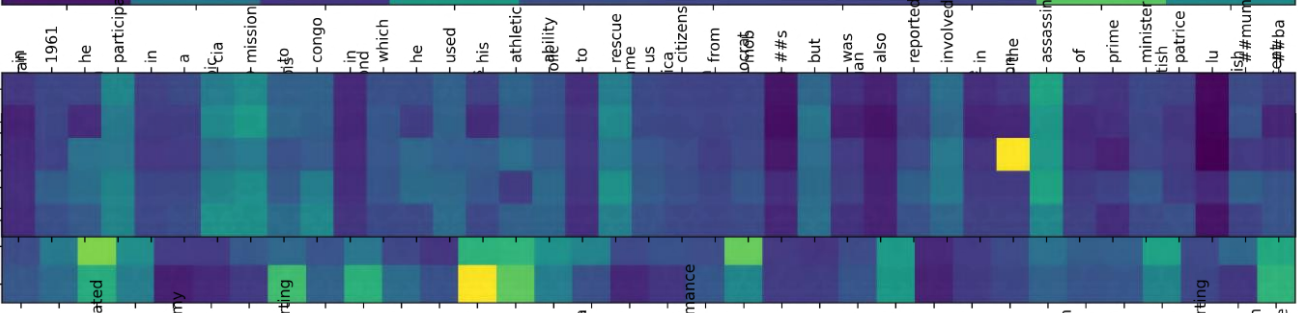

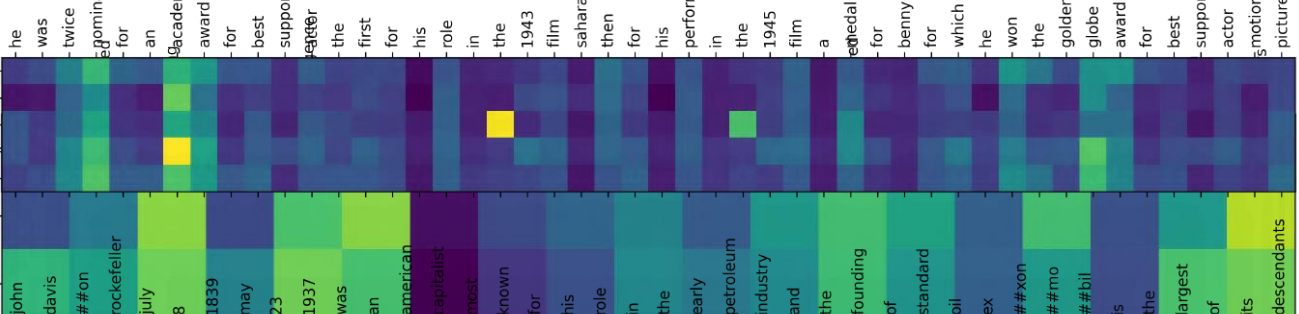

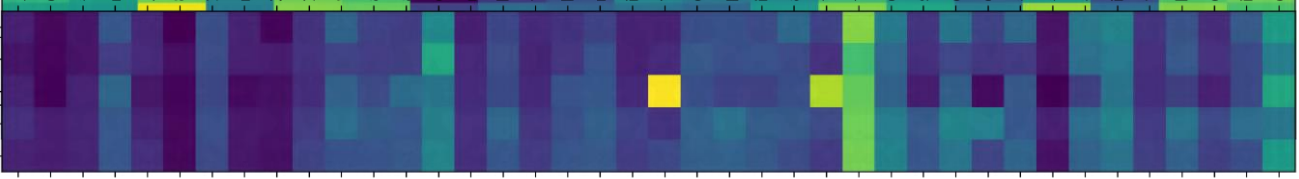

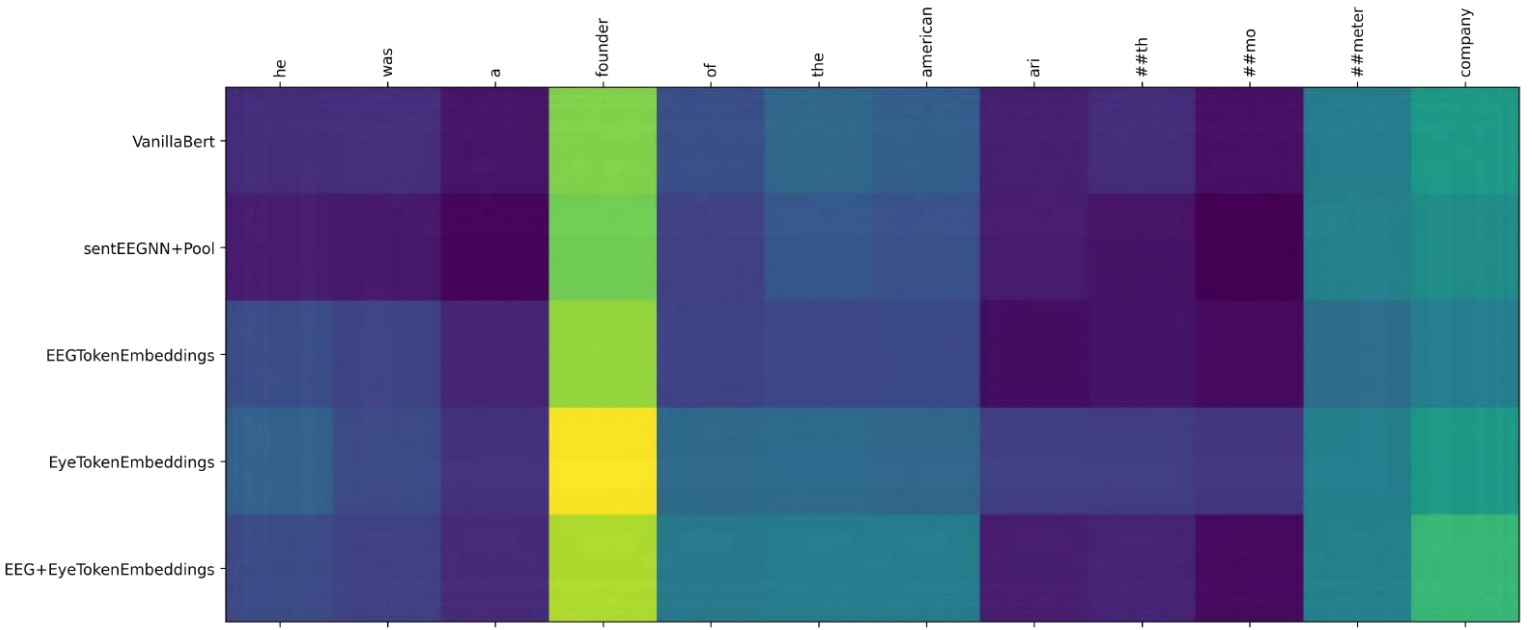

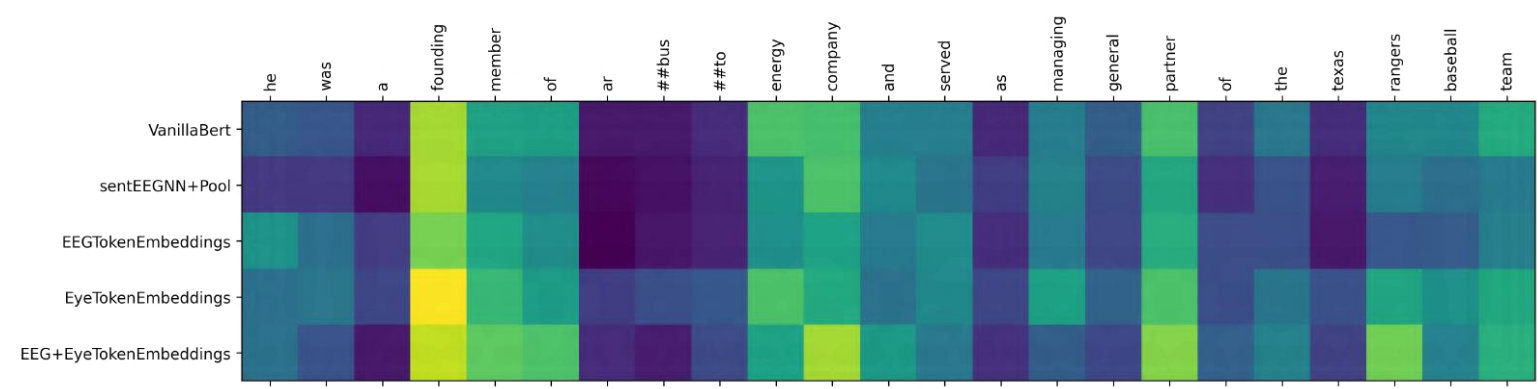

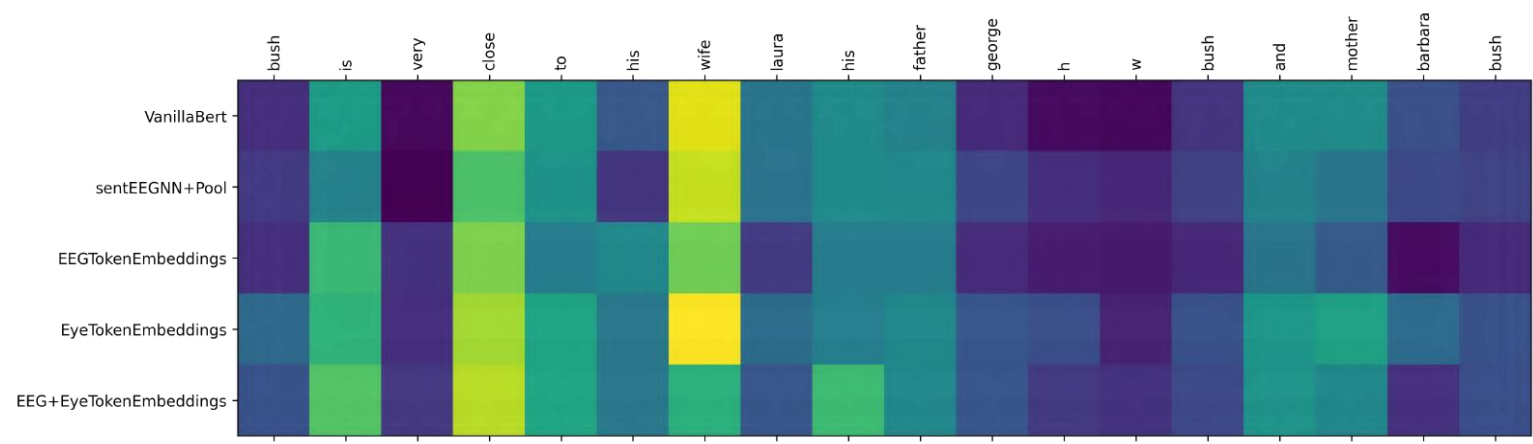

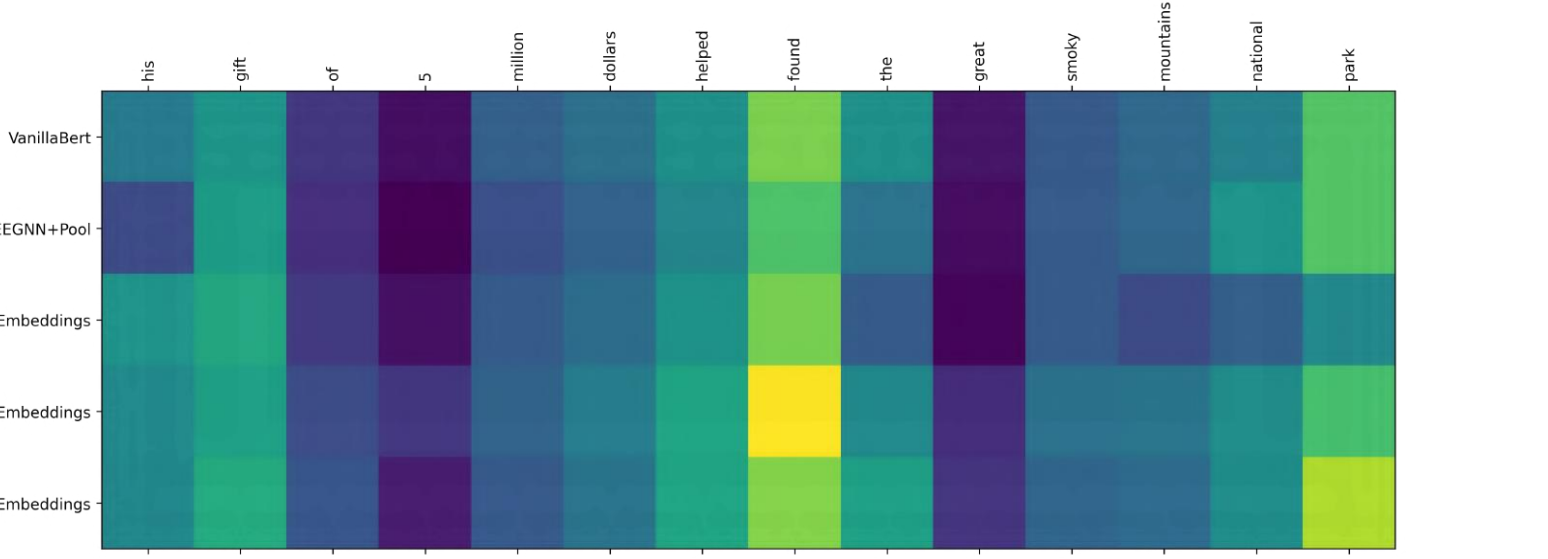

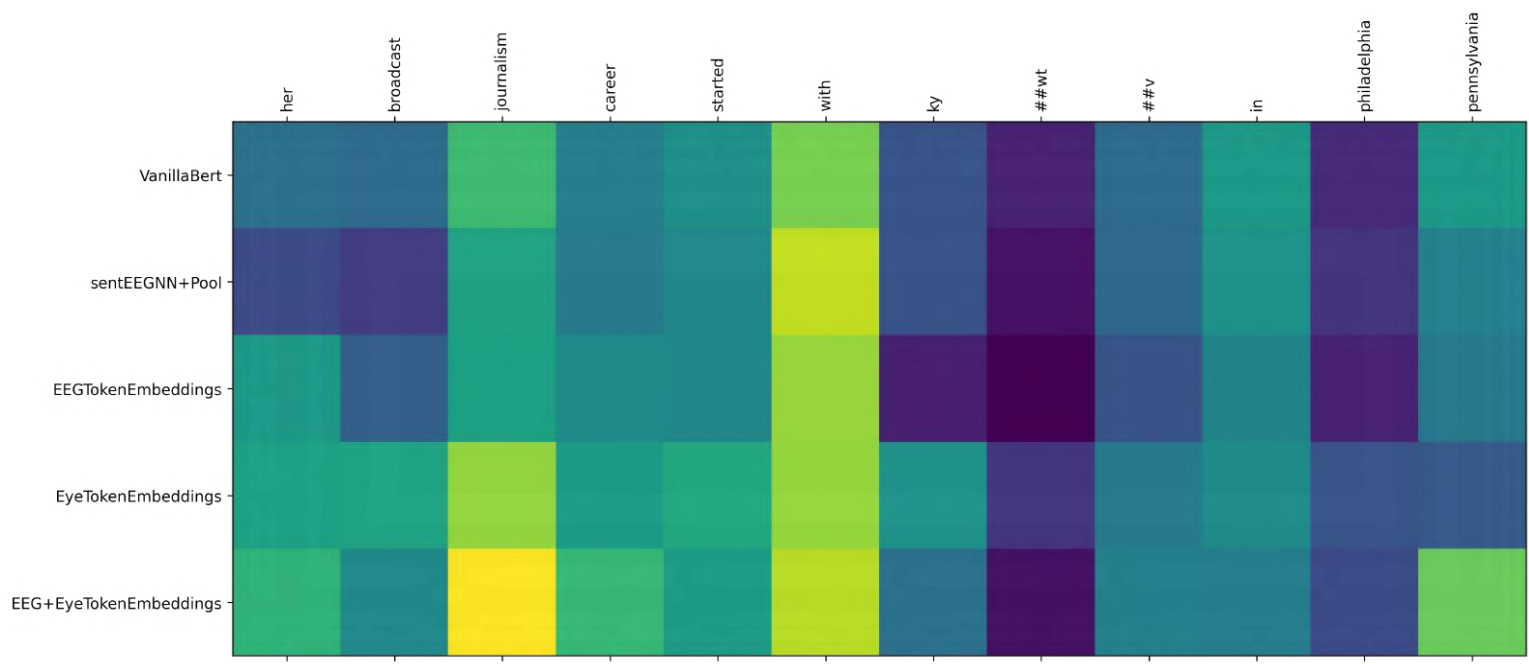

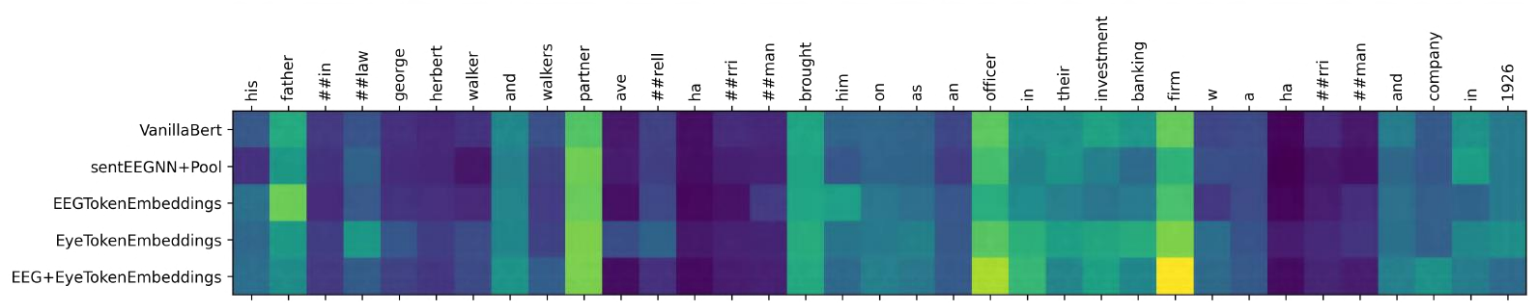

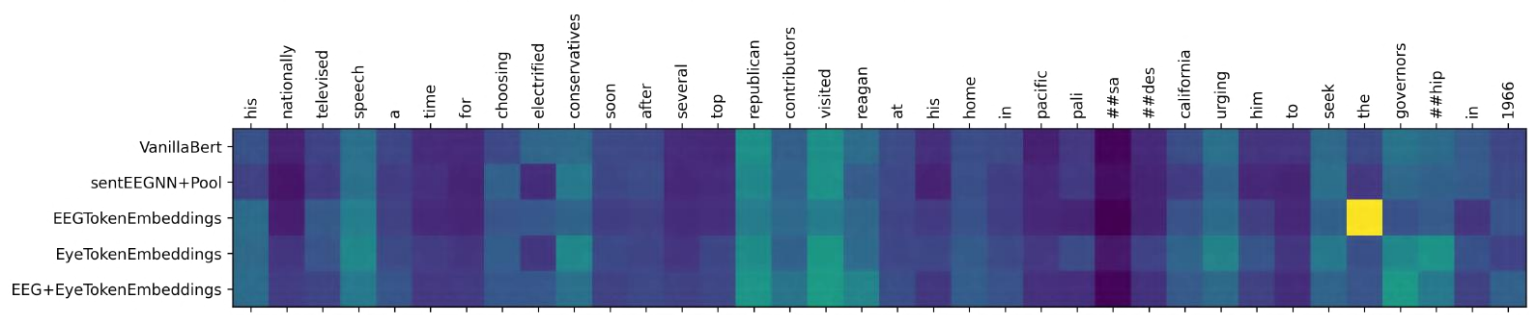

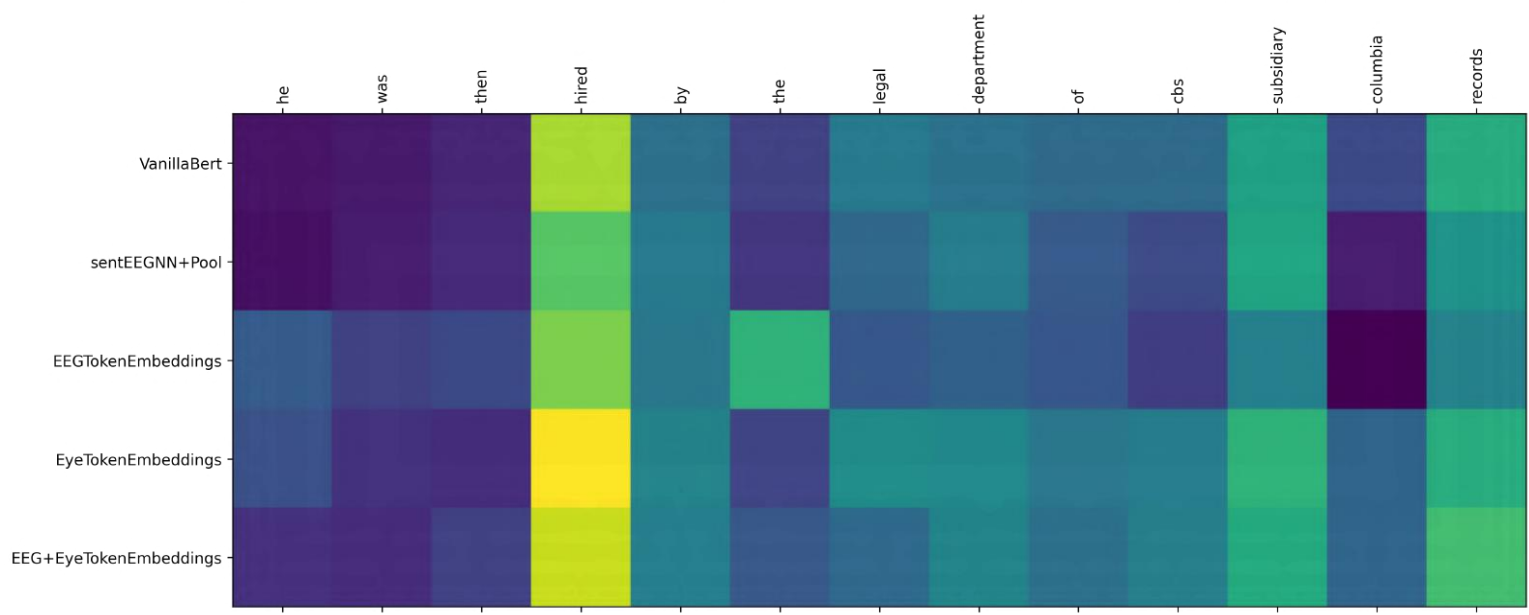

# Appendix C: Heat maps for incoming self-attention weights for the test corpus of the cognitive attention mask model

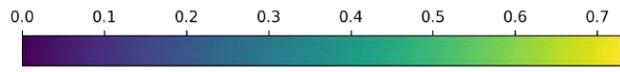

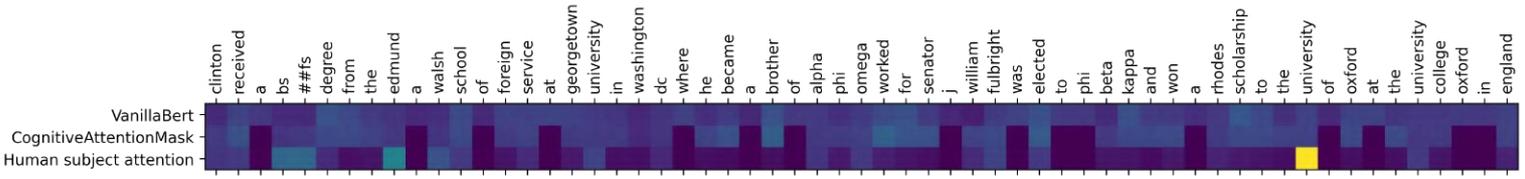

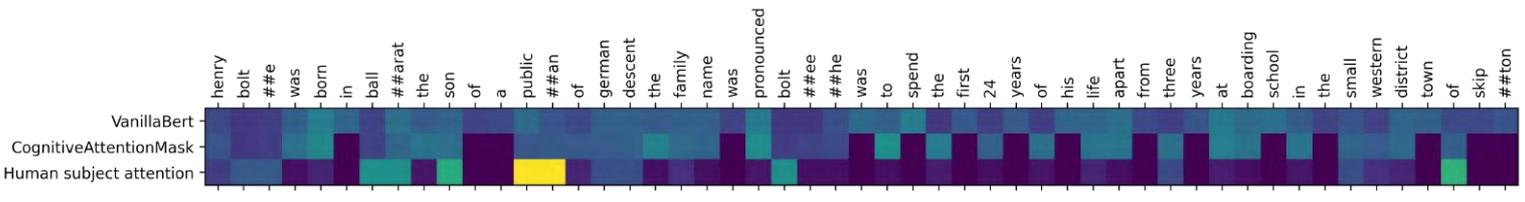

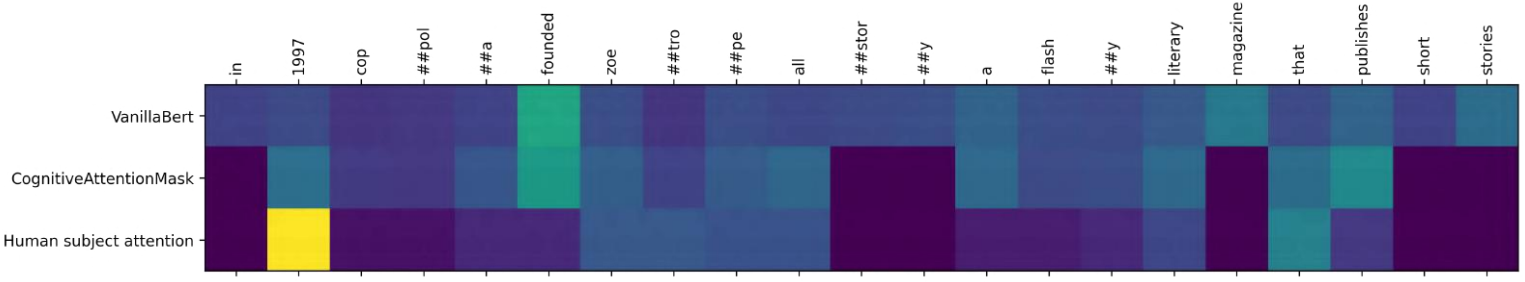

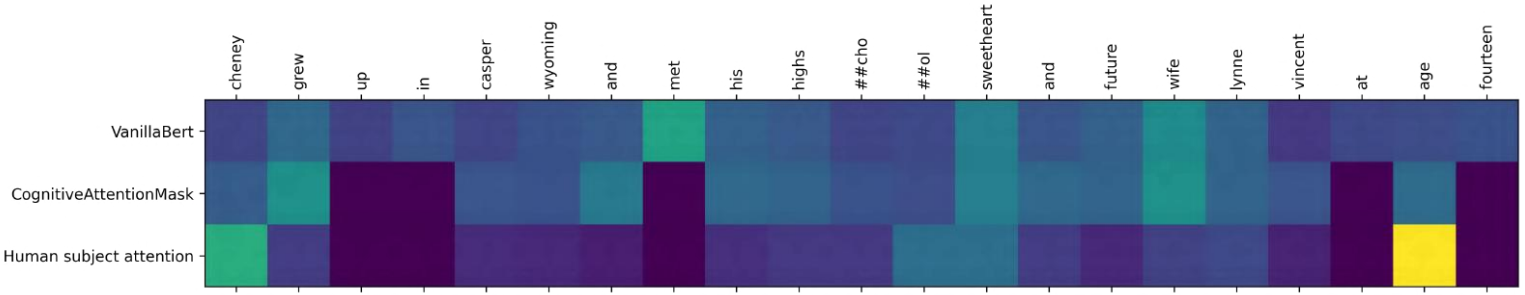

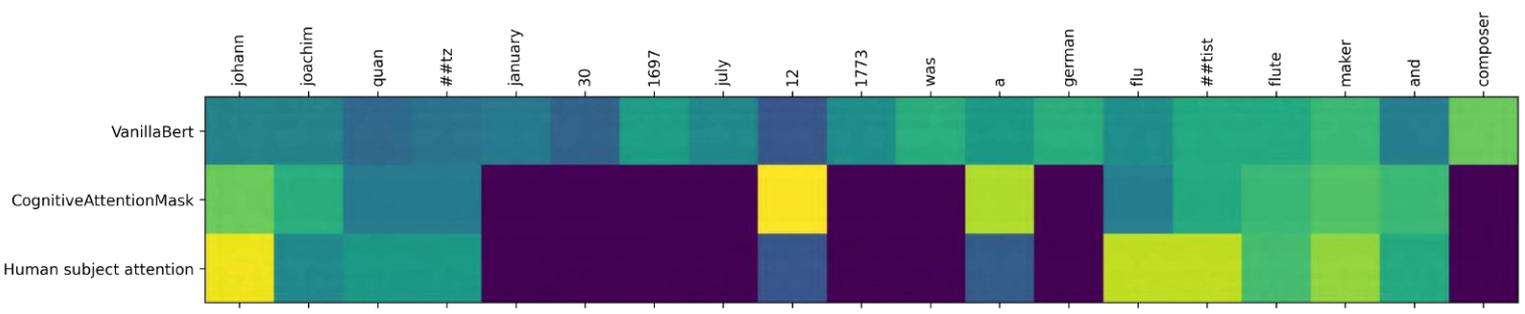

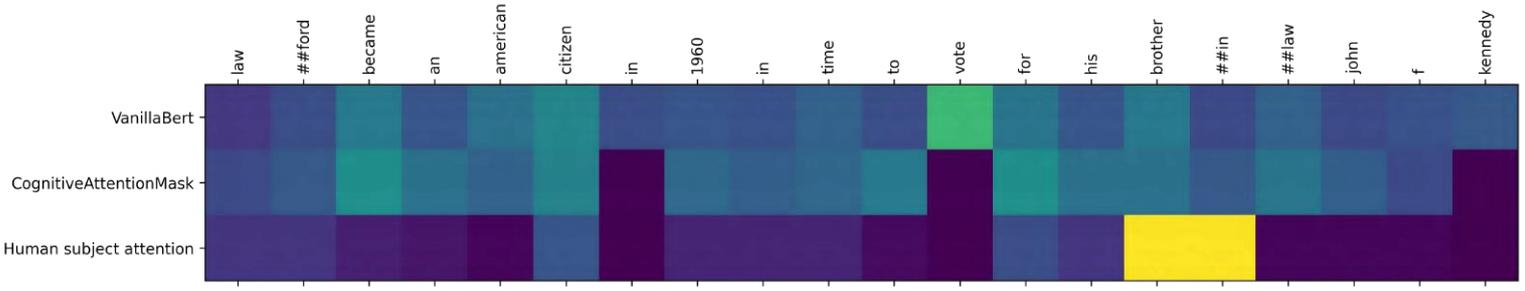

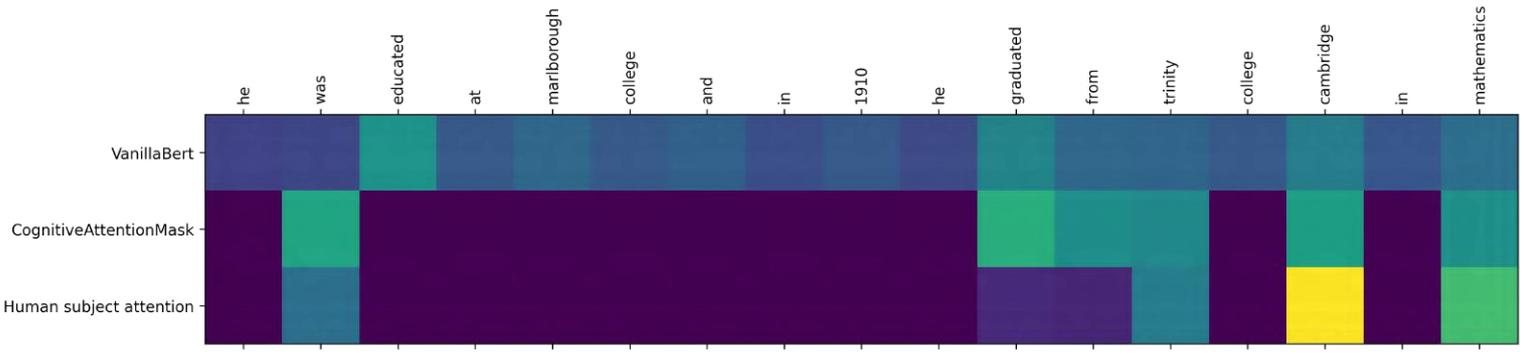

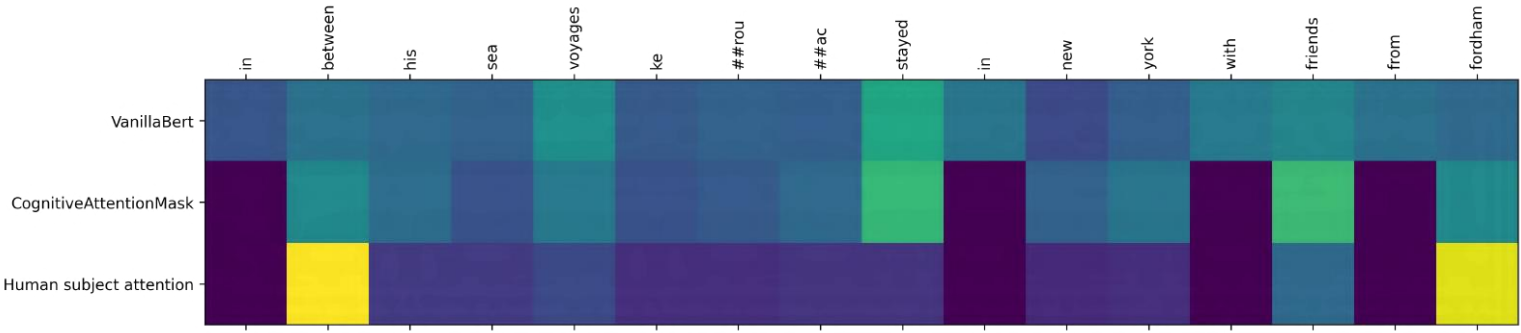

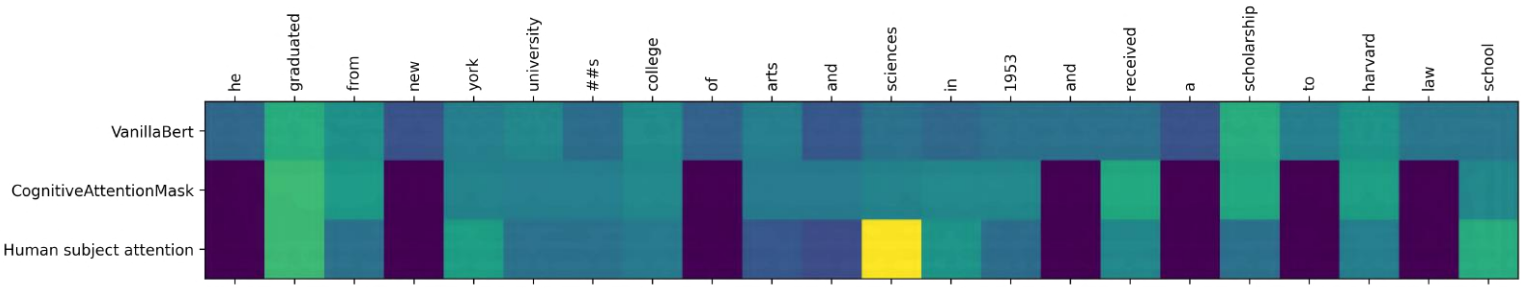

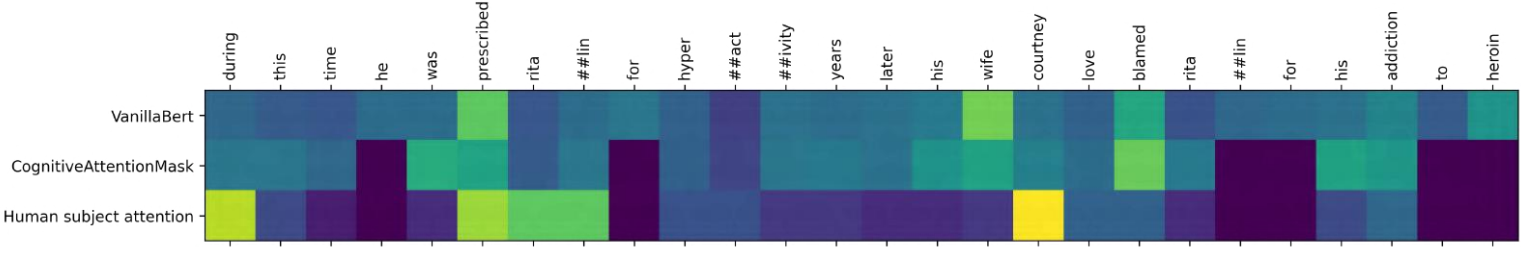

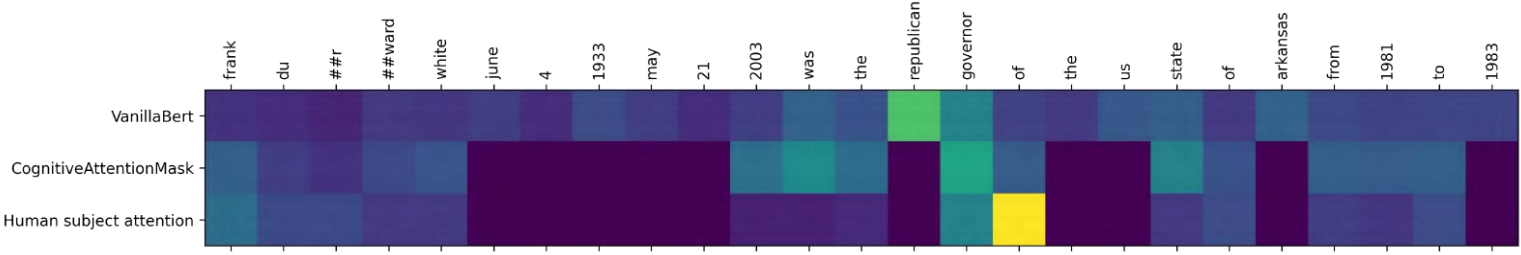

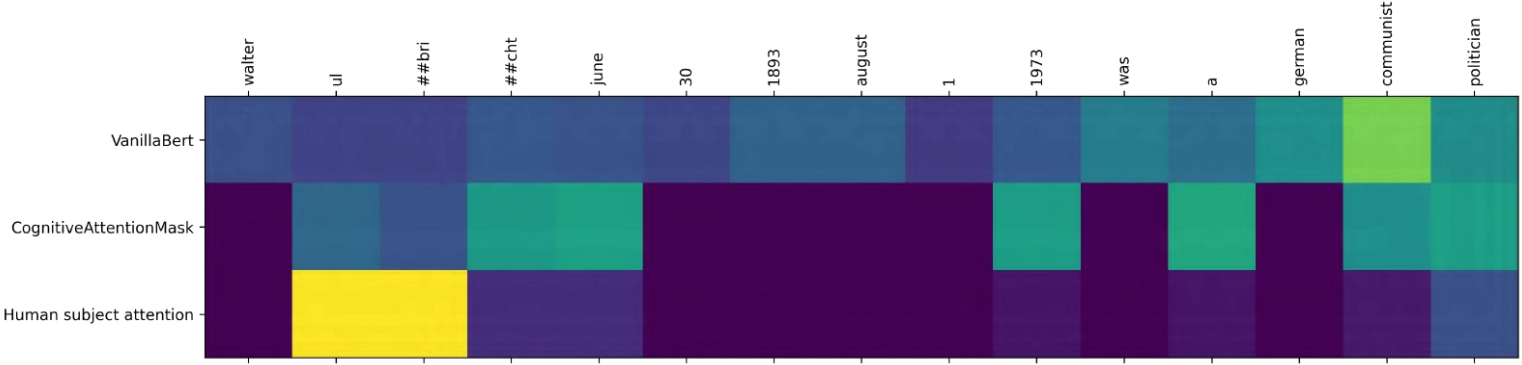

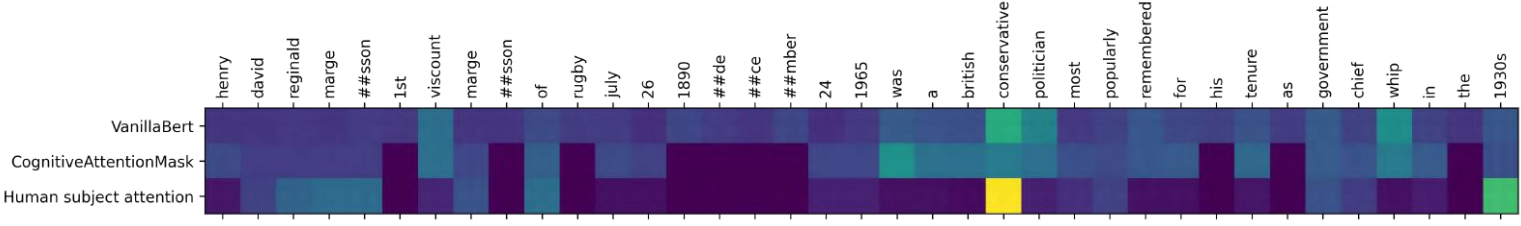

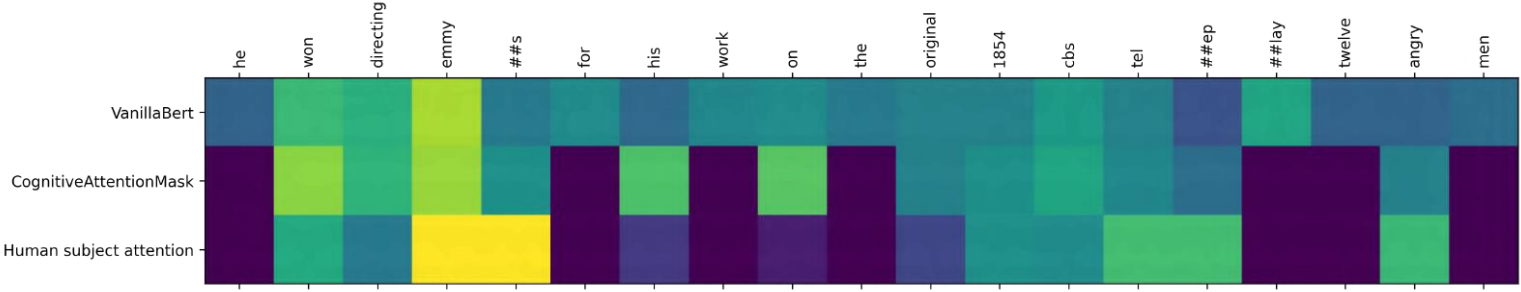

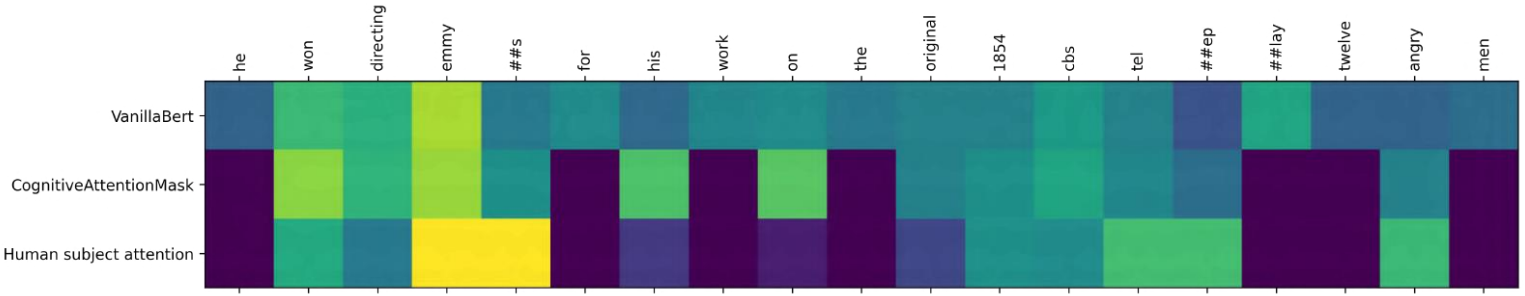

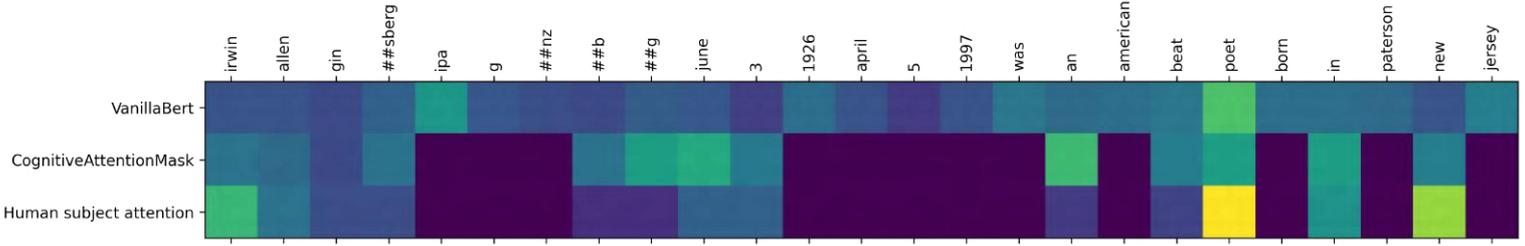

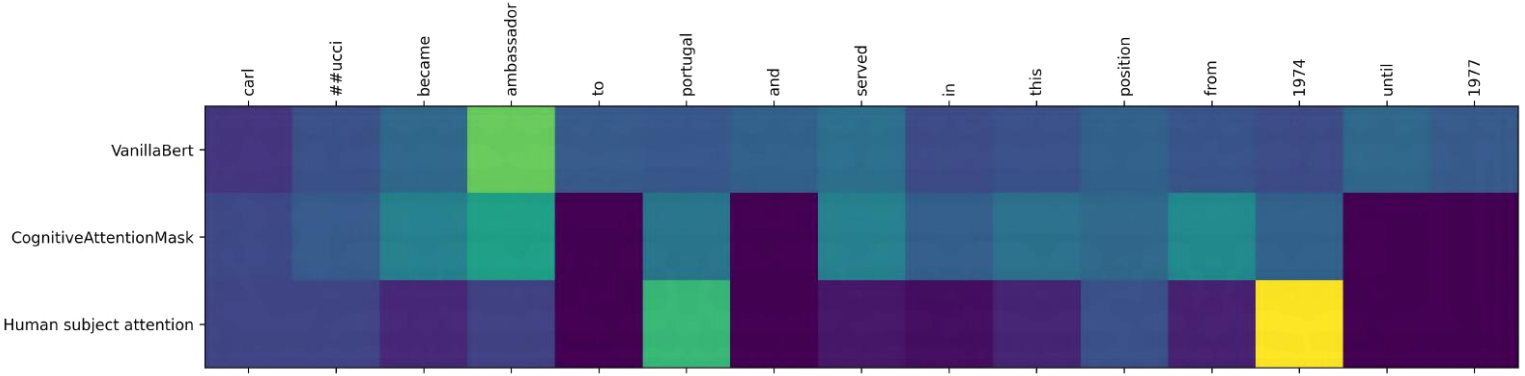

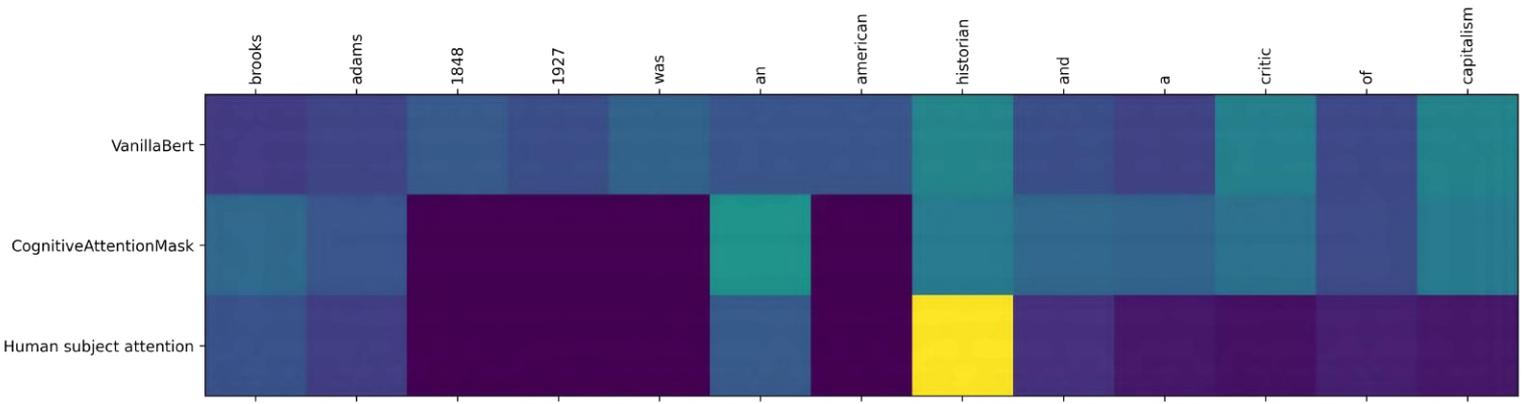

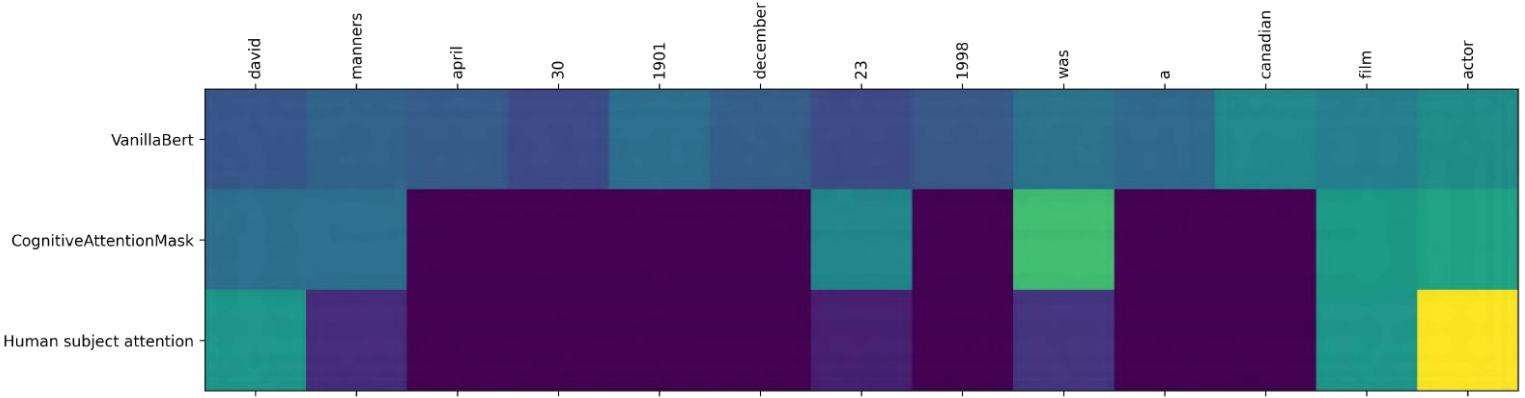

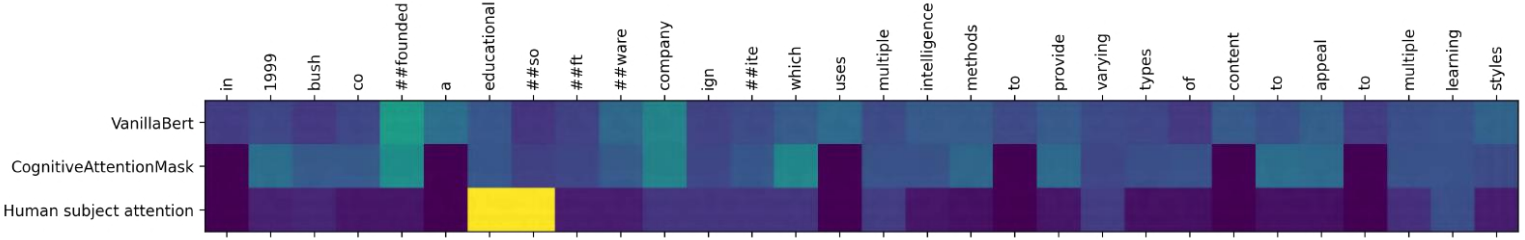

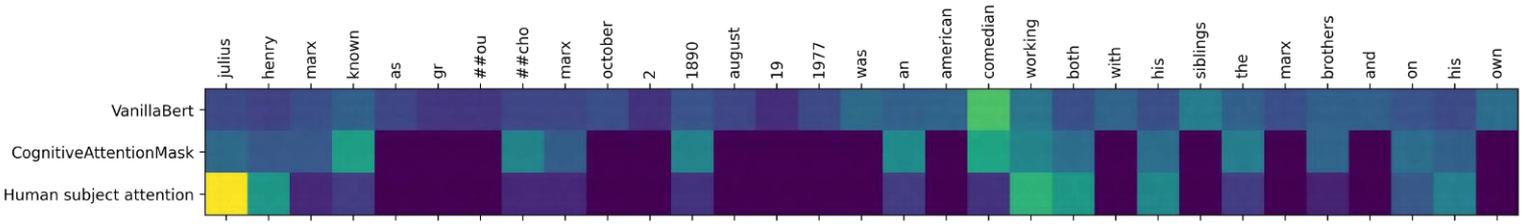

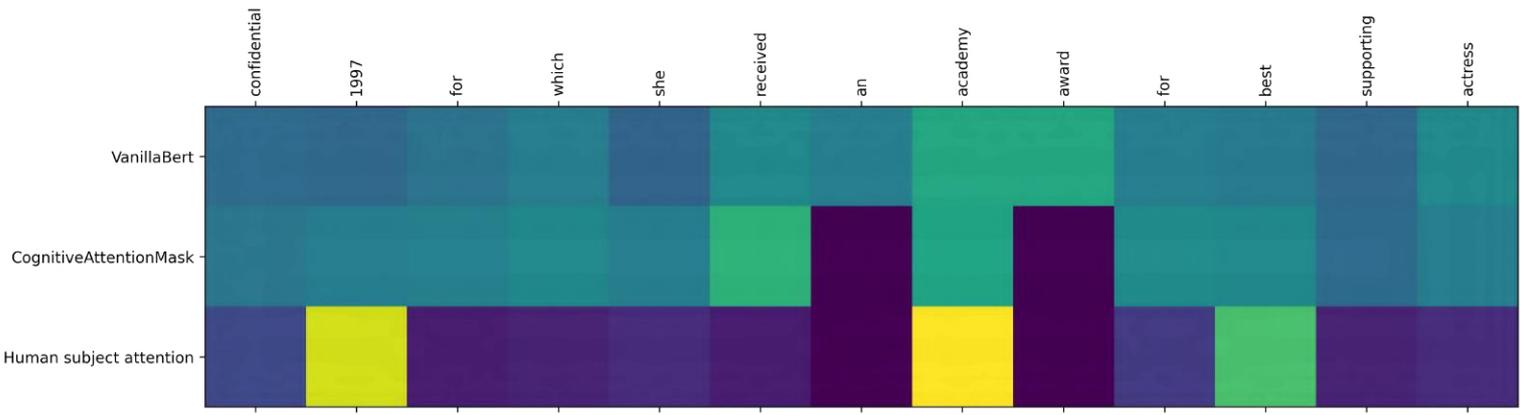

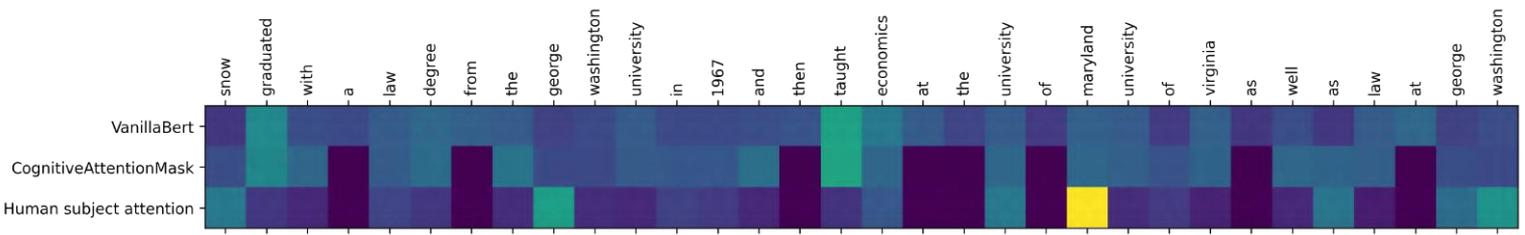

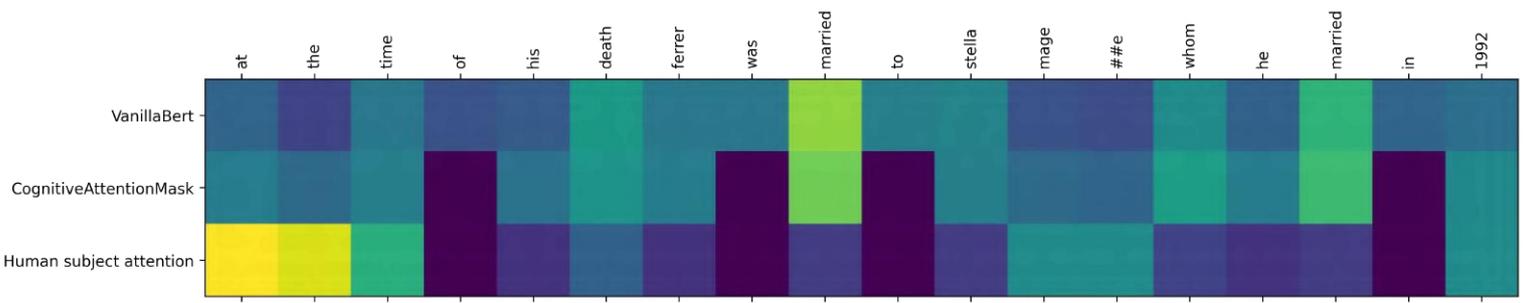

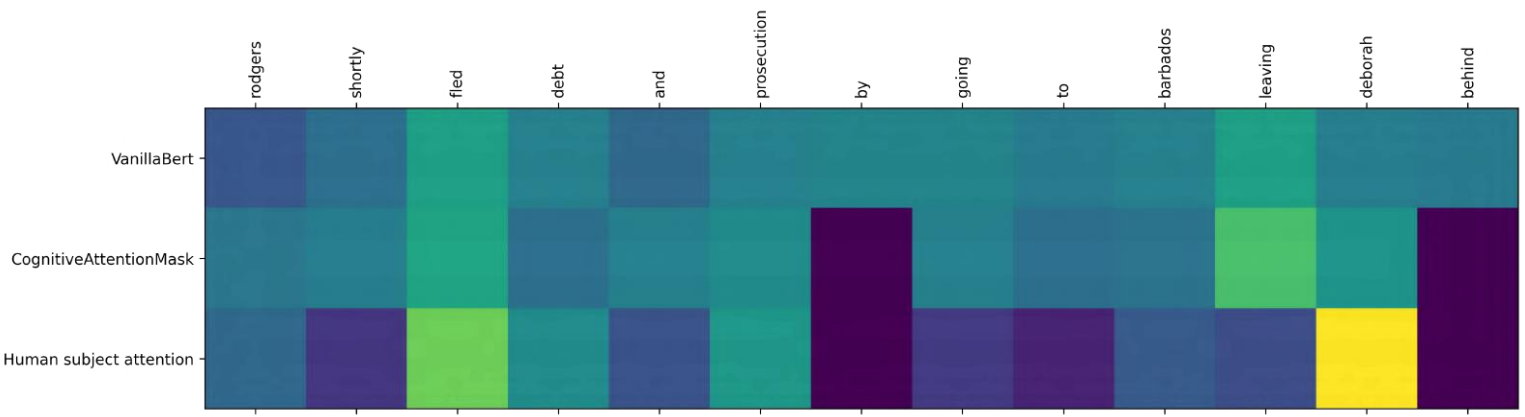

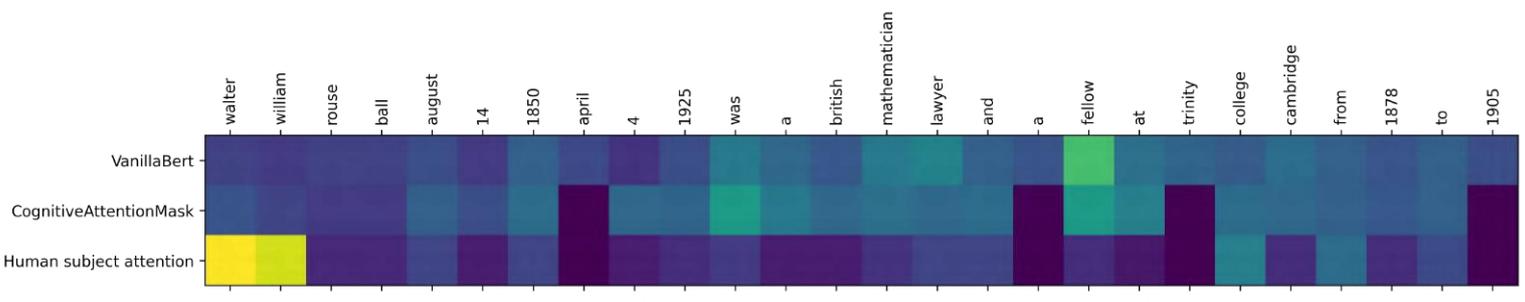

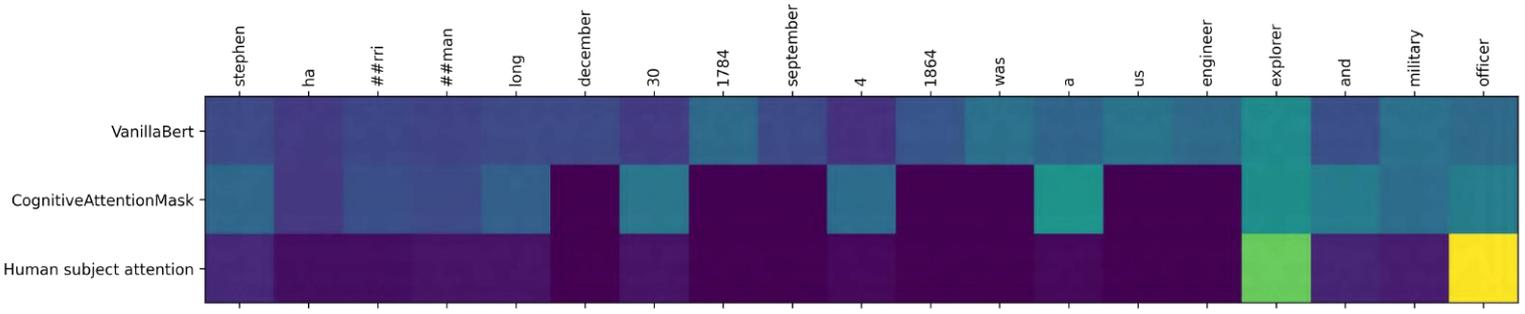

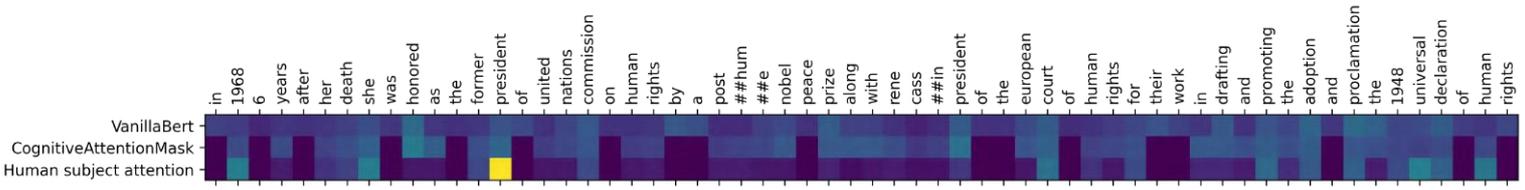

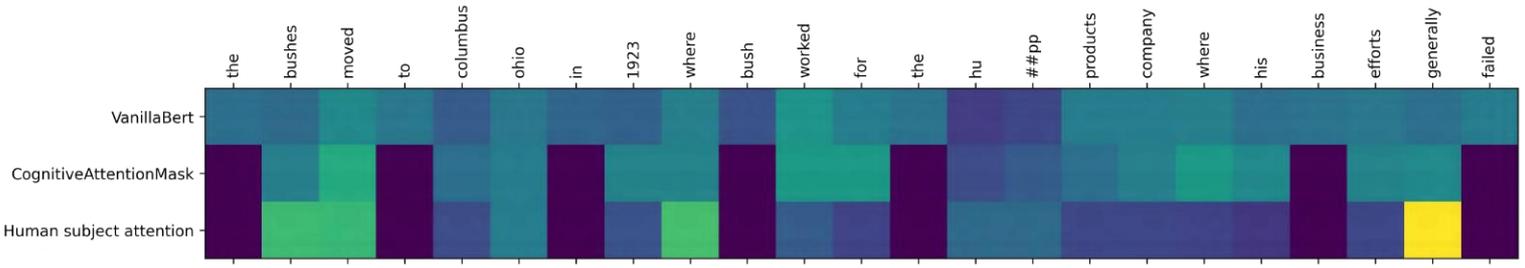

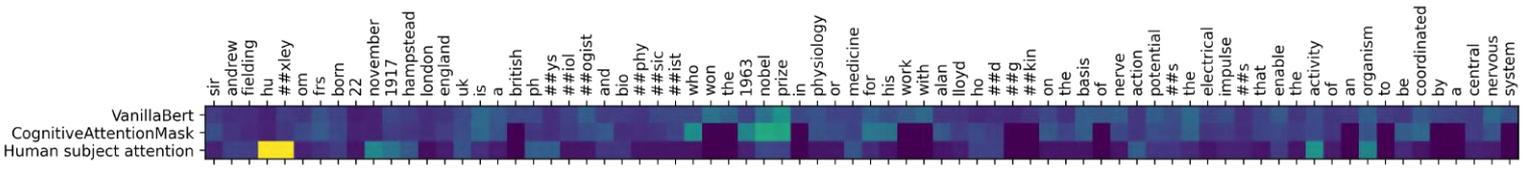

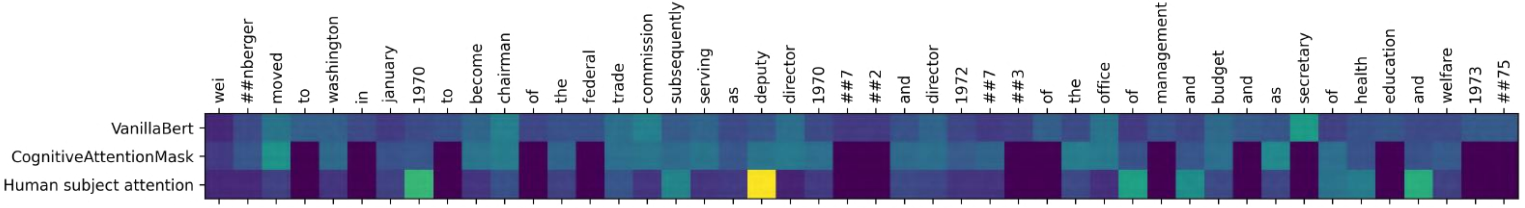

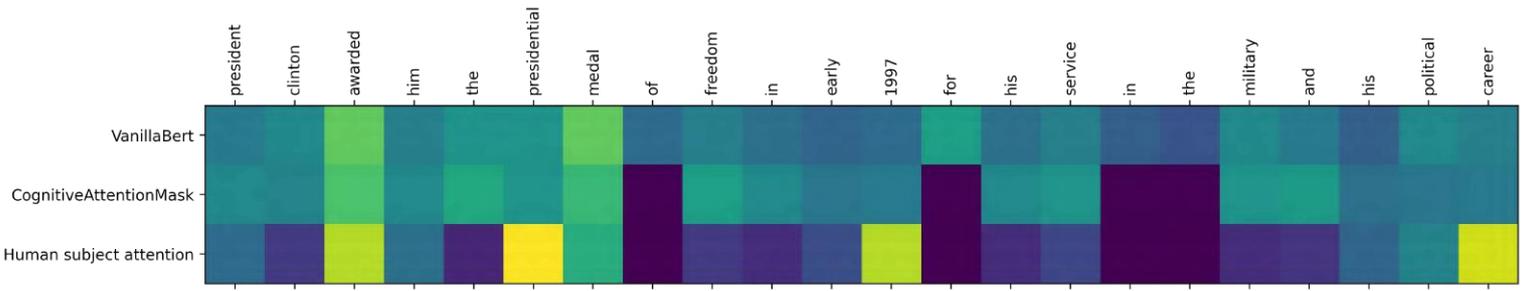

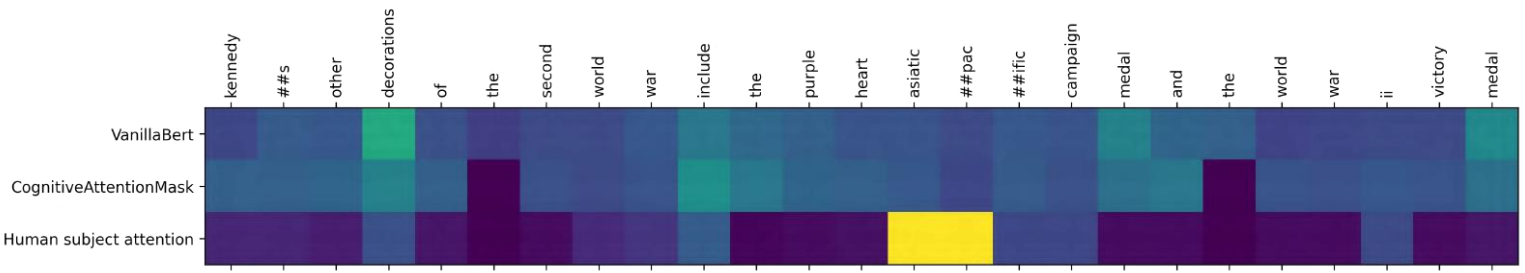

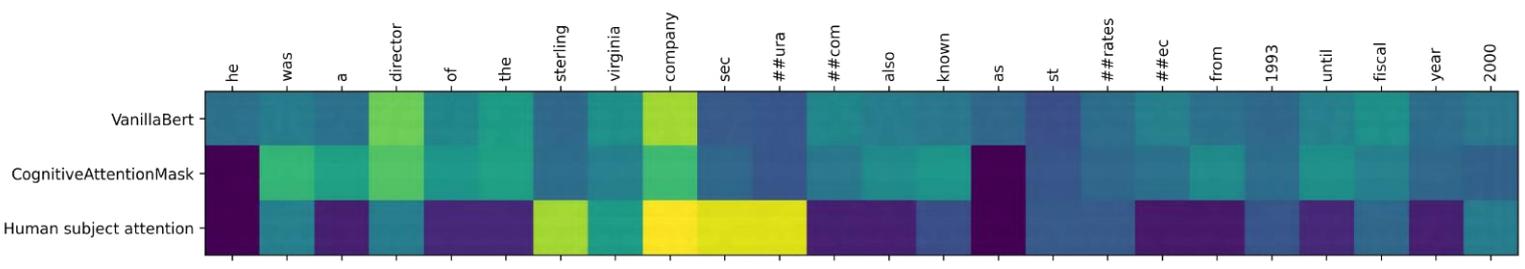

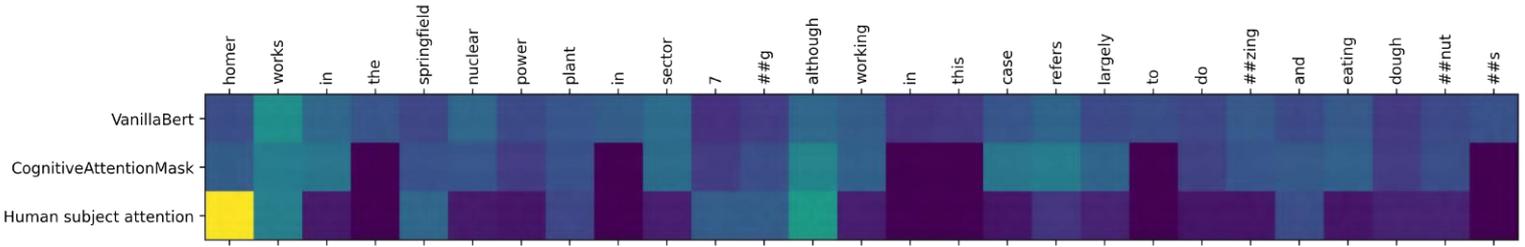

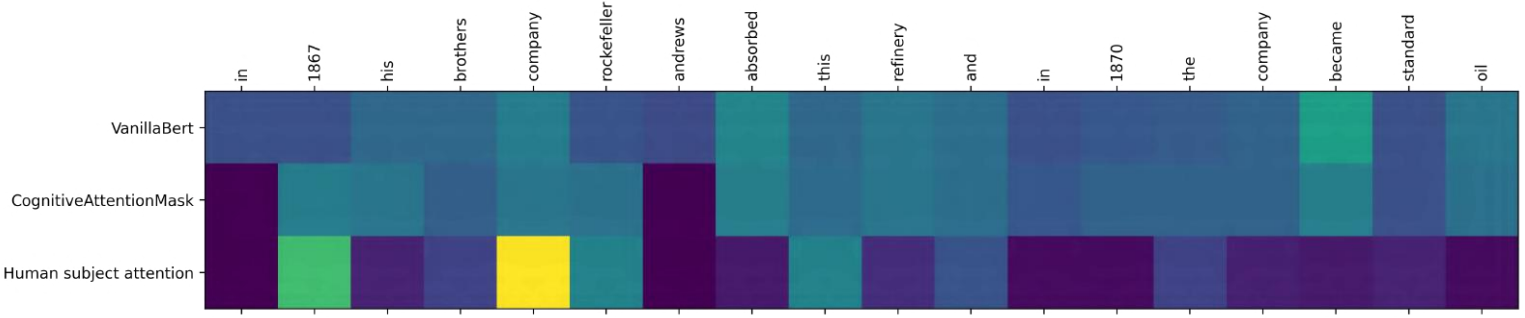

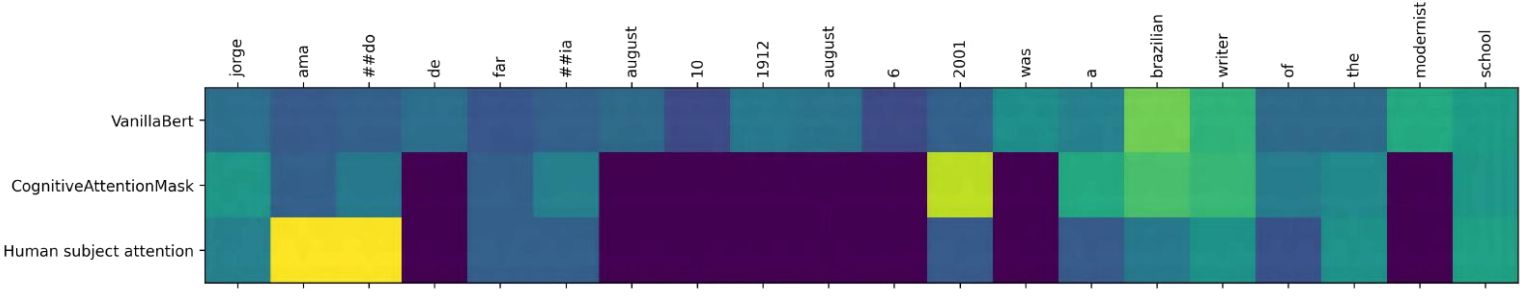

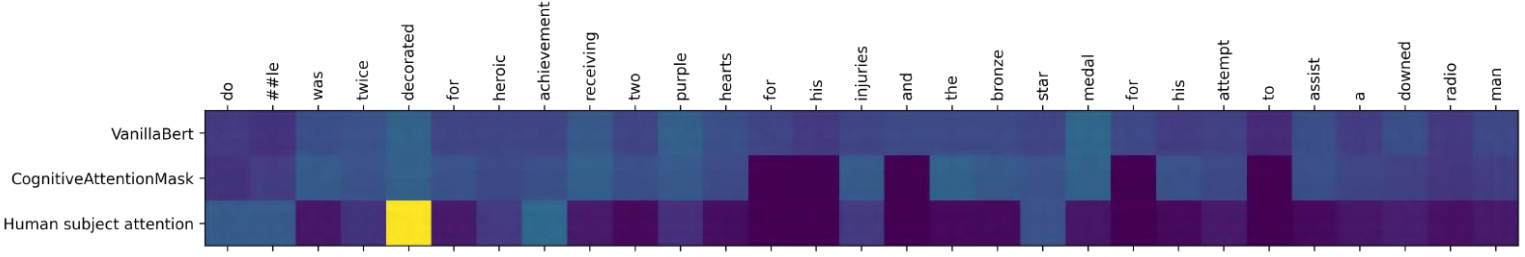

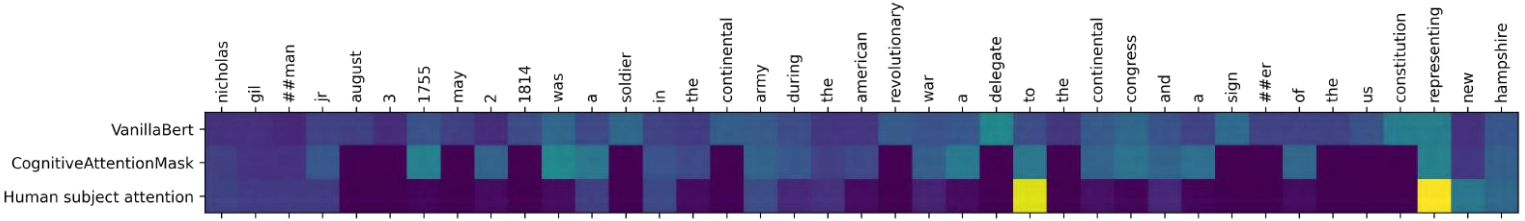

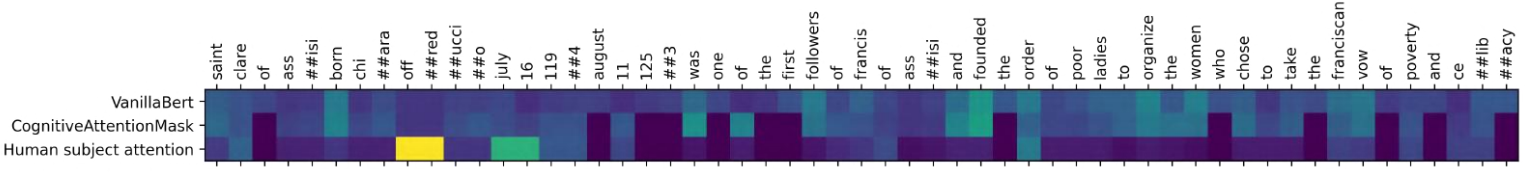

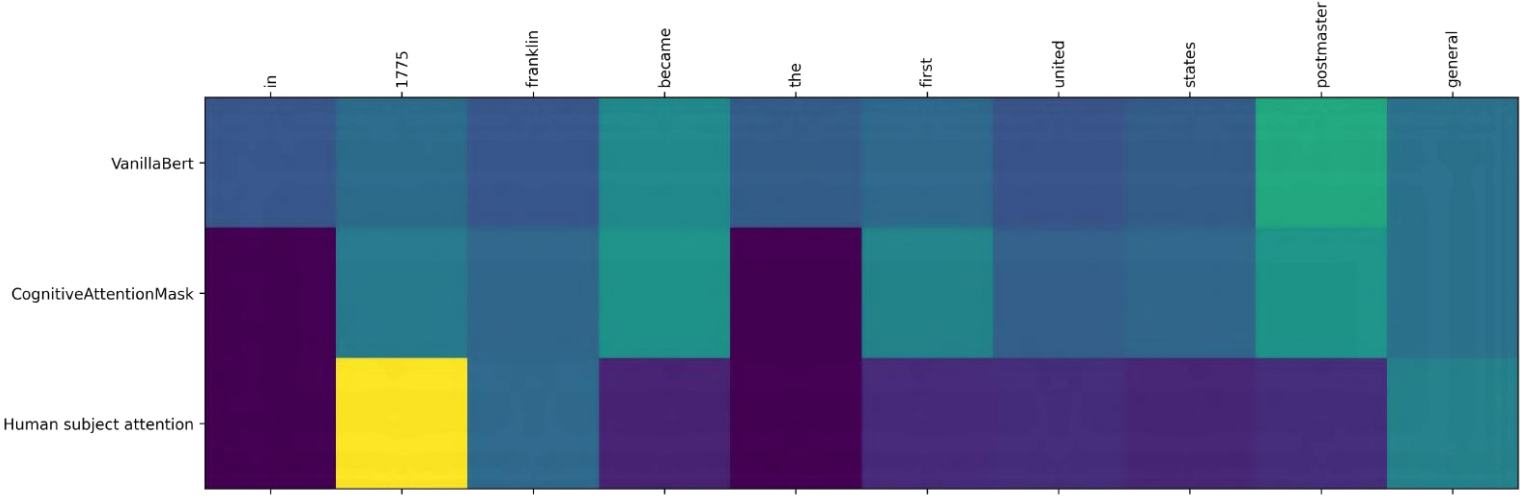

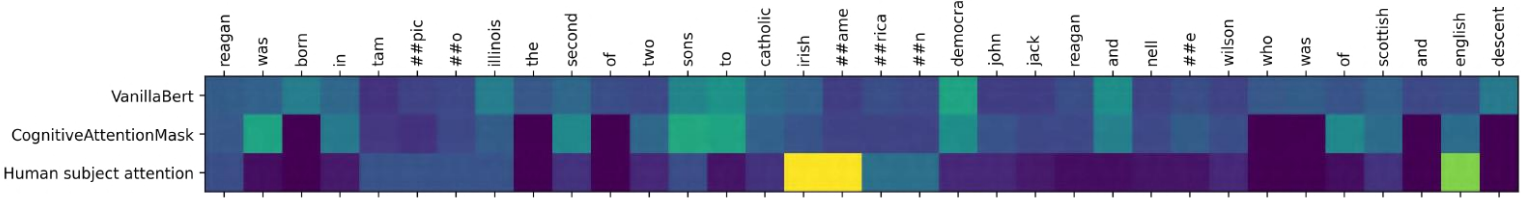

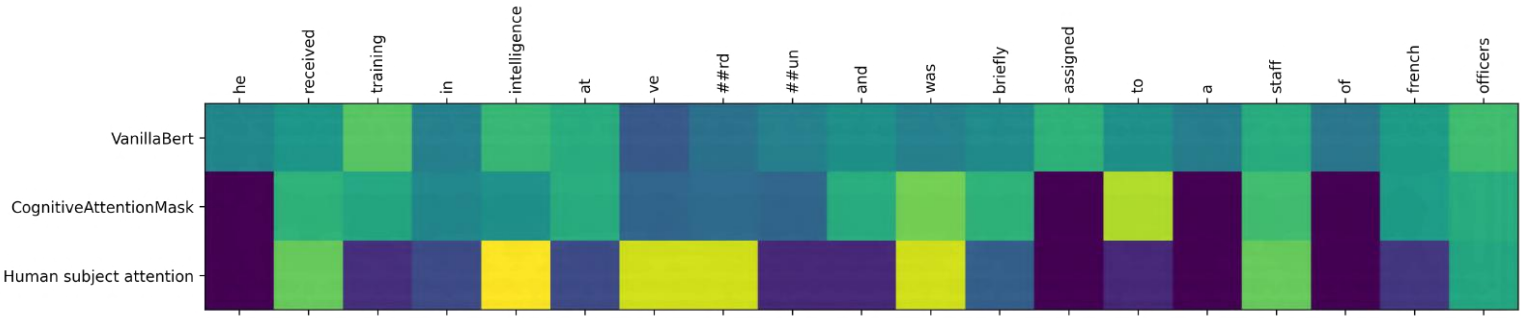

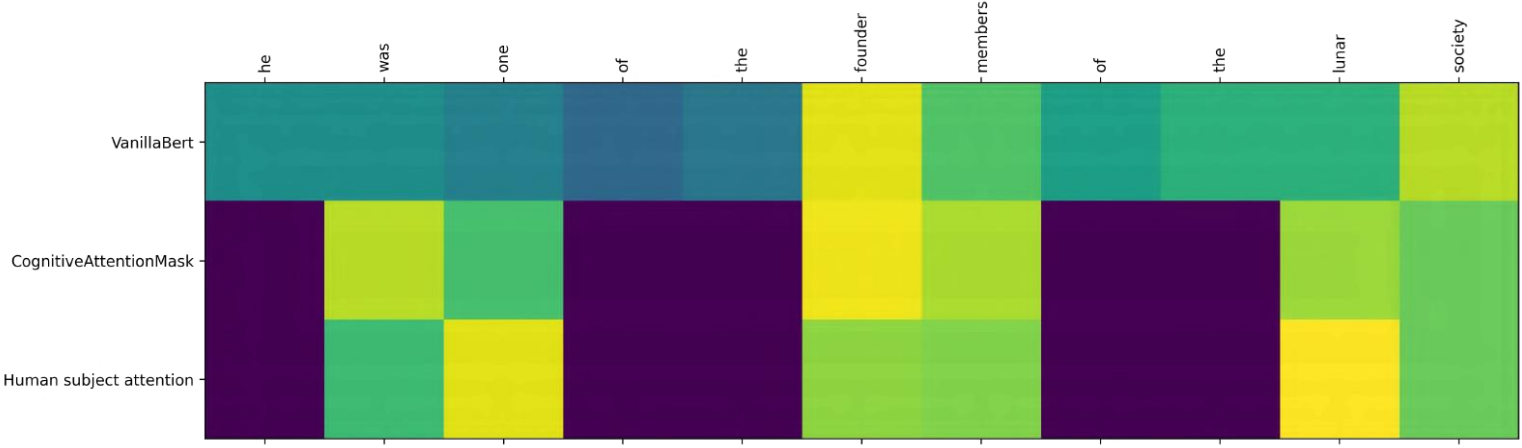

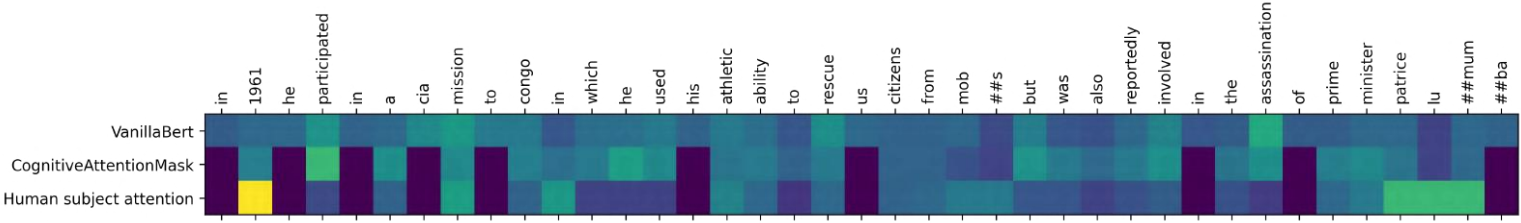

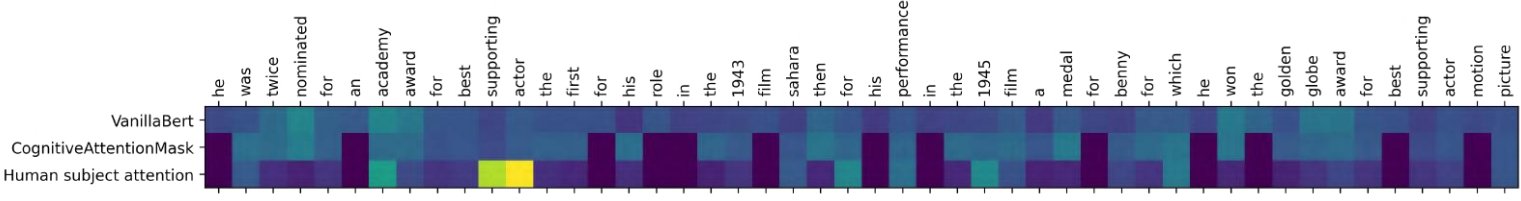

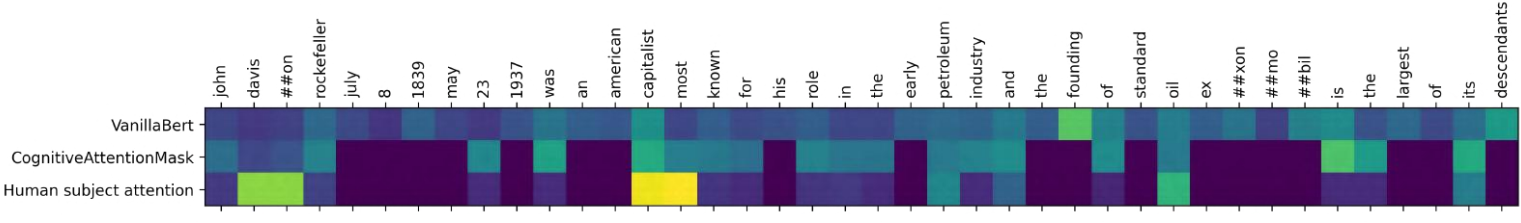

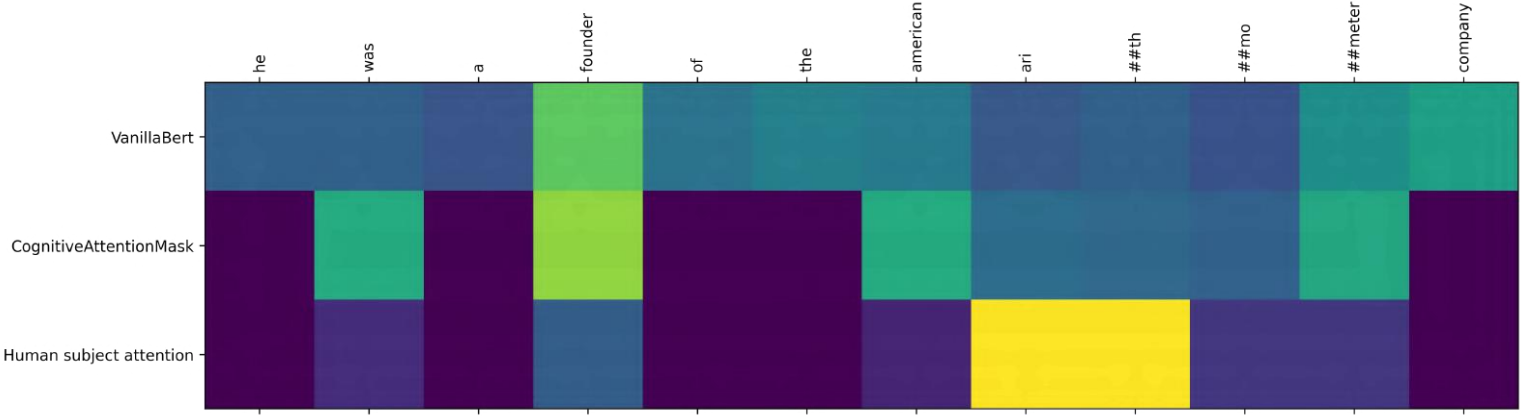

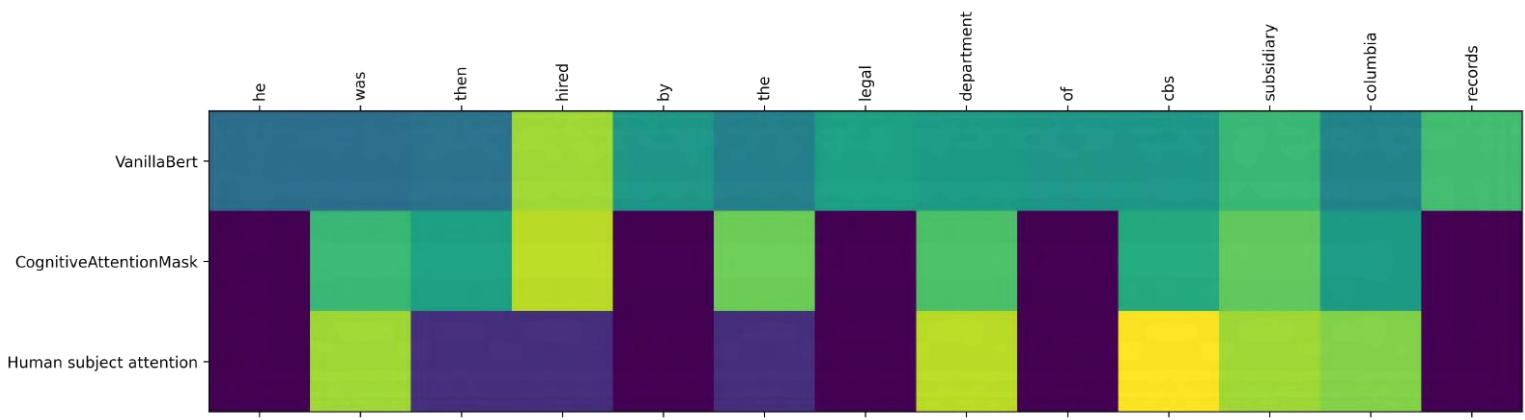